\documentclass[12pt,a4paper]{report}
\usepackage[left=2cm,right=2cm,bottom=2cm]{geometry}

\usepackage[utf8]{inputenc}
\usepackage{bm}
\usepackage{graphicx}
\usepackage{amsmath}
\usepackage{hyperref}
\hypersetup{
    colorlinks=true,
    linkcolor=blue,
    urlcolor=blue,
    citecolor=blue
}
\usepackage{mathrsfs}
\usepackage{amssymb}
\usepackage{amsfonts}
\usepackage{latexsym}
\usepackage{amsthm}
\usepackage{longtable}
\usepackage{subcaption}
\usepackage[linesnumbered,ruled,vlined]{algorithm2e}
\usepackage[toc, page, title]{appendix}

\usepackage{color}

\usepackage{array}
\newcolumntype{P}[1]{>{\centering\arraybackslash}p{#1}}

\newcommand\norm[1]{\left\lVert#1\right\rVert}

\usepackage{pifont}
\newcommand{\cmark}{\ding{51}}%
\newcommand{\xmark}{\ding{55}}%

\usepackage[nottoc]{tocbibind}
\makeatletter
\renewcommand\p@subsubsection{\thesubsection.} 
\makeatother

\begin{document}

\thispagestyle{empty}

\begin{center}
   \vspace*{1cm}

   \textbf{\textsc{A PROJECT REPORT ON}} \\
   \vspace{0.5cm}
   \textbf{{\Large LEARNING CONTROL POLICIES}} \\
   \vspace{0.4cm}
   \textbf{{\Large FOR IMITATING HUMAN GAITS}}

   \vspace{0.5cm}
   
   \textit{Submitted in partial fulfillment of the \\ requirements for the award of the degree of}

    \vspace{0.5cm}
   \textbf{\textsc{BACHELOR OF TECHNOLOGY}} \\
   \vspace{0.2cm}
   \textbf{\textit{in}} \\
   \vspace{0.2cm}
   \textbf{\textsc{MECHANICAL ENGINEERING}} \\
   
   \vspace{0.75cm}
   
   \textit{Submitted by:}\\
   \vspace{0.3cm}
   \textbf{Utkarsh Aashu Mishra}
   
   \vspace{0.75cm}
   
   \textit{Supervised by:}\\
   \vspace{0.3cm}
   \textbf{Prof. Pushparaj M. Pathak} \\
   \vspace{0.2cm}
   \textit{and} \\
   \vspace{0.2cm}
   \textbf{Prof. Auke J. Ijspeert} \\
        
   \vspace{0.5cm}

   \begin{figure}[ht]
   \centering
    \begin{subfigure}{.4\textwidth}
      \centering
      \includegraphics[width=0.9\linewidth]{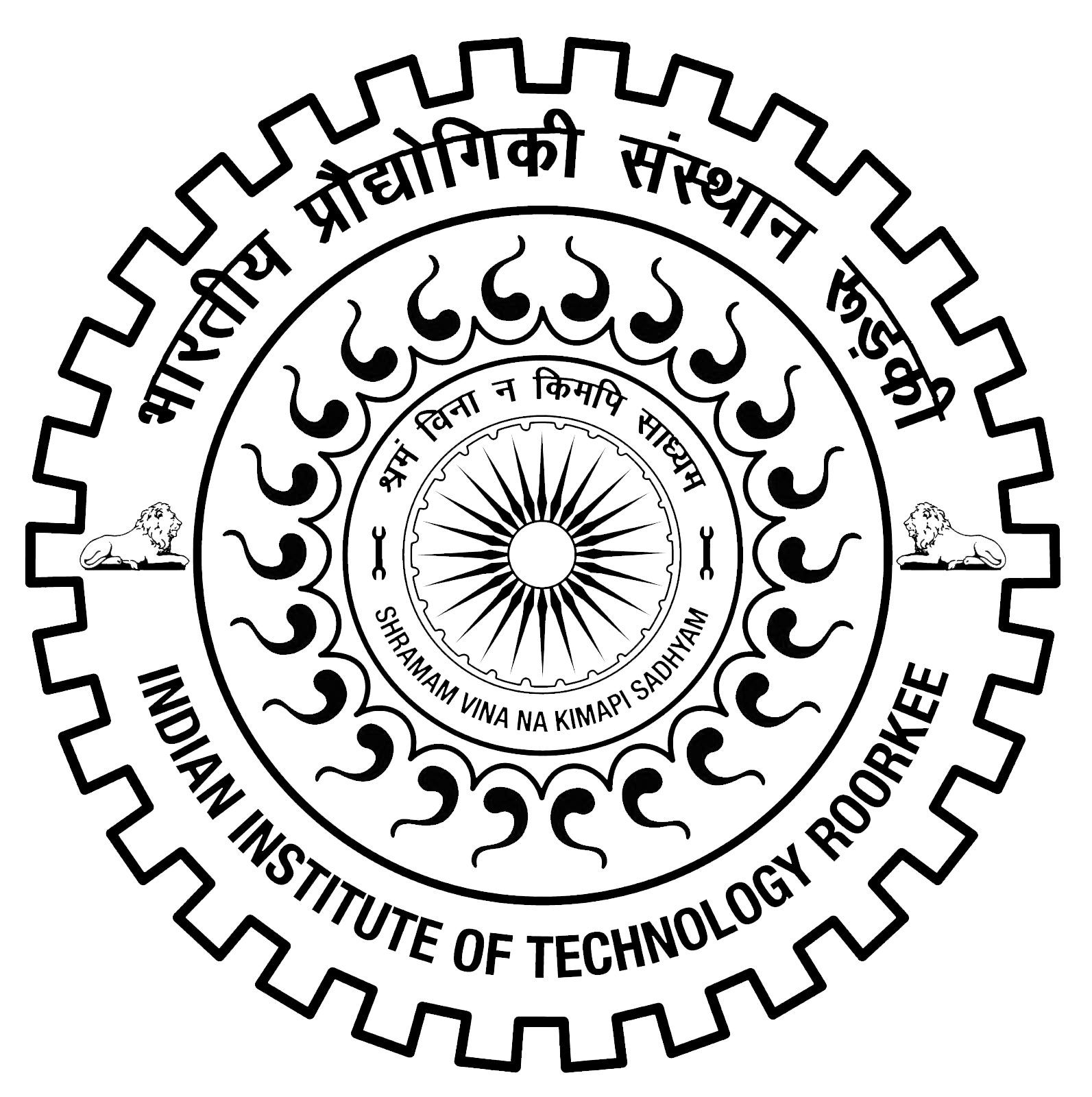}
    \end{subfigure}%
    \begin{subfigure}{.4\textwidth}
      \centering
      \includegraphics[width=0.9\linewidth]{./epfl-biorob.png}
    \end{subfigure}
    \end{figure}
        
    \vspace{0.5cm}
   \textbf{\textsc{DEPARTMENT OF MECHANICAL AND INDUSTRIAL ENGINEERING}} \\
   \vspace{0.2cm}
   \textbf{\textsc{INDIAN INSTITUTE OF TECHNOLOGY ROORKEE}} \\
   \vspace{0.2cm}
   \textbf{\textsc{ROORKEE - 247667 (INDIA)}} \\
   \vspace{0.2cm}
   \textbf{\textsc{APRIL, 2021}} \\
        
\end{center}

\newpage

\thispagestyle{empty}

\chapter*{Declaration by Candidate}

I hereby declare that the work carried out in this report entitled, \textbf{``Learning Control Policies for Imitating Human Gaits"} is presented for the subject MIN-400B (B. Tech Project) and submitted to the Department of Mechanical and Industrial Engineering, Indian Institute of Technology(IIT) Roorkee (India). The work is an authentic record of own work carried out during the period from \textbf{September~2020} to \textbf{April~2021} under the supervision of Prof.~Pushparaj~M.~Pathak, MIED IIT Roorkee and Prof.~Auke~J.~Ijspeert, Biorobotics Laboratory, EPFL Switzerland. The record embodied in this project report is not submitted for the award of any other degree or diploma in any institute.

\begin{figure}[ht]
      \includegraphics[scale=0.08]{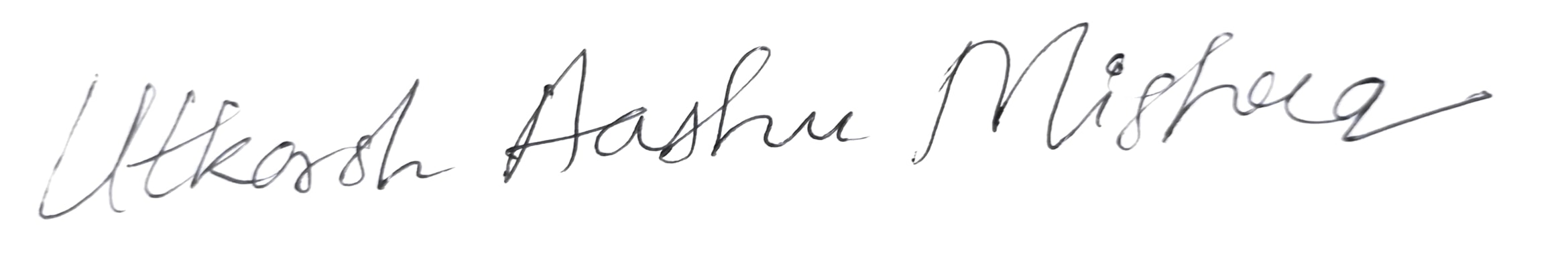}
\end{figure}
\noindent
\_\_\_\_\_\_\_\_\_\_\_\_\_\_\_\_\_\_\_\_\_\_\_\_\_\_\_\_\_\_\_\_\_\_\_\_\_\_\_\_\_\_\_\_\_\_\_\_\_\_\_ \\
Utkarsh Aashu Mishra \\
17117093, B. Tech 4th Year\\
Mechanical and Industrial Engineering Department \\
Indian Institute of Technology(IIT) Roorkee (India)

\bigskip
\bigskip

\noindent
April 30, 2021 \\
Roorkee, Uttarakhand

\bigskip
\bigskip

\hrule

\bigskip
\bigskip

\noindent
This is to certify that the above statement made by the candidate is correct to the best of my
knowledge.

\bigskip
\bigskip
\bigskip
\bigskip

\noindent
\textbf{Prof. Pushparaj M. Pathak \\
Professor \\
Mechanical and Industrial Engineering Department \\
Indian Institute of Technology(IIT) Roorkee (India)
}

\thispagestyle{empty}

\newpage

\pagenumbering{Roman}

\chapter*{Acknowledgement}
\addcontentsline{toc}{chapter}{Acknowledgement}

First and foremost, I would like to express my sincere gratitude towards my supervisors, \textbf{Prof.~Pushparaj~M.~Pathak} and \textbf{Prof.~Auke~J.~Ijspeert}, for their support and the freedom they afforded me in defining my project. I want to convey my special thanks to Prof.~A.~J.~Ijspeert for giving me this opportunity to conduct my thesis in collaboration with his Biorobotics Laboratory at the Swiss Federal Institute of Technology Lausanne (EPFL). I want to thank \textbf{Dr.~Dimitar~Stanev} for continually guiding me at every step of the project, helping me realize concepts, and constantly motivating me to achieve more. I have covered a lot of ground in learning to be an independent researcher, and it would not have been possible without their guidance and unwavering belief in my ability.

I would also like to thank everybody in the laboratory for creating an environment conducive to research and intellectual discourse. I would particularly like to thank Nikolaos~Kokkinis~Ntrenis and Lionel~Blond\'{e} for all the stimulating discussions. Towards the evaluation panel, I would like to thank Prof.~S~H.~Upadhyay, Prof.~D~M.~Joglekar, and Prof.~V.~Gaur for their constant evaluations and suggestions to leverage the quality of the presentation. Finally, I express my thankfulness to my family and friends for their continuous support during this venture.

\bigskip
\bigskip
\bigskip

\noindent
Sincerely,\\
Utkarsh Aashu Mishra

\newpage

\chapter*{Abstract}
\addcontentsline{toc}{chapter}{Abstract}

The work presented in this report introduces a framework aimed towards learning to imitate human gaits. Humans exhibit movements like walking, running, and jumping in the most efficient manner, which served as the source of motivation for this project. Skeletal and Musculoskeletal human models were considered for motions in the sagittal plane, and results from both were compared exhaustively. While skeletal models are driven with motor actuation, musculoskeletal models perform through muscle-tendon actuation. Model-free reinforcement learning algorithms were used to optimize inverse dynamics control actions to satisfy the objective of imitating a reference motion along with secondary objectives of minimizing effort in terms of power spent by motors and metabolic energy consumed by the muscles. On the one hand, the control actions for the motor actuated model is the target joint angles converted into joint torques through a Proportional-Differential controller. While on the other hand, the control actions for the muscle-tendon actuated model is the muscle excitations converted implicitly to muscle activations and then to muscle forces which apply moments on joints. Muscle-tendon actuated models were found to have superiority over motor actuation as they are inherently smooth due to muscle activation dynamics and don't need any external regularizers. Finally, a strategy that was used to obtain an optimal configuration of the significant decision variables in the framework was discussed. All the results and analysis are presented in an illustrative, qualitative, and quantitative manner. Supporting video links are provided in the Appendix. All simulation are done using \textregistered \textsc{Opensim}.

\bigskip
\bigskip

\noindent
\textbf{Keywords:} Artificial Intelligence, Machine learning, Reinforcement Learning, Imitation, Bipedal Locomotion, Musculoskeletal Human Models

\bigskip
\bigskip
\bigskip

\newpage
\tableofcontents
\listoffigures
\listoftables

\newpage

\pagenumbering{arabic}

\chapter{Introduction}

Movement is essential for human well-being, and its interpretation is essential for characterizing healthy and pathological conditions. Implementing robust controllers for achieving complex movements through simulation is an active research area that provides us with important insights. The proposed project aims towards simulating healthy movement based on learned control policies while trying to imitate them. Notably, the example of such movements can be walking, running, etc. Based on reference motion capture trajectories, a controller is established using the state-of-the-art Deep Reinforcement Learning (DRL) based control policies. The project's main contribution is an imitation controller that can effectively control a full-body skeletal and musculoskeletal model to achieve the desired gait, being robust to disturbances, and provide the inverse dynamics analysis. The work will enable an analysis of pathological gaits using learned mimicking of healthy gait policies and enhancing the development of clinically relevant research methods.

\section{Motivation}

The motivation of the project lies in the convergence of robotics and biomechanics with the application of reinforcement learning to define more robust and healthy gaits. There is growing popularity of reinforcement learning as an optimal control strategy, and it competes well with effective single-shooting and collocation-based methods. However, dealing with continuous torques produces spiky output signals, which are then smoothened using various filtering regulators like low pass, finite and infinite impulse. In this sense, humans perform inherently smooth motions with implicit regulations. This comes from the muscle-tendon actuated architecture and control of human gaits. Muscle actuation is key to understanding more “human-like” movements. In terms of biomechanics, imitating a subject task eventually leads to learning the task, and once a task is learned, it can be improved on various other criteria. For example, if we know how to walk, we can walk at varying speeds based on our step timing and frequency. Considering a pathological subject task, the motion capture data of their defected gaits can be learned through imitation. This opens the door to a wide range of analyses on virtual surgeries and their remedial effects on bringing a pathological gait closer to a healthy gait.

\section{Highlights}

The main highlights of the presented work are as follows:
\begin{enumerate}
    \item \textbf{A Modular Framework:} Introduces a modular framework as a test-bed for various reinforcement learning and control algorithms. The framework consists of skeletal motor actuated human model environments with a particular interest in exploring human movements from a robotics perspective. It also consists of musculoskeletal muscle-tendon actuated human model environments to understand and analyze various physiological gaits. Thus, the framework lies on the boundaries of robotics and bio-mechanics and takes inspiration from both.
    \item \textbf{Analysis of Walking gait:} An extensive analysis of walking gait with motor and muscle-tendon actuated models is drawn based on their joint moment profiles, quality of resulting gait, and learning of behaviors with minimal data. With data of only a single gait cycle, the resultant motion could generalize for infinite walking. Further, it was observed that the movements generated by the muscle-tendon actuators were very smooth and lacked any spikes accompanied by the motions with motor actuation.
    \item \textbf{Analysis of Multiple Movements:} To validate the modularity of the framework and analyze more movements in addition to walking, common day-to-day activities like running and jumping were performed, and the imitated motions were analyzed. Both high jumps and continuous jumping patterns were visualized and achieved by modulating the use of the phase variable. Further, pathological models with locked knees were trained to follow the healthy walking gait, and the resultant motion was found intuitive to how people with similar leg disabilities behave naturally.
    \item \textbf{Extensive Analysis of the Observations and Reward terms:} The terms included in the observation space and reward function were analyzed based on the performance of the resulting motion being trained with or without their presence. It was concluded that the model requires the knowledge of the relation between the generalized coordinate space and task space to generate a feasible motion and the foot positions to learn a stable limit cycle based on heel strikes. Further, the activation states of the muscles are required for a better non-linear approximation of the activation dynamics. Most importantly, the phase variable corresponding to the progress level of any gait helps the controller understand the periodicity between observations and rewards. 
\end{enumerate}

\section{Report Organization}

Following this, various concepts and terminologies are discussed in Section \ref{sec:related} along with an exploration of the previous associated works in the domains around which this project revolves. The structure of the \textsc{Opensim} environment and the functioning of the skeletal and musculoskeletal models are discussed in Section \ref{sec:osim_envs}. Section~\ref{sec:algorithms} describes the reinforcement learning algorithms used to validate the framework as well as the formulation of various terms of the reward formulating the surrogate objective. Section~\ref{sec:results} gives the simulation results obtained using the proposed approach for both the motor and muscle-tendon actuated models for walking, running, and jumping. Additional details on the decision variables are also provided in this section. Those results are extensively discussed in Section~\ref{sec:discuss}, and significant findings are accumulated. Conclusions and future work are drawn in Section~\ref{sec:conclusion}. Finally, Appendix \ref{appendix:A} provides extra information on the other variants of reinforcement learning testbeds explored during the project, and Appendix \ref{appendix:B} provides all the video links for the movements and comparisons generated and used in this project.

Hereafter, a healthy gait is used to refer to a movement exhibited by a patient with no disorders, and the framework is in no way constrained to the model and reference data used in the experiments. The framework is indeed modular to such changes. Finally, terms like agent, policy, and actor are used to represent the controller from reinforcement learning algorithms, and these terms can be used interchangeably.

\section{Concepts, Frameworks and Related Work}
\label{sec:related}

\subsection{OpenSim}

The project solely uses OpenSim \cite{c1} framework to conduct simulations with the musculoskeletal and skeletal models. OpenSim is open-source software for biomechanical modeling, simulation, and analysis. The framework enables a wide range of studies, including analysis of human gait dynamics, studies of sports performance, simulations of surgical procedures, analysis of joint loads, design of medical devices, and animation of both human and animal movements. It gives us access to perform complex inverse dynamics analysis and forward dynamics simulations. 

Using the OpenSim simulator, we set out to make a simulated human model learn a controller coordinating its limbs to align with the succession of recorded poses from patient gait. Such coordination is first done via the application of torques on its joints and then on a more complex scenario in which the simulated agent directly applies activations on its muscles.

\subsection{Reference Data: Formulation and Types}

Reference data forms the basis of imitation and is generated via motion capture techniques using body markers and sensors for kinematics and ground reaction forces (GRFs). Motion Capture is the technique of recording the movement of various segments of the subject under concern in the form of markers. The OpenSim - SimTrack \cite{c1} enables researchers to generate dynamic simulations of movement from motion capture data. An inverse kinematics (IK) problem is solved to determine the model generalized coordinate values that best reproduce the raw marker data obtained from motion capture. This is formulated as a least-squares problem that minimizes the differences between the measured marker locations and the model’s virtual marker locations, subject to joint constraints.

In this project, the formulation is used to develop the desired trajectories based on higher-order objective functions and hence helped in the development of the basis of performing the learning experiments. The formulation eventually provides a deeper insight into algorithms already provided by OpenSim, such as Computed Muscle Control(CMC) \cite{c2} algorithm. In order to get fine data and verify the algorithmic flow of the process, we use data generated from predictive simulation in OpenSim-SCONE \cite{c7} and closely resembling experimental data without noises and miscalibration. 

The formulations of solving dynamic models of human skeletal and musculoskeletal systems are being addressed employing various optimization techniques, dealing with optimal control problems for data tracking and predictive simulations. Predictive simulation is a method to predict novel movements without any reference trajectory. Such simulations are used to develop optimized gaits based on a high-level objective function and required tasks. Predictive simulation can be solved through various trajectory tracking and optimal control methods. In contrast, in the tracking case, the objective is to minimize the difference between the model's behavior and a target set of experimental data, such as joint kinematics and ground reaction forces. To achieve both these cases, Direct Collocation (DC) \cite{c4} methods are used as a form of optimal control \cite{c5}. Furthermore, various open-source frameworks like OpenSim MOCO \cite{c6}, and OpenSim SCONE \cite{c7} continuously solve predictive simulation problems using the same technique. 

\begin{figure}[hp]
      \centering
      \includegraphics[width=\linewidth]{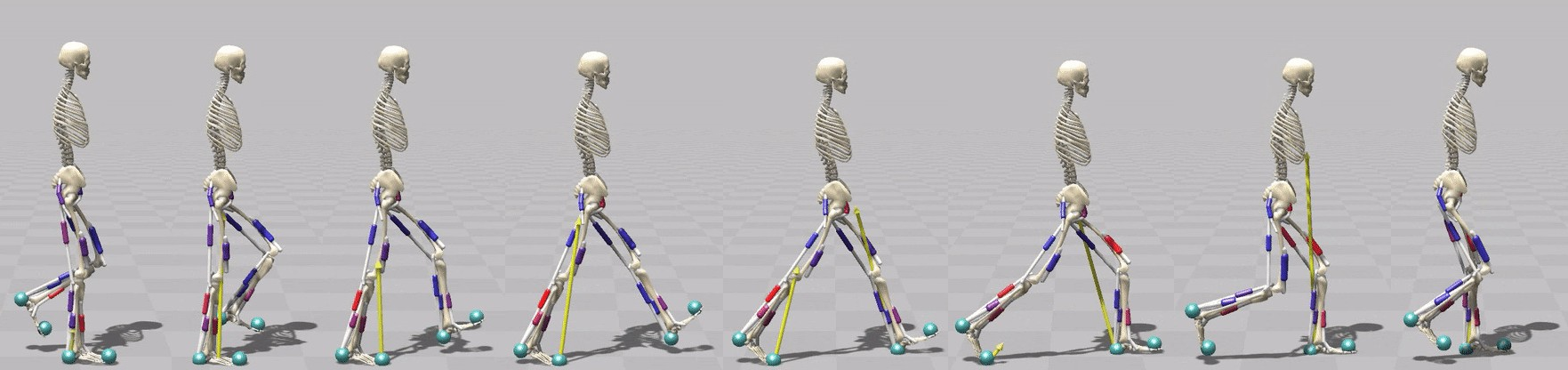}
      \caption{Walking Gait with Predictive Simulation based on Direct Collocation}
      \label{fig:intro}
   \end{figure}
   
SCONE solves to obtain a controller using a DC approach by getting piece-wise control curves based on a high-level objective. The objective function consists of minimizing the energy consumption, which is the muscles' metabolic rate and minimizing the limiting forces experienced by the joints. A joint experiences a limiting force as a compensation for the extra moment applied to it. This keeps the joints bounded to respect physical constraints. For musculoskeletal systems, control curves for the excitation of each muscle is obtained using the direct-collocation approach.

\subsection{Reinforcement Learning based Imitation}

Human intelligence has proved to be the most advanced learning system. The nature of the human mind to learn specific tasks and generalize the skills learned from the tasks to perform a similar unseen task has been a motivation for scientists and researchers worldwide. This is yet to be achieved by developing autonomous systems and intelligent robots. While such agents may find it challenging to learn new tasks with the same agility and coherence, as depicted by humans, they might take inspiration from humans. One such concept lies in imitating humans' behaviors for a specific task and acquiring skills from human demonstrations.

Imitation learning, also known as Learning from Demonstration, has been successfully applied to solve many different tasks in complex domains such as helicopter flight, playing table tennis, and accomplishing human-like reaching motions, among many others. While imitation learning has the benefit of comparatively fast skill acquisition, it requires high-quality demonstration of a human expert. In this project, the technique is used to imitate based on the Predictive simulation optimized gait results, and the human demonstrated motion capture results.

While imitation learning purely depends on the reference data and resembles more to supervised techniques, reinforcement learning-based imitation considers both the reference data and the added objectives. The goal is to imitate the reference data along with minimizing the overall supplementary objective function. For example, while we see someone doing a task, we know the way their limbs move, but we don't know the most optimal way to move them. The human controller cumulatively imitates the movement as well as performs it most efficiently. Here, efficiency is strictly concerning the added objective.

With DC, on the one hand, various researches have explored the effectiveness of Reinforcement Learning \cite{c8} as a solution to robust learning and developing skillset generalizable over multiple tasks. Various physics-based methods with RL settings have shown remarkable results by imitating reference human motions and comparing with different action parameterizations \cite{c9} comprising of torques, muscle-activations, target joint angles, and target joint angle velocities. The works have evaluated the choice of action space in terms of learning time, policy robustness, motion quality, and policy query rates. DeepMimic \cite{c10} successfully achieved imitation behavior, along with learning additional task objectives. Although the concept's dynamics compensation made the black box dynamics of the environment more straightforward, the strategy of providing dynamic Stable Proportional-Derivative control opened up further explorations.

Muscle-tendon activation-based human simulation and control over varying tasks and pathological conditions \cite{c11} also introduced an interesting supervised learning framework inside the RL agent. The setup optimizes muscle activations based on joint accelerations and trains a supervised network using the demerit of large samples required for RL algorithms. Parallelly, interesting works focussed on specific tasks like playing basketball \cite{c12} to modulate RL agents in order to perform them precisely. All these simulations are different from what humans face in real-time while performing the actions themselves. To bridge this Sim-to-Real gap, residual forces are introduced \cite{c13} which tend to apply various invisible dynamic forces to sustain the RL agent from getting into the traps of simulations and exhibit precise motion tracking.

\newpage

\chapter{Methodology}

\section{OpenSim Environment}
\label{sec:osim_envs}

\begin{figure}[hp]
      \centering
      \includegraphics[width=0.8\linewidth]{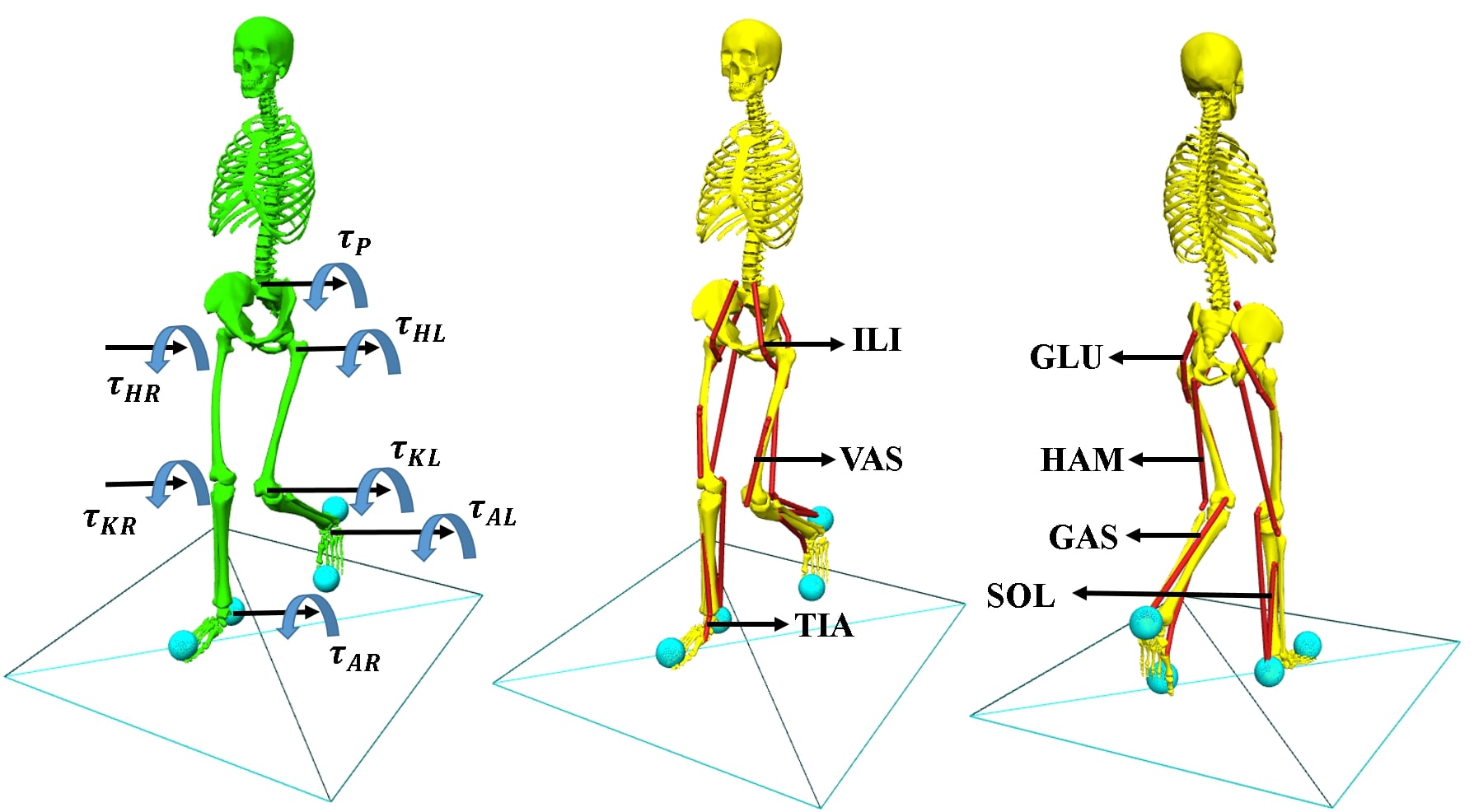}
      \caption{The Skeletal (Green - Torque Actuated, Left) and Musculoskeletal (Yellow - MTU Actuated, Center-Front, Right-Back) Human Models used in the experiments}
      \label{fig:modelpic}
   \end{figure}

The musculoskeletal model is made up of simulated bones, joints, and muscle-tendon units (MTU) as depicted in Fig. \ref{fig:modelpic}. Since the task is to design and implement a musculoskeletal controller, the agent’s outputs — the controls — effectively are either torques to be applied on joints or muscle excitations to be applied on each of the constituting MTU. When the model is torque actuated, the muscles are removed, and it becomes a skeletal model.

The representations $\tau_\square$ denotes the torque at a joint, and the other abbreviations mentioned in Fig. \ref{fig:modelpic} are described in the table below:

\begin{longtable}{|c|c|c|c|}
\hline
\textbf{Abbreviation} & \textbf{Meaning} & \textbf{Abbreviation} & \textbf{Meaning}\\
\hline
\multicolumn{2}{|c}{\textbf{Torque Actuated}} & \multicolumn{2}{|c|}{\textbf{MTU Actuated}} \\
\hline
P & pelvis tilt (torso) &  HAM &  Hamstring\\
\hline
HR &  Hip Flexion Right &  GLU &  Gluteus Maximus\\
\hline
KR &  Knee Angle Right &  ILI &  Iliopsoas\\
\hline
AR &  Ankle Angle Right &  VAS &  Vastii\\
\hline
HL &  Hip Flexion Left &  GAS &  Gastrocnemius\\
\hline
KL &  Knee Angle Left &  SOL &  Soleus\\
\hline
AL &  Ankle Angle Left &  TIA &   Tibialis Anterior\\
\hline
\caption{Model Abbreviations}
\label{abbreviations}
\end{longtable}

\subsection{Models}
\label{sec:osim_envs_models}

Due to the high computational cost, a simplified model with fewer DoFs was selected to carry out our prototyping workload. The overall workflow consisted of evaluating several reinforcement learning algorithms with different track records on locomotion tasks and cross-validating several reward (reinforcement signal) function that was hand-engineered for the task at hand.

The more complex model is used with the set of hyperparameters considered the best from our preliminary prototyping efforts. We consider a 2D - model with 7 degrees of freedom (DoFs) and 14 muscles in our experiments. By design, the agent's state/observation and control vectors (respectively, input and output vectors) depend directly on the number of DoFs of the agent. The higher the number of DoFs, the more complex the simulation, making the controller harder to learn.

\subsubsection{Torque Actuated Model}

The torque actuated model resembles a robotic system with motors used to actuate using torques. Derived from the Lagrangian formulation, the dynamics for such a system can be simply given by:
\begin{equation}
    \pmb \tau = \bf D(\bf q)^{-1} \left(  \bf M(\bf q)\bf \ddot{q} + \bf C(\bf q, \bf \dot{q}) \bf \dot{q} - \bf J^T_c \bf f_c - \pmb \tau_{ext}\right) 
\end{equation}
where $\bf f_c$ is the vector of the constrained forces and $\pmb \tau$ is the vector of the joint torques. $\bf M, \bf C$ and $\bf J_c$ represent the Mass matrix, Coriolis matrix, and the Jacobian matrix as a function of the generalized coordinates and their velocities. The Jacobian matrix maps the constraint force vector to the generalized coordinate space. The contact dynamics with the ground is modeled using the Hunt-Crossley model as inelastic collisions. The ground reaction force, $F_{GR}$ is given by:
\begin{gather}
    F_{GR} = F_S + F_D \\
    F_S = k(x^n) \\
    F_D = \lambda (x^n) \dot{x}
\end{gather}
where, the GRF, $F_{GR}$ depends on the displacement, $x$ and displacement rate, $\dot{x}$, of the contact sphere (blue spheres in the feet in Fig. \ref{fig:modelpic}). The stiffness parameter, $k$, and the power parameter, $n$, can be found through Hertz's elastic theory and $\lambda$ represents the damping parameter. We rely on our neural network-based non-linear approximation to model the GRF relations with the inverse dynamics.

\subsubsection{Muscle Tendon Unit (MTU) Actuated Model}

The Muscle Tendon Unit is simply a Hill-type model defined as a series combination of an elastic element with a parallel combination of an elastic and contractile element as shown in Fig. \ref{fig:muscle} below. 
\begin{figure}[ht]
      \centering
      \includegraphics[scale=0.25]{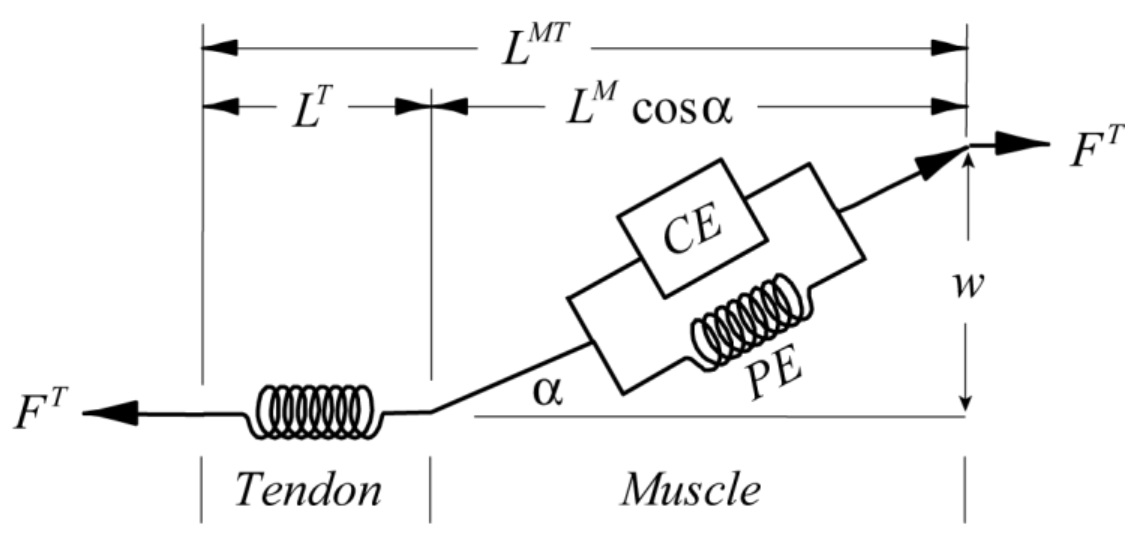}
      \caption{Mathematical representation of a Muscle-Tendon Unit}
      \label{fig:muscle}
   \end{figure}
The contractile element (CE) models the muscle fibers responsible for actively generating force ($F^{CE}$) depending on the current activation level (a). On the other hand, the parallel-elastic element (PE) models the passive forces ($F^{PE}$) generated by the muscle fibers, while the serial-elastic element (SE) models the tendon dynamics. The angle between the line of action of muscle fibers and the line of action of tendon forces is known as the pennation angle, $\alpha$.

The dynamics for this model is similar to that of the torque actuated model as given in Eq.~(1) except that the generalized forces due to the MTUs are added, eventually resulting in:
\begin{equation}
    \sum_{m \in \mathcal{M}} \bf J^T_m f_m(a_m) = \bf M(\bf \ddot{q}) + \bf C(\bf q, \bf \dot{q}) \bf \dot{q} - \bf J^T_c \bf f_c - \pmb \tau_{ext} 
\end{equation}
where $\mathcal{M}$ represents the set of muscles, $\bf J_m$ represents the Jacobian matrix to transform the muscle forces into the generalized coordinate space. The force due to a muscle is given by:
\begin{equation}
    f_m = F_{max} (a_m \times f_l \times f_v + f_{PE})
\end{equation}
where $F_{max}$ is the maximum isometric force that a muscle can apply. $f_l$ and $f_v$ are functions of the fiber length and fiber velocity of a muscle represented as a function of standard muscle behavioral curves. $f_{PE}$ is the passive force generated by the parallel elastic element. For more information, refer to \cite{wang2012}.

\newpage

\section{Observations and State Configuration}
\label{sec:osim_envs_obs}

\subsection{Observations for Motor actuation}

The observation and the state dictionary for the skeletal model were formulated as a 90 Dimensional state space. The table below documents the detail of each of the entities appearing in the dictionary. Each contact vector is a wrench vector consisting of the force and moment at the contact sphere. The contact wrench vector was normalized using the weight of the model for the contact forces and the weight moment (i.e.,~weight~=~75.164~Kg times the model's height~=~1.80~meter) for the contact moments. 

\begin{longtable}{|c|c|c|c|c|}
\hline
\textbf{Num} & \textbf{Name} & \textbf{Min} & \textbf{Max} & \textbf{Metric Unit}\\
\hline
0 & Gait Phase & 0 & 1 & None \\
\hline
\multicolumn{5}{|c|}{coordinate pos} \\
\hline
1 & pelvis\_tilt(rotation) &  -1.57079633 &  1.57079633 &  rad\\
\hline
2 &  hip\_flexion\_r(rotation) &  -2.0943951 &  2.0943951  & rad\\
\hline
3  & knee\_angle\_r(rotation) &  -2.0943951 &  0.17453293 &  rad\\
\hline
4  & ankle\_angle\_r(rotation) &  -1.57079633 &  1.57079633 &  rad\\
\hline
5  & hip\_flexion\_l(rotation) &  -2.0943951 &  2.0943951 &  rad\\
\hline
6  & knee\_angle\_l(rotation) &  -2.0943951 &  0.17453293 &  rad\\
\hline
7 &  ankle\_angle\_l(rotation) &  -1.04719755 &  1.04719755 &  rad\\
\hline
\multicolumn{5}{|c|}{coordinate vel} \\
\hline
8-14 &  all previous names &  unknown &  unknown &  rad/$s$\\
\hline
\multicolumn{5}{|c|}{coordinate acc} \\
\hline
15-21 &  all previous names &  unknown & unknown &  rad/$s^2$\\
\hline
\multicolumn{5}{|c|}{body pos relative to pelvis} \\
\hline
22-24 &  femur\_r\_(x,y,z)(position) &  -inf &  +inf &  meter\\
\hline
25-27 &  tibia\_r\_(x,y,z)(position) &  -inf &  +inf &  meter\\
\hline
28-30 &  talus\_r\_(x,y,z)(position) &  -inf &  +inf &  meter\\
\hline
31-33 &  calcn\_r\_(x,y,z)(position) &  -inf &  +inf &  meter\\
\hline
34-36 &  toes\_r\_(x,y,z)(position) &  -inf &  +inf  & meter\\
\hline
37-39 &  femur\_l\_(x,y,z)(position) &  -inf &  +inf &  meter\\
\hline
40-42 &  tibia\_l\_(x,y,z)(position) &  -inf &  +inf &  meter\\
\hline
43-45 &  talus\_l\_(x,y,z)(position) &  -inf &  +inf &  meter\\
\hline
46-48 &  calcn\_l\_(x,y,z)(position) &  -inf &  +inf &  meter\\
\hline
49-51 &  toes\_l\_(x,y,z)(position) &  -inf &  +inf  & meter\\
\hline
52-54 &  center\_of\_mass\_(x,y,z)(position) &  -inf &  +inf  & meter\\
\hline
\multicolumn{5}{|c|}{body vel} \\
\hline
55-90 & all previous names(of body) &  -inf  & +inf &  m/s\\
\hline
\caption{Observation and State Dictionary}
\label{obsmodel}
\end{longtable}

\subsection{Modifications for MTU actuation}

For MTU actuated models, the desired muscle properties were added for all the 14 muscles corresponding to the left and right limbs. The following Table \ref{tab:muscleparams} contains the characteristics added to the already existing observations as given in Table \ref{obsmodel}.

\begin{longtable}{|c|c|c|c|c|}
\hline
\textbf{Num} & \textbf{Name} & \textbf{Min} & \textbf{Max} & \textbf{Metric Unit}\\
\hline
91 & hamstrings\_r (activation) &  0 & 1 &  unknown\\
\hline
92 & glut\_max\_r (activation) &  0 & 1 &  unknown\\
\hline
93  & iliopsoas\_r (activation) &  0 & 1 &  unknown\\
\hline
94  & vasti\_r (activation) & 0 & 1 &  unknown\\
\hline
95  & gastroc\_r (activation) & 0 & 1 &  unknown\\
\hline
96  & soleus\_r (activation) &  0 & 1 &  unknown\\
\hline
97 &  tib\_ant\_r (activation) &  0 & 1 &  unknown\\
\hline
98 & hamstrings\_l (activation) &  0 & 1 &  unknown\\
\hline
99 &  glut\_max\_l (activation) &  0 & 1 &  unknown\\
\hline
100  & iliopsoas\_l (activation) &  0 & 1 &  unknown\\
\hline
101  & vasti\_l (activation) & 0 & 1 &  unknown\\
\hline
102  & gastroc\_l (activation) & 0 & 1 &  unknown\\
\hline
103  & soleus\_l (activation) &  0 & 1 &  unknown\\
\hline
104 & tib\_ant\_l  (activation) &  0 & 1 &  unknown\\
\hline
\multicolumn{5}{|c|}{Muscle Fiber Length} \\
\hline
105-118 &  all previous names &  unknown &  unknown &  unknown\\
\hline
\multicolumn{5}{|c|}{Muscle Fiber Velocity} \\
\hline
118-132 &  all previous names &  unknown & unknown &  unknown\\
\hline
\caption{Observation and State Dictionary added for MTU actuated tasks}
\label{tab:muscleparams}
\end{longtable}

\section{Actions and Control of Torque Actuated Model}
\label{sec:osim_envs_torque_acts}

The user provides a motion capture clip or a kinematic reference trajectory. The goal is to learn a control policy (a.k.a. controller) that produces motions that resemble the reference data. To achieve this, the model's action space is constructed as the joint configuration to be achieved in the next time step. Thus, the agent predicts the desired joint configuration passed through a PD controller to generate the required torques.

A possible argument can be fitting the data for the next step directly from the reference motion data. However, the values of the gains in the PD controller might not provide torques capable enough to perform precise tracking. Thus, the agent plans the following configuration, learning about the PD gains recursively. The joint torques based on the PD law ($K_P = 50$ for ankle and $100$ for rest; $K_D = 2$ for ankle and $5$ for rest) :

\begin{equation}
\pmb \tau = \bf K_P \times (\bf q_{pred} - \bf q_{curr}) + \bf K_D \times ( - \bf \dot{q}_{curr})
\end{equation}

The range of action values depends on whether we are working on joint space, torque space(joints), or muscle activation space. For muscles, the range is: [0, 1]. With regards to the torques, the range is [-1, 1], where this value is mapped automatically from OpenSim depending on the ``max actuation" we have set (default = 200Nm). The action space limits:

\begin{longtable}{|c|c|c|c|c|}
\hline
\textbf{Num} & \textbf{Name} & \textbf{Min} & \textbf{Max} & \textbf{Metric Unit}\\
\hline
0 & pelvis\_tilt(rotation) &  -1.57079633 &  1.57079633 &  rad\\
\hline
1 &  hip\_flexion\_r(rotation) &  -2.0943951 &  2.0943951  & rad\\
\hline
2  & knee\_angle\_r(rotation) &  -2.0943951 &  0.17453293 &  rad\\
\hline
3  & ankle\_angle\_r(rotation) &  -1.57079633 &  1.57079633 &  rad\\
\hline
4  & hip\_flexion\_l(rotation) &  -2.0943951 &  2.0943951 &  rad\\
\hline
5  & knee\_angle\_l(rotation) &  -2.0943951 &  0.17453293 &  rad\\
\hline
6 &  ankle\_angle\_l(rotation) &  -1.04719755 &  1.04719755 &  rad\\
\hline

\caption{Action Space}
\label{actmodel}
\end{longtable}

\section{Actions and Control of MTU Actuated Model}
\label{sec:osim_envs_muscle_acts}

The actions provided by the controller are the required excitations, denoted as $u$, for all the muscles, which is a one-dimensional vector of size 14. Then, the predicted excitation level is passed to the muscles. However, this does not imply that for a muscle with current activation state, $a$, directly jumps to an activation state, $u$. This transition is inherently smoothened and regulated by the activation dynamics, which is a differential equation given by:

\begin{equation}
    \frac{da}{dt} = \frac{u-a}{b(a,u)}
\end{equation}
and
\begin{equation}
    b(a,u) = \begin{cases}
t_{act}(0.5+1.5a) & :u > a\\
t_{deact}/(0.5+1.5a) & :u < a 
\end{cases}
\end{equation}
where, $t_{act}$ and $t_{deact}$ are the activation and deactivation time constants.

After getting the control actions, Eq.~(8) is solved, and the new activations are used to calculate muscle forces according to Eq.~(6) and integrate over one simulation timestep using Eq.~(5).

However, in any case, both the torque space and muscle actuation space are inter-convertible based on the following convex optimization:
\begin{gather*}
    \min_a \ \frac{1}{m} \sum_{m \in \mathcal{M}} a^p_m \\
    \text{Such that} \ \pmb \tau = \bf R(\bf q) \ F_{max} \ [a_m]_{activation}
\end{gather*}
where $\bf R(\bf q)$ is the moment arm matrix to convert muscle forces to joint momnets.

\section{Epsiode Termination}

Each episode is terminated when it fulfilled/satisfies one of following conditions:

\begin{itemize}
\item torso height $<$ 0.75 meter, this indicates that the agent is going to fall.
\item $|$max coordinate limiting forces$|$ $>$ 1000, to prevent the slow down of the integrator.
\item $|$max acceleration$|$ $>$ 10000, to prevent the agent from ”flying”.
\item simulation time $>=$ N (two walking steps of demonstration in N timesteps).
\end{itemize}

\newpage

\section{Training Data}
\label{sec:osim_envs_training}

The training data for the first phase comprised of the reference walking motion of a healthy individual. The gait data should ideally be produced by optimizing and inverse kinematics based on marker data from the motion capture experimental data of the subject. 
\begin{figure}[ht]
     \centering
      \includegraphics[width=\linewidth]{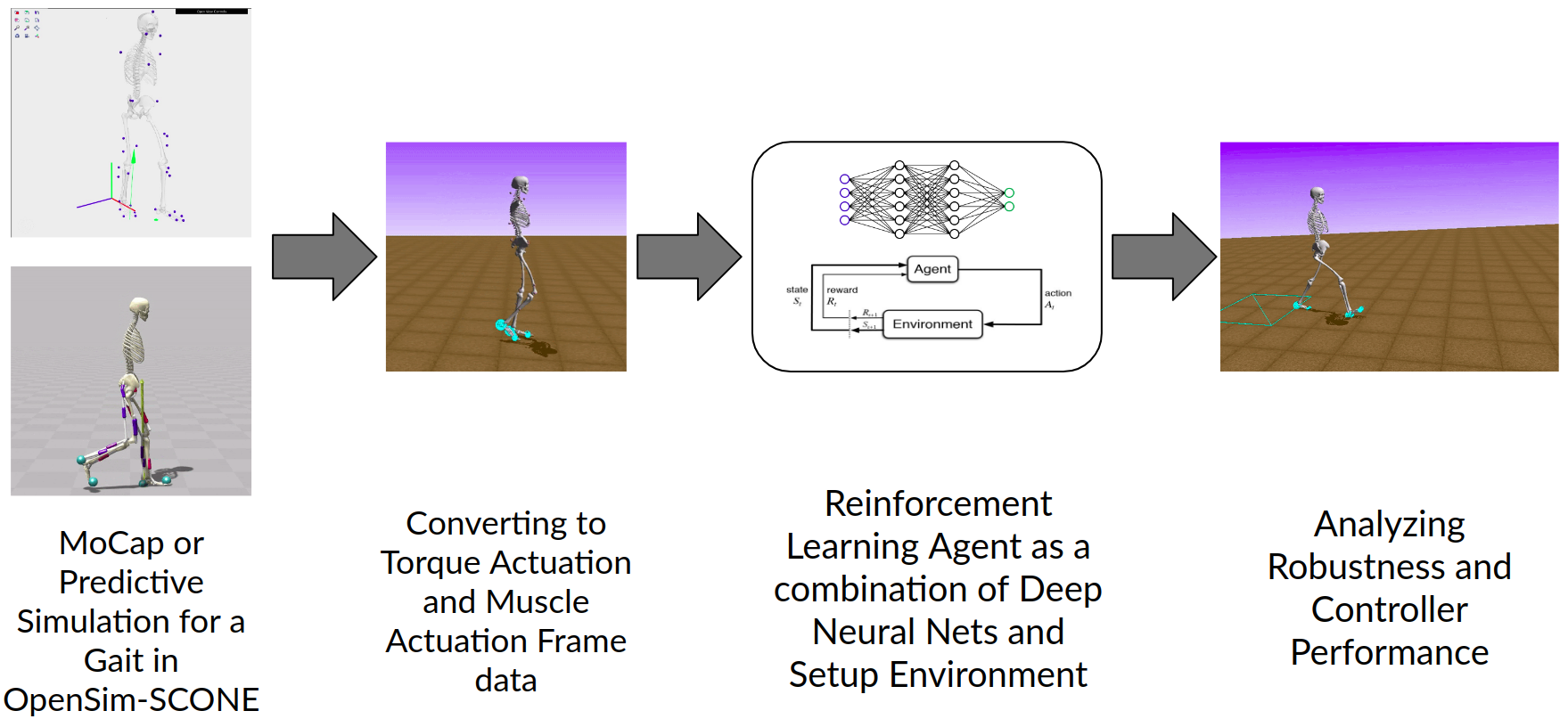}
      \caption{Methodology Followed for the Imitation Objective}
      \label{fig:method}
   \end{figure}
After the motion capture data provides the body position markers and velocities, inverse kinematics is solved with ground reaction forces (GRFs) and contact spheres. The GRFs are modeled as Hunt-Crossley Forces with specified stiffness, dissipation and static, dynamic, viscous frictions, and transition velocities coefficients. Specific joint damping, stiffness, and transition coefficients were employed to make the model more realistic. After solving the kinematics, the optimized gait was obtained, which minimizes the difference between model positions and marker data. The gait was converted to a torque actuated model by removing the muscles. This model behaves like a bipedal robot with motors for each joint coordinate.
   
The final static optimization was calculated to find the optimal muscle forces and required muscle activations to perform the desired movement. Muscle excitations for all the 14 concerned muscles were obtained and used as the ground truth values. For a given set of required joint torques at each joint, there can be multiple combinations of muscle activations possible. Hence, to reach optimal solutions, the ground truth activation values will play a significant role.
   
However, for simplicity, relatively noise-free and accurate reference data was obtained through predictive simulation via SCONE. The data contained the same information dictionary as can be obtained by the static optimization of the motion capture experimental marker data. The ground truth values of only the coordinate positions, velocities, and observable body positions and velocities were provided to the reinforcement learning algorithm not directly but in the reward function, giving the agent an intuition if it is able to make the model perform the desired gait effectively.

This should be noted that the choice of the observation space was carefully done based on actual observable properties by a normal human being. Specific data like the GRFs were not provided numerically to the model as those forces cannot be interpreted physically by us. All we know that a contact has been made and the impact of the contact on the joint moments.

\newpage

\section{Algorithms}
\label{sec:algorithms}

\subsection{Proximal Policy Optimization (PPO)}
\label{sec:algorithms_ppo}

Proximal policy optimization (PPO) \cite{c14} is a type  of policy gradient methods for RL with some modifications and has some of the benefits of trust region policy optimization (TRPO) \cite{c15}. They are more general and have better sample complexity in terms of empirical evaluations. Also, it performs very well for continuous control tasks.

As the main objective of PPO is to (approximately) maximize on each iteration it’s loss function:
\begin{equation}
L_t(\theta) = \hat{E}_t [L^{CLIP}_t (\theta) - c_1 L^{VF}_t (\theta) + c_2 S[\pi_\theta](s_t)]
\end{equation}
where $c_1$, $c_2$ are coefficients, $S$ denote entropy bonus to ensure sufficient
exploration, $L^{VF}_t$ is a squared-error loss of Value function $(V_\phi(s_t) - V^{target}_t)^2$ 
which is used to approximate the accumulated rewards, and $L^{CLIP}_t$ is Clipped Surrogate Objective where we try to maximize.

\begin{equation}
L^{CLIP}_t (\theta) = \hat{E}_t [\min(r_t(\theta) \hat{A}_t, clip(r_t(\theta), 1 - \epsilon, 1 + \epsilon) \hat{A}_t)]
\end{equation}
where $r_t(\theta)$ denotes the probability ratio $r_t(\theta) = \frac{\pi_\theta(a_t|s_t)}{{\pi_\theta}_{old}(a_t|s_t)}$.

$\hat{A}_t$ is an estimator of the advantage function at timestep $t$, and epsilon is a hyperparameter, say $\epsilon = 0.2$. The maximization of the first term inside the min in Eq. (11) would lead to an excessively large policy update, so we try to penalize changes to the policy by removing the incentive for moving $r_t$ outside of the interval $[1-\epsilon, 1+\epsilon]$.

\begin{algorithm}[thpb]
\SetAlgoLined
\KwResult{Trained Value Function $V$ and Optimal Policy ($\pi$)}
 Input: Initial Policy parameters $\theta_0$, value function parameters $\phi_0$\;
\For{k = 1, 2, \dots, max timesteps}{
    Collect State-Action trajectory $D_k = \{\tau_i\}$ using current policy $\pi_\theta = \pi(\theta_k)$ \;
    Calculate Reward-to-go, $\hat{R}_i = \sum_{(s_t, a_t, s_{t+1}) \in \tau_i} R(s_t, a_t, s_{t+1})$ $\forall \tau_i \in D_k$\;
    With current value function, $V_{\phi_k}$, calculate advantage estimates, $\hat{A}_t$ \;
    Update policy by maximizing PPO-Clip objective:
    \begin{equation*}
        \theta_{k+1} = \arg \max_{\theta} \frac{1}{|D_k|T} \sum_{\tau \in D_k} \sum^T_{t = 0} \min \left( \frac{\pi_\theta(a_t|s_t)}{\pi_{\theta_k}(a_t|s_t)} A^{\pi_{\theta_k}}(s_t, a_t), g(\epsilon, A^{\pi_{\theta_k}}(s_t, a_t)) \right)
    \end{equation*}
    where
    \begin{equation*}
        g(\epsilon, A) = 
        \begin{cases}
            (1+\epsilon)A & A \geq 0\\
            (1-\epsilon)A & A < 0
        \end{cases} 
    \end{equation*}\\
    Update Value function using:
    \begin{equation*}
        \phi_{k+1} = \arg \min_{\phi} \frac{1}{|D_k|T} \sum_{\tau \in D_k} \sum^T_{t = 0} (V_\phi(s_t) - \hat{R}_t)^2
    \end{equation*}
}
 \caption{Proximal Policy Optimization Algorithm}
\end{algorithm}

\subsection{Deep Deterministic Policy Gradient (DDPG)}
\label{sec:algorithms_ddpg}

Deep Deterministic Policy Gradient (DDPG) \cite{c16} is an actor-critic algorithm. The setting works as a model-free algorithm to understand the interactions between the environment and the model instead of just understanding the dynamics of the model. The overall framework is based on the deterministic policy gradient. The DDPG algorithm relies on finding the action maximizing the action-value function, thus it operates over continuous action spaces.

In standard notations, an agent’s behavior is characterized by a policy, $\pi$, which maps states to a probability distribution over the actions $\pi : S \longrightarrow P(A)$. However, on DDPG, the policy is deterministic as it is described by the name, which means that it outputs the actual actions instead of the probability selection of the action. Nevertheless, the deterministic policy is updated by a stochastic one; due to stochasticity, we try to bound the corresponding output actions in the environment limits.

Therefore, we have two actor and two critic networks, a target for each. Two pairs of them (actor-critic) will be on-policy, the other two will be off-policy. 

As actor function, $\mu(s|\theta^\mu)$, specifies the current policy by deterministically mapping states to a specific action. The critic $Q(s, a)$ is learned using the Bellman equation in Q-learning.

\begin{algorithm}[thpb]
\SetAlgoLined
\KwResult{Trained $Q$-Function and Optimal Policy ($\mu$)}
 Input: Initial Policy parameters $\theta$, $Q$-function parameters $\phi$, Replay Buffer $\mathcal{D}$\;
 Set Target parameters equal to main parameters $\theta_{targ} \gets \theta$, $\phi_{targ} \gets \phi$\;
 \Repeat{convergence}{
  Observe state $s$ and select action $a = \text{clip}(\mu_\theta(s) + \epsilon, a_{low}, a_{high})$, where $\epsilon \sim \mathcal{N}$ \;
  Execute $a$ in the Environment \;
  Observe next state $s'$, reward $r$ and done signal $d$ to check if $s'$ is terminal \;
  Store $(s, a, r, s', d)$ in replay buffer $\mathcal{D}$ \;
  \If{$s'$ is terminal}{
  Reset Environment state \;
  }
  \If{time to update}{
   	\For{ however many updates}{
   	Randomly sample a batch of transitions, $\mathcal{B} = \{(s, a, r, s', d)\}$ from $\mathcal{D}$\;
   	Compute Targets
   	\begin{gather*}
   	y(r, s', d) = r + \gamma (1-d) Q_{\phi_{targ}}(s', \mu_{\theta_{targ}}(s'))
   	\end{gather*} \\
   	Update $Q$-function by one step of gradient descent
   	\begin{gather*}
   	\nabla_\phi \frac{1}{|\mathcal{B}|} \sum_{(s, a, r, s', d) \in \mathcal{B}} (Q_\phi (s,a) - y(r, s', d))^2
   	\end{gather*} \\
   	Update policy by one step of gradient ascent
   	\begin{gather*}
   	\nabla_\theta \frac{1}{|\mathcal{B}|} \sum_{s \in \mathcal{B}} Q_\phi (s, \mu_\theta(s))
   	\end{gather*} \\
   	Update target networks
   	\begin{gather*}
   	\phi_{targ} \gets \rho \phi_{targ} + (1-\rho) \phi \\
   	\theta_{targ} \gets \rho \theta_{targ} + (1-\rho) \theta
   	\end{gather*}
   	}
   }
 }
 \caption{Deep Deterministic Policy Gradient Algorithm}
\end{algorithm}

The critic network will be
\begin{equation}
Q^\mu(s_t, a_t) = E_{r_t,s_{t+1} \sim E}[r(s_t, a_t) + \gamma Q^\mu (s_{t+1}, \mu(s_{t+1}))]
\end{equation}

A commonly used stochastic off-policy algorithm, is the greedy one
\begin{equation}
\mu(s) = arg max_a Q(s, a)
\end{equation}

The target actor will be doing the evaluation having an exploration policy $\mu'$ by adding noise sampled from a noise process N.

\begin{equation}
\mu'(s_t) = \mu(s_t|\theta^\mu_t) + N
\end{equation}

where N can be chosen to suit the environment actions space.

On each iteration the actions are chosen according to target actor network. As a result, on each iteration we try to minimize the following loss function:

\begin{equation}
L(\theta, Q) = E_{s_t \sim \rho^\beta,a_t \sim \beta,r_t \sim E}[(Q(s_t,a_t|\theta^Q) - y_t)^2]
\end{equation}

where $\beta$ is a stochastic policy, and $y_t = r(s_t, a_t) + \gamma Q(s_{t+1}, \mu(s_{t+1})|\theta^Q)$.

\newpage

\section{Reward Function Formulation}
\label{sec:algorithms_reward}

For this segment, the reward described in the Scalable Muscle-Actuated Human Simulation and Control (MASS) \cite{c11} was used with some modifications to get results on the OpenSim simulator using as imitation reference from the healthy gait data-set. Reward formulation:

\begin{equation}
R_{imitation} = r_q \times r_e \times r_c
\end{equation}

where $r_q$, $r_e$ and $r_c$ represent the pose imitation, end-effector imitation and center of mass (CoM) imitation respectively. The imitation rewards are intended to match the current and reference motions in terms of joint angles, end-effector positions, and CoM.

\begin{gather}
r_q = exp(-\sigma_q \sum_j \norm{\hat{q}_j (\phi) - q_j}^2) \\
r_e = exp(-\sigma_e \sum_e \norm{\hat{p}_e (\phi) - p_e}^2) \\
r_c = exp(-\sigma_c \norm{\hat{p}_{COM} (\phi) - p_{COM}}^2) 
\end{gather}

Here, all the individual rewards are normalized such that each of them $\in [0, 1]$, keeping the cumulative maximum achievable reward one as well. Furthermore, the hat symbols indicate desired values taken from the reference data, $j$ is the index of joints, and $e$ is the end-effectors index. The joint configurations are represented by joint angles. In our experiments, the weights are $\sigma_q = 10.0$, $\sigma_e = 20.0$ and $\sigma_c = 15.0$.

The multiplication gets rewards when both terms are rewarded, which makes sense because joint angles, end-effector positions, and movement of CoM are closely related to each other.

\subsection{Cost of Transport}

The cost of transport is defined as the energy spent in performing the locomotion gait. For torque actuated systems with motors, this is directly proportional to the work done by all the motors. A simple formulation for the cost of transport, $E$, for a single $i^{th}$ simulation time step is given by:

\begin{equation}
    E_i = \sum_{m \in \mathcal{T}} \tau^m_i \ (\theta^m_{i+1} - \theta^m_i) 
\end{equation}

where $\mathcal{T}$ is the actuator space for motor torques, $ \tau^m$ and $\theta^m$ denote the torque and angular position of the $m^{th}$ motor actuator.

\subsection{Metabolic Rate for MTU}

Metabolic rate can be defined as the cumulative energy consumption by all the muscle-tendon units. This is because for the MTU actuated systems, the energy obtained by metabolism is utilized by the muscles to do work. Hence, the intensity of work is directly proportional to food requirements and metabolic rate. The framework presented in this work considers the model proposed in \cite{wang2012}.

The rate of energy expenditure by a muscle is the sum of heat released and the mechanical work done by that muscle and is given by:
\begin{equation}
    \dot{E} = \dot{A} + \dot{M} + \dot{S} + \dot{W}
\end{equation}
where $\dot{A}$ is the activation heat rate of the muscle generated in response to excitation, $u$, and $\dot{M}$ is the maintenance heat rate to maintain the muscle at the current activation, $a$. The expressions for all the terms are as follows:

\begin{gather}
    \dot{A} = Mass_{muscle} \ f_A(u) \\
    \dot{M} = Mass_{muscle} \ g(l^{CE}/l^{opt}) \ f_M(a)
\end{gather}
where, $l^{CE}$ and $l^{opt}$ are the current fiber length and optimal fiber length respectively. The forms of the functions $f_A$, $f_M$ and $g$ as proposed by \cite{wang2012} are:
\begin{gather}
    f_A(u) = 40 \ \lambda \sin (\frac{\pi}{2} u) + 133 \ (1 - \lambda)(1 - \cos (\frac{\pi}{2} u)) \\
    f_M(a) = 74 \ \lambda \sin (\frac{\pi}{2} u) + 111 \ (1 - \lambda)(1 - \cos (\frac{\pi}{2} u))
\end{gather}
and $\lambda$ denote the fraction of Type I fibers in a given muscle.

Finally, the other two terms in Eq.~(21) is given by:
\begin{gather}
    \dot{S} = \frac{1}{4} \ F^{MTU} \ \{-v^{CE}\}_{+} \\
    \dot{W} = \ F^{CE} \ \{-v^{CE}\}_{+} 
\end{gather}
modelling the muscle shortening heat rate and the mechanical work rate by the active element. Here, $F^{MTU}$ is the net muscle force composed of both the active and passive forces, whereas $F^{CE}$ represents only the active force. The final effort for the current time step is given by:

\begin{equation}
    E_i = \sum_{m \in \mathcal{M}} \dot{E}_m
\end{equation}

\newpage

\chapter{Results}
\label{sec:results}

This section includes the Actor and Critic Neural Networks and their properties, the training rewards, and network losses and finally concludes with the comparison between the learned and the desired walking motion.

\section{Training Hyper-parameters}
\label{sec:results_hyperparams}

RL algorithms, or in general, machine learning (ML) algorithms, involve significant number of hyperparameters which have to be determined to get the desired behavior. While the direct, first-level model parameters, which are determined during training, the upper-level tuning parameters have to be carefully optimized to achieve maximal performance. In order to converge to an appropriate hyperparameter configuration for a specific task at hand, users of RL algorithms can resort to default values of hyperparameters that are specified in implementing software packages or manually configure them, for example, based on recommendations from the literature, experience or trial-and-error. Thus, in the presented setting, separate trainings were conducted for DDPG and PPO to converge to specific hyperparameters by trial-and-error method. Those hyperparameters are respectively given in Table \ref{ddpgparam} and \ref{ppoparam} below. A more detailed discussion for the MTU actuation case is discussed in Section~\ref{sec:tuning} of this chapeter.

\begin{longtable}{|c|c|c|}
\hline
\textbf{Hyperparameter} & \textbf{For Torque Actuation} & \textbf{For MTU Actuation}\\
\hline
Policy for training &  \multicolumn{2}{c|}{Multi-Layer Perceptron (MLP)}\\
\hline
Hidden Nodes for actor & \multicolumn{2}{c|}{512, 512} \\
\hline
Hidden Nodes for critic & \multicolumn{2}{c|}{512, 512}\\
\hline
Activation function & \multicolumn{2}{c|}{Hyperbolic Tan}\\
\hline
Discount factor (gamma, $\gamma$)  & 0.99 & 0.995\\
\hline
learning rate for actor &  0.0003  & 0.0004\\
\hline
learning rate for critic &  0.003 &  0.004\\
\hline
GAE lambda ($\lambda$) &  \multicolumn{2}{c|}{0.95}\\
\hline
Batch size &  \multicolumn{2}{c|}{128}\\
\hline
action noise & \multicolumn{2}{c|}{OrnsteinUhlenbeckActionNoise} \\
\hline
soft update coefficient (tau, $\tau$) & \multicolumn{2}{c|}{0.001}\\
\hline
Replay Buffer size & \multicolumn{2}{c|}{50000}\\
\hline

\caption{Final Hyperparameters for DDPG Algorithm}
\label{ddpgparam}
\end{longtable}

\begin{longtable}{|c|c|}
\hline
\textbf{Hyperparameter} &  \textbf{For MTU Actuation}\\
\hline
Policy for training &  Multi-Layer Perceptron (MLP)\\
\hline
Hidden Nodes for Policy & 512, 512 \\
\hline
Activation function & Hyperbolic Tan\\
\hline
Discount factor (gamma, $\gamma$)  & 0.995\\
\hline
learning rate for policy &  0.0003\\
\hline
learning rate for value function &  0.001\\
\hline
Clip ratio ($\epsilon$) &  0.2\\
\hline
\caption{Final Hyperparameters for PPO Algorithm}
\label{ppoparam}
\end{longtable}

\section{Training Results for Torque Actuated Model}
\label{sec:results_torque}

\begin{figure}[ht]
\begin{subfigure}{.5\textwidth}
  \centering
  \includegraphics[width=0.9\linewidth]{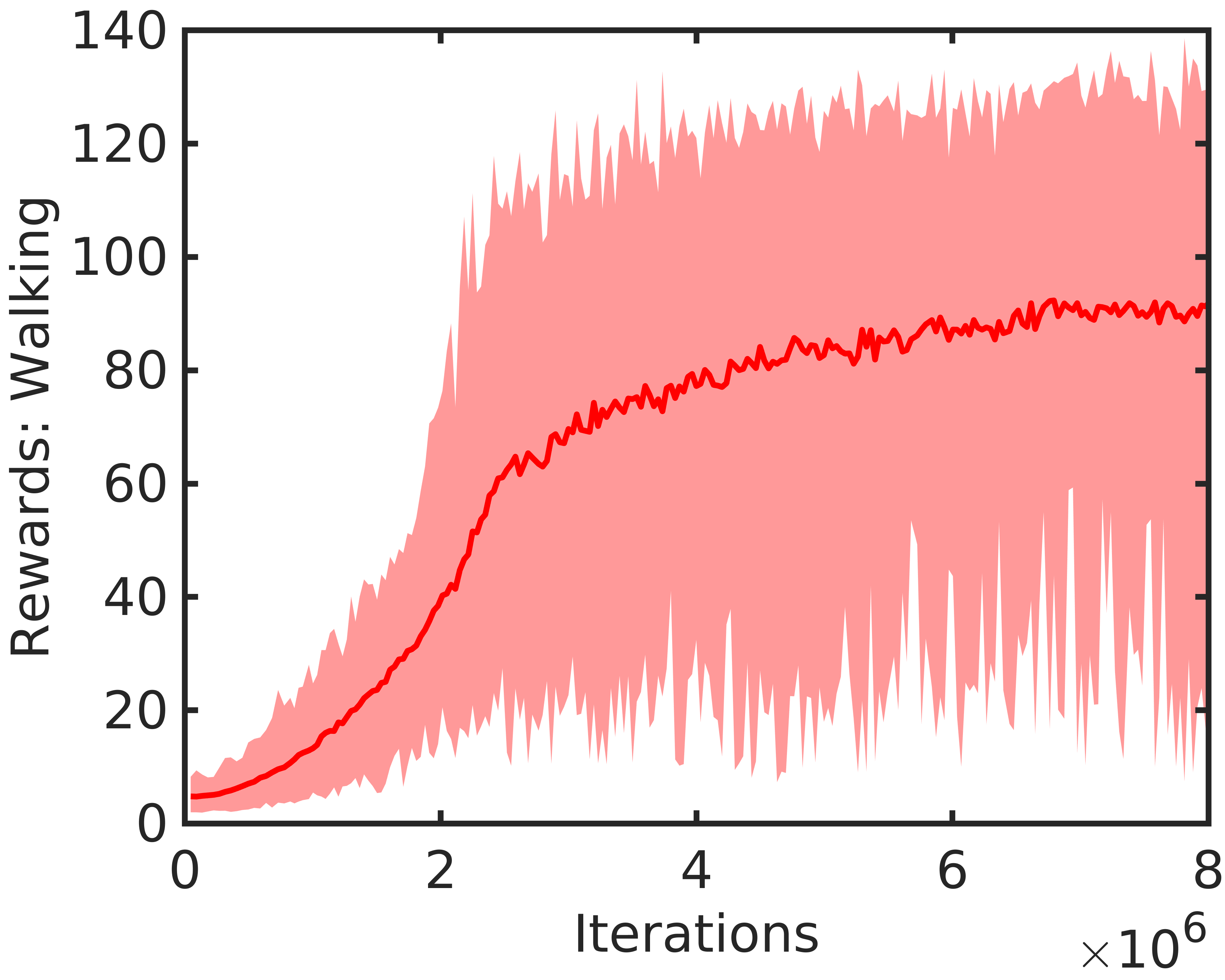}
  \caption{PPO Reward Curve}
  \label{fig:sfig11}
\end{subfigure}%
\begin{subfigure}{.5\textwidth}
  \centering
  \includegraphics[width=0.9\linewidth]{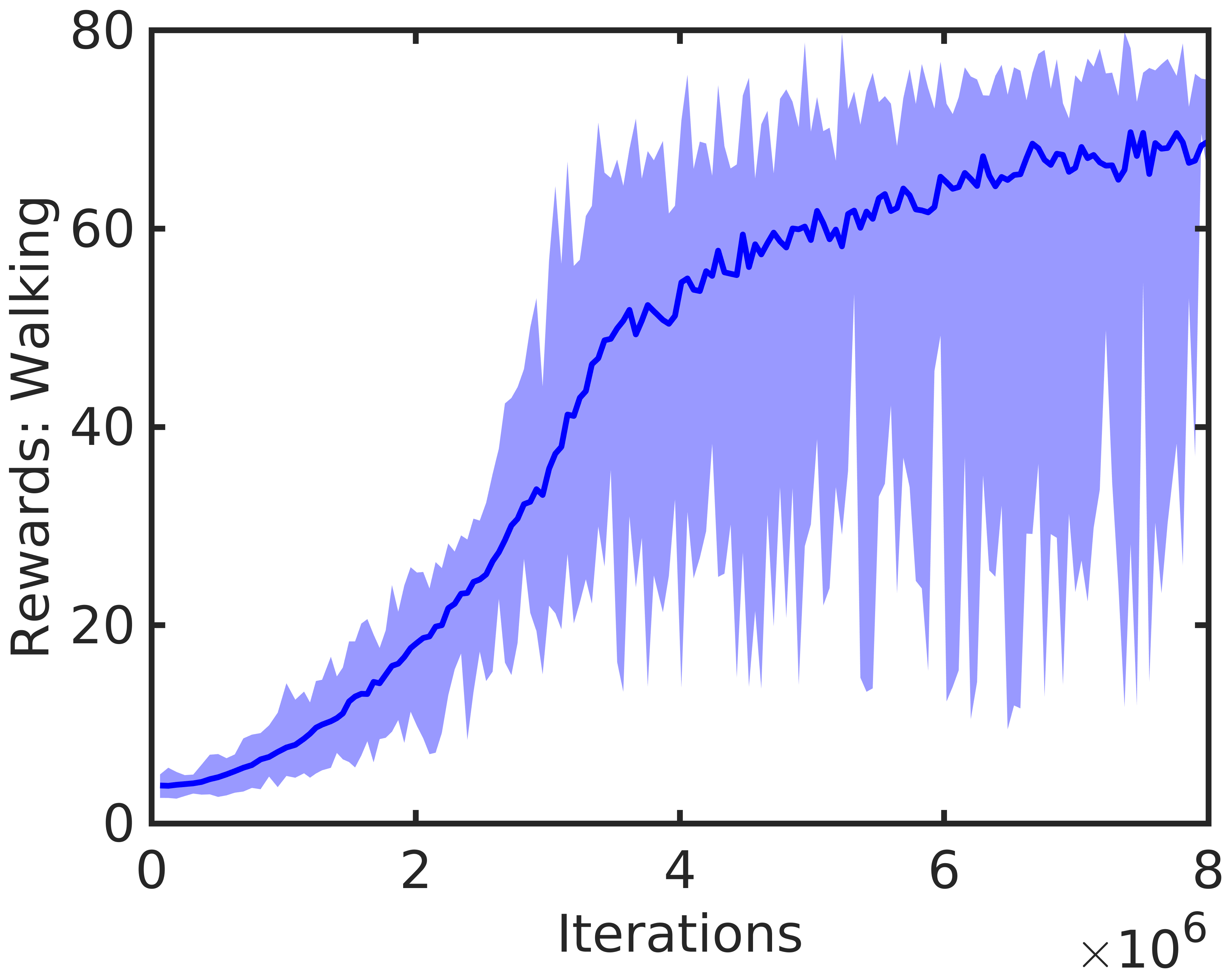}
  \caption{DDPG Reward Curve}
  \label{fig:sfig223}
\end{subfigure}
\caption{Rl Algorithm Reward evolution in walking experiment when trained for 8M timesteps (iterations)}
\label{fig:reward}
\end{figure}

The training progress is basically determined by the increment in the rewards obtained by the agents. The reward curves obtained from both the algorithms PPO and DDPG are shown in Fig. \ref{fig:reward}. PPO attains a maximum reward of 130 with a mean around 80 whereas DDPG attain a maximum reward of 78 with a mean around 70. The variance of rewards for DDPG is less due to its deterministic nature as compared to the stochastic policy given by PPO. Both the algorithms are trained for eight million timesteps. The comparison results with the reference motion are shown in the next section.

   
The actor and critic losses also play a significant role. Although a decrease in a loss might suggest that the actor is becoming deterministic precisely, RL problems might not always guarantee that the agent is progressing towards a better performance. 


\subsection{Compared with Reference Walking Gait}

The reference motion consists of motion capture data of walking gait for 1000 timesteps. Given that, to test the actual capabilities of imitation learning, only two human walking step data were provided (nearly 130 timesteps), and then the agent was left to generalize further. In the process, the agent learns the gait and generalizes it to walk infinitely without tipping or falling. The significant learnings from the imitation setup were that the agent learns to use the gait phase provided in the observation dictionary and learns symmetry accordingly.
   
\begin{figure}[t!]
      \centering
      \includegraphics[scale=0.15]{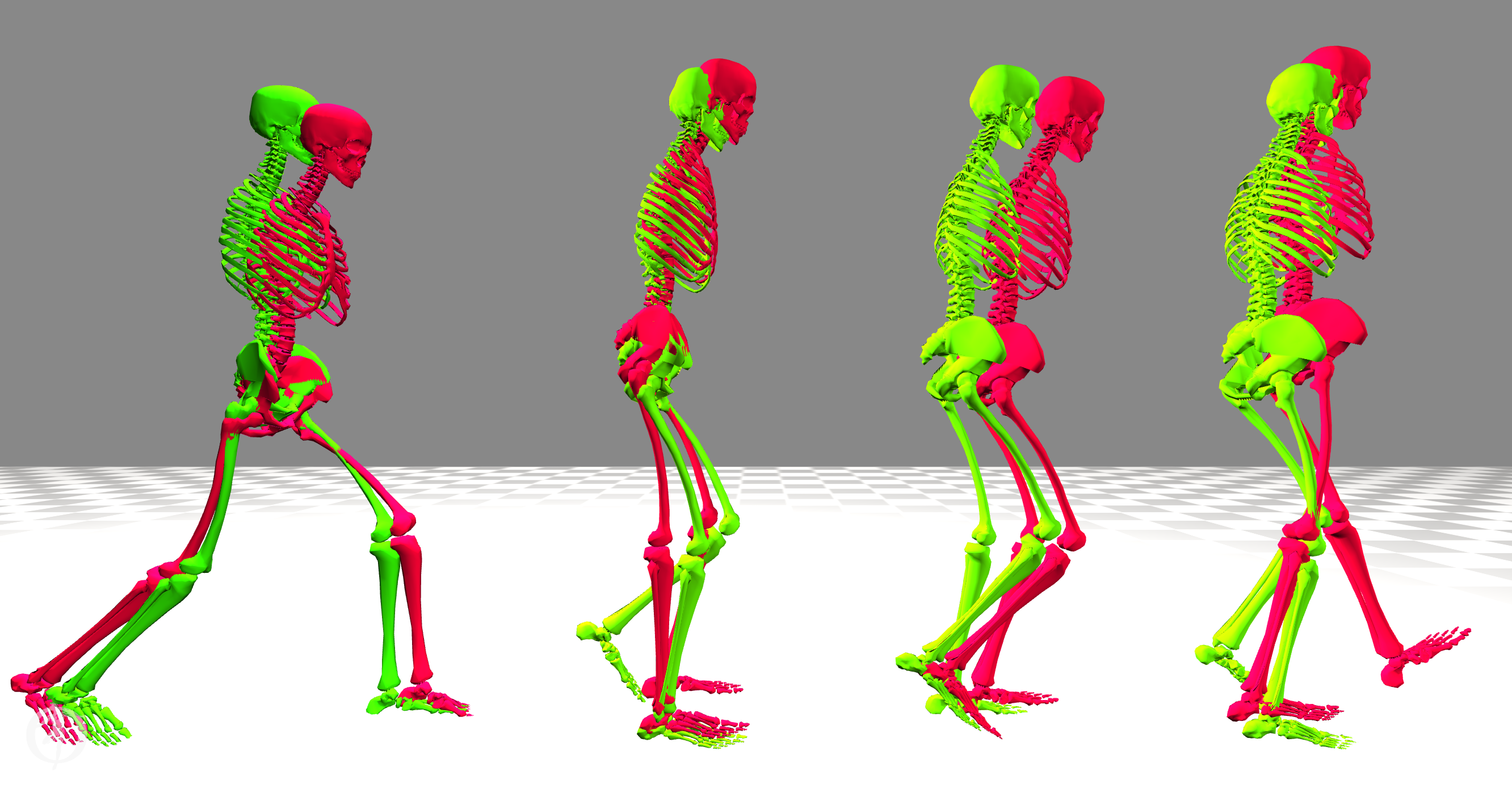}
      \caption{Comparison between Desired (Green) and Learned (Red) walking gait using the DDPG RL Algorithm}
      \label{fig:compare}
   \end{figure}
   
Fig. \ref{fig:compare} depicts the quality of imitation for the first two steps, i.e., the first two instances shown. For the next two instances, for which the agent was not trained, the model takes larger foot-steps. In that case, although the quality and nature of the gait remains identical to the reference, the difference between the reference and the trained motion becomes significant for the same time steps.

\begin{figure}
\centering
\begin{subfigure}{.32\textwidth}
  \centering
  \includegraphics[width=\linewidth]{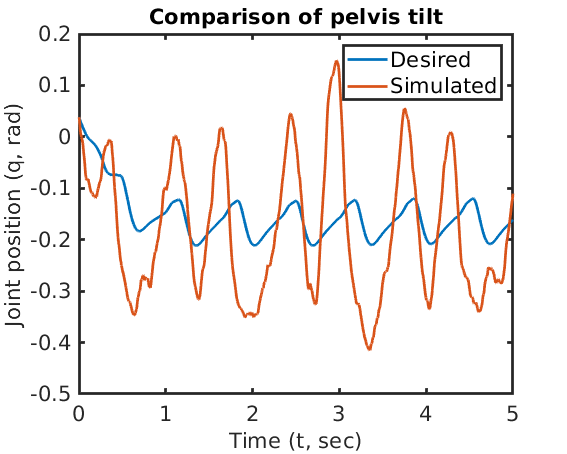}
  \caption{Pelvis Tilt Angle}
  \label{fig:sfig13}
\end{subfigure}%
\begin{subfigure}{.32\textwidth}
  \centering
  \includegraphics[width=\linewidth]{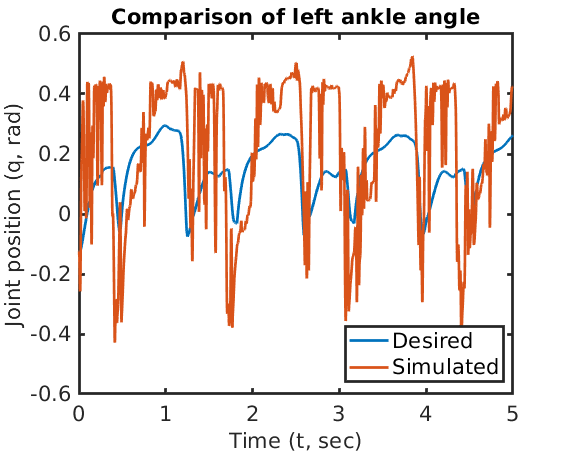}
  \caption{Left Ankle Angle}
  \label{fig:sfig225}
\end{subfigure}
\begin{subfigure}{.32\textwidth}
  \centering
  \includegraphics[width=\linewidth]{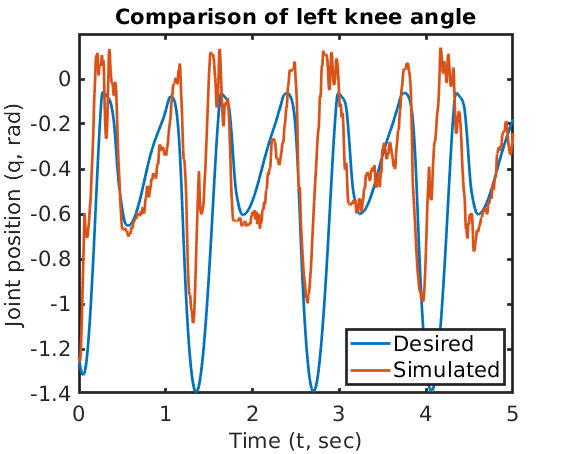}
  \caption{Left Knee Angle}
  \label{fig:sfig14}
\end{subfigure}%
\begin{subfigure}{.32\textwidth}
  \centering
  \includegraphics[width=\linewidth]{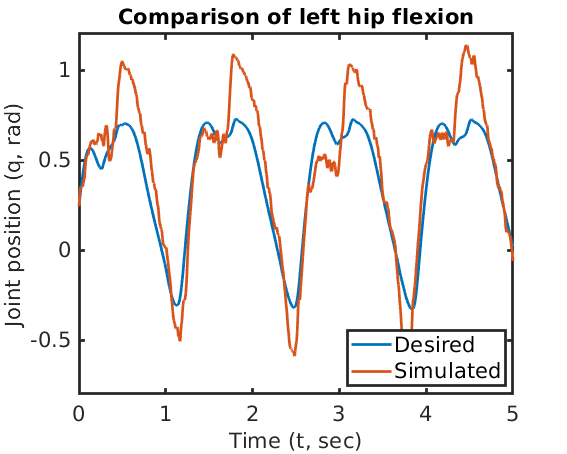}
  \caption{Left Hip Flexion}
  \label{fig:sfig226}
\end{subfigure}
\caption{Comparison of Desired and Learned Gait Patterns. The generalization between states can be easily verified based on the limit cycles learned for each of the coordinates.}
\label{fig:comparison}
\end{figure}

\begin{figure}
\centering
\begin{subfigure}{.32\textwidth}
  \centering
  \includegraphics[width=\linewidth]{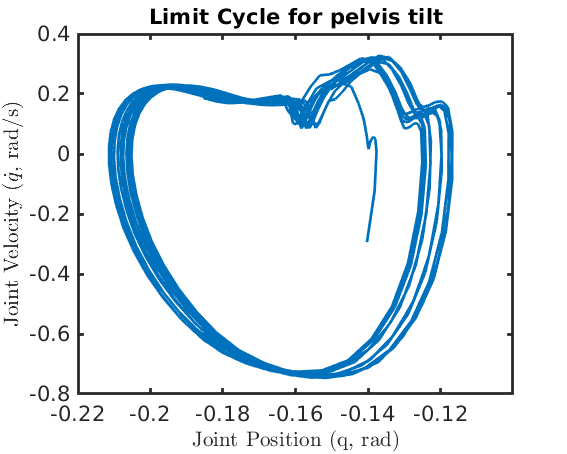}
  \caption{Pelvis Tilt Angle}
  \label{fig:sfig15}
\end{subfigure}%
\begin{subfigure}{.32\textwidth}
  \centering
  \includegraphics[width=\linewidth]{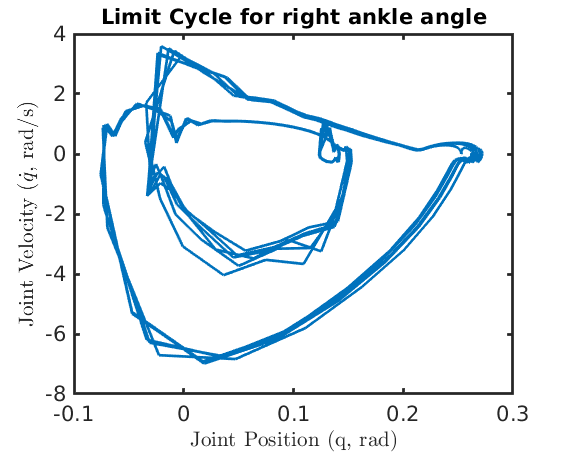}
  \caption{Right Ankle Angle}
  \label{fig:sfig227}
\end{subfigure}
\begin{subfigure}{.32\textwidth}
  \centering
  \includegraphics[width=\linewidth]{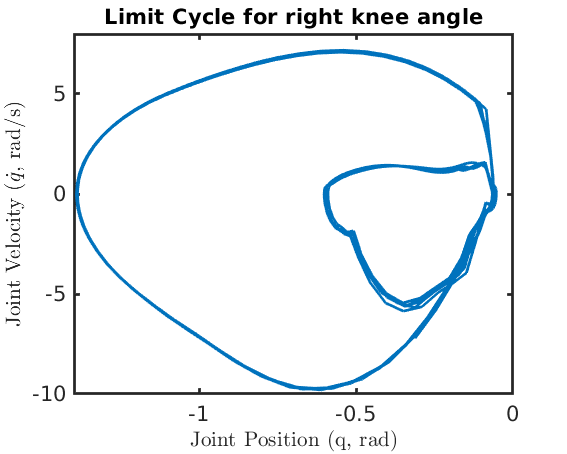}
  \caption{Right Knee Angle}
  \label{fig:sfig16}
\end{subfigure}%
\begin{subfigure}{.32\textwidth}
  \centering
  \includegraphics[width=\linewidth]{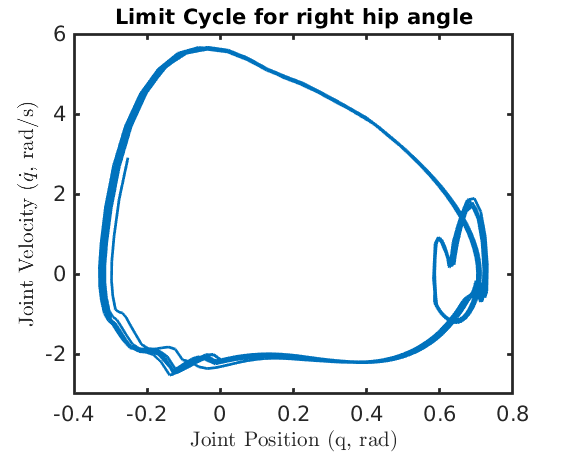}
  \caption{Right Hip Flexion}
  \label{fig:sfig228}
\end{subfigure}
\caption{Limit Cycles/ Phase Plots for the Learned Gait drawn between the joint position and joint velocities. Results are symmetrical.}
\label{fig:lcycles}
\end{figure}

The following figure \ref{fig:comparison} shows the tracking results for the pelvis tilt and the left hip, knee, and ankle. The data clearly concludes that the agent successfully learns the symmetry associated between the left and right Cody segments and the similarity associated between gait phases. The learned agent generates phase/ limit cycles, which corresponds to the learned walking behavior. The limit cycles are shown in figure \ref{fig:lcycles}.

For video illustrations, kindly refer \footnote{\href{https://utkarshmishra04.github.io/videos/BTP2021/comparison_walking_motor.mp4}{utkarshmishra04.github.io/videos/BTP2021/comparison\_walking\_torque.mp4}} link.

\newpage

\subsection{Additional Gaits Formulated}

Human gaits have a wide range of diversity, and the presented formulation should hold true for such a wide range of gaits. Some elementary forms of gaits that children generally learn during their early stages of learning are walking, running, and jumping. However, this should be noted that although we aim towards imitating healthy gaits, the people performing such gaits might not be healthy themselves. An example of this can be a person with a locked knee (which cannot be folded) trying to learn how to walk from observing a healthy person. Such type of motions trials were conducted and observed. As illustrated in Fig. \ref{fig:addgaits}, four more varieties of experiments were conducted in addition to the walking gait. The gaits and their descriptions are as follows:

\begin{figure}[t!]
      \centering
      \includegraphics[scale=0.12]{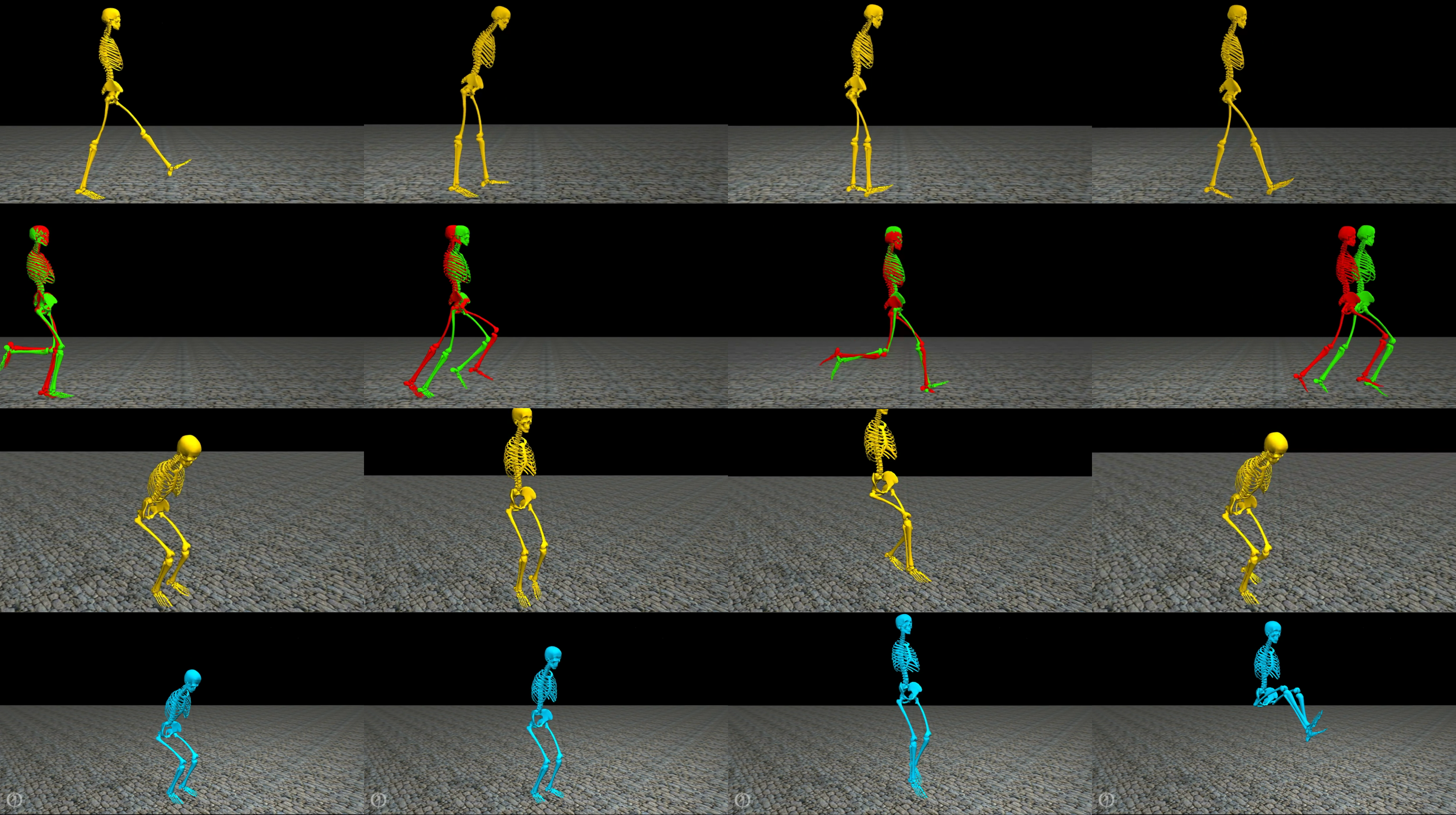}
      \caption{Additional Gaits implemented and imitated using the torque actuation framework, namely (from top to bottom): Prosthetic Walking (with a Locked Knee), Running, Continuous Jumping and High Jump.}
      \label{fig:addgaits}
   \end{figure}

\begin{itemize}
    \item \textbf{Prosthetic Walking:} This simulation study was performed on a modified human model representing a locked knee on one of the legs. The imitation reference data was the same as that of healthy walking. The RL policy formulated a strategy to walk with almost the same horizontal speed as the reference data. This study gives us insights into how people with physiological difficulties can also learn by observing healthy individuals. 
    
    \item \textbf{Running:} Running gait was performed to get an intuition on how the inclusion of phase variable into observations and relative change in phase of walking and running helps to realize periodic gaits. 
    
    \item \textbf{High Jump and Continuous Jumping:} High Jump was simulated in SCONE, and the predictive simulation gait was obtained. While the phase variable efficiently dealt with the jumping task, the landing was not learned, and hence to get a good landing behavior, an extra objective was added to balance itself while landing. This led to continuous jumping behavior.
\end{itemize}

Another variant of environment used with RL framework discussed in the project is discussed in Appendix \ref{appendix:A}. Related video links are present in Appendix \ref{appendix:B}.

The motivations behind the addition of additional gait analysis is the biological conclusion that it is very difficult to predict the functional outcome of a clinical treatment in patients with cerebral palsy (CP), which often leads to follow-up treatments. Therefore, this formulations aims to create a modular framework that will enable the clinicians to compare the effect of different treatments by RL predictions and to determine which treatment has the highest potential of improving gait performance before the intervention. This is eventually expected to increase the quality of clinically relevant research.
   
\newpage

\section{Training Results for MTU Actuated Model}
\label{sec:results_muscle}

An almost similar dynamics govern the MTU actuated model as that of the motor actuated model except for the fact that the joint moments are the torques applied by the muscles by exerting forces at their contact positions.

\begin{figure}[ht]
      \centering
      \includegraphics[width=\linewidth]{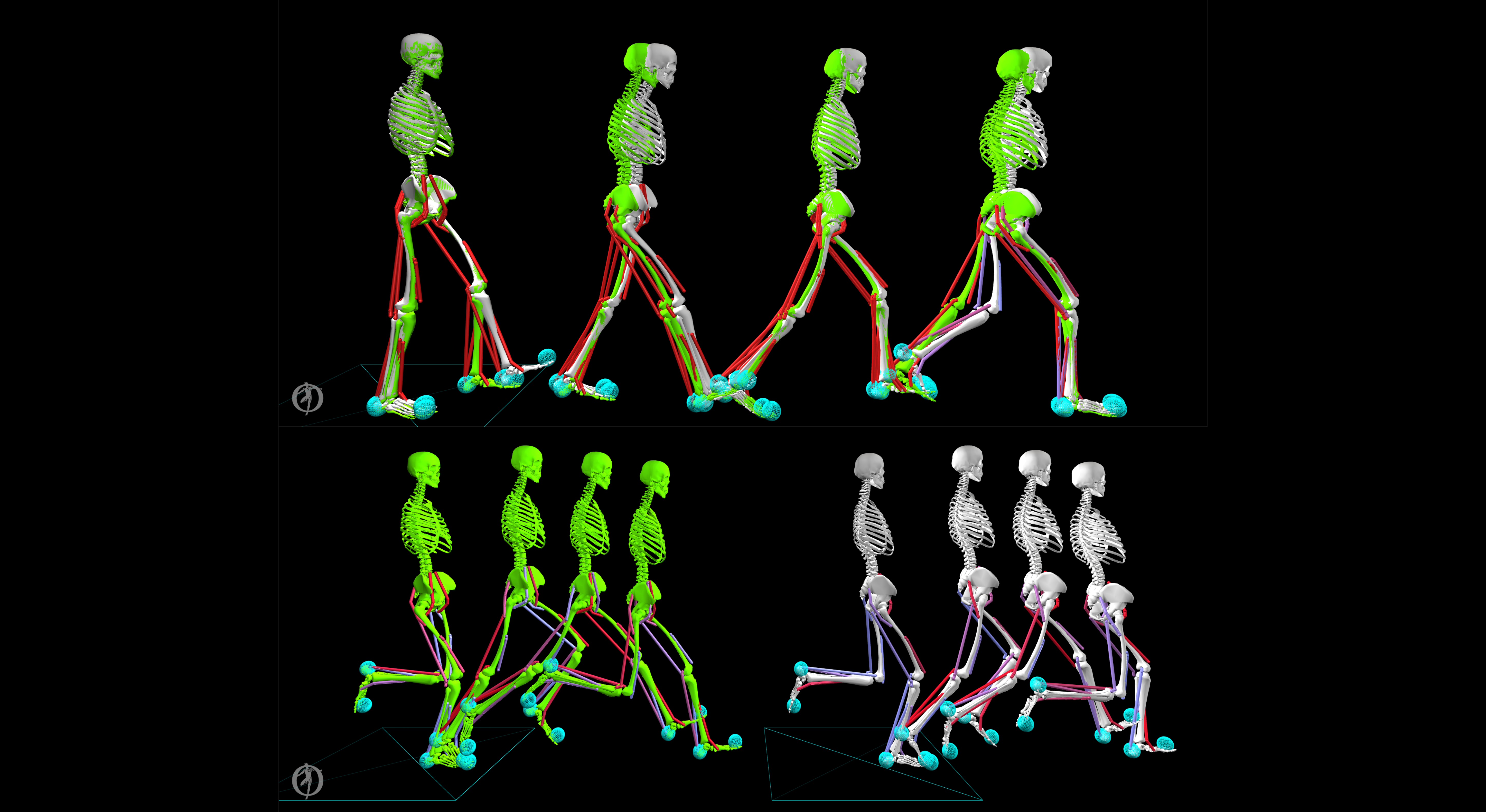}
      \caption{MTU Actuated walking and running gait (Grey Model) using the PPO RL Algorithm compared with the desired healthy gait (Green Model). Muscles are colored proportional to their activation. 0 activation corresponds to blue and 1 activation corresponds to red.}
      \label{fig:muswalk}
   \end{figure}
   
The walking movement obtained using MTU actuated model, as shown in Fig. \ref{fig:muswalk}, was analyzed based on the learned phases of the joints and muscle activation pattern. Finally, a comparison is drawn between the joint moments from training using torque actuated and MTU actuated models based on the same reference kinematic data.

\subsection{Comparison for Walking Reference Gait: Phase Cycles and Muscle Activation}

\begin{figure}
\centering
\begin{subfigure}{.35\textwidth}
  \centering
  \includegraphics[width=0.9\linewidth]{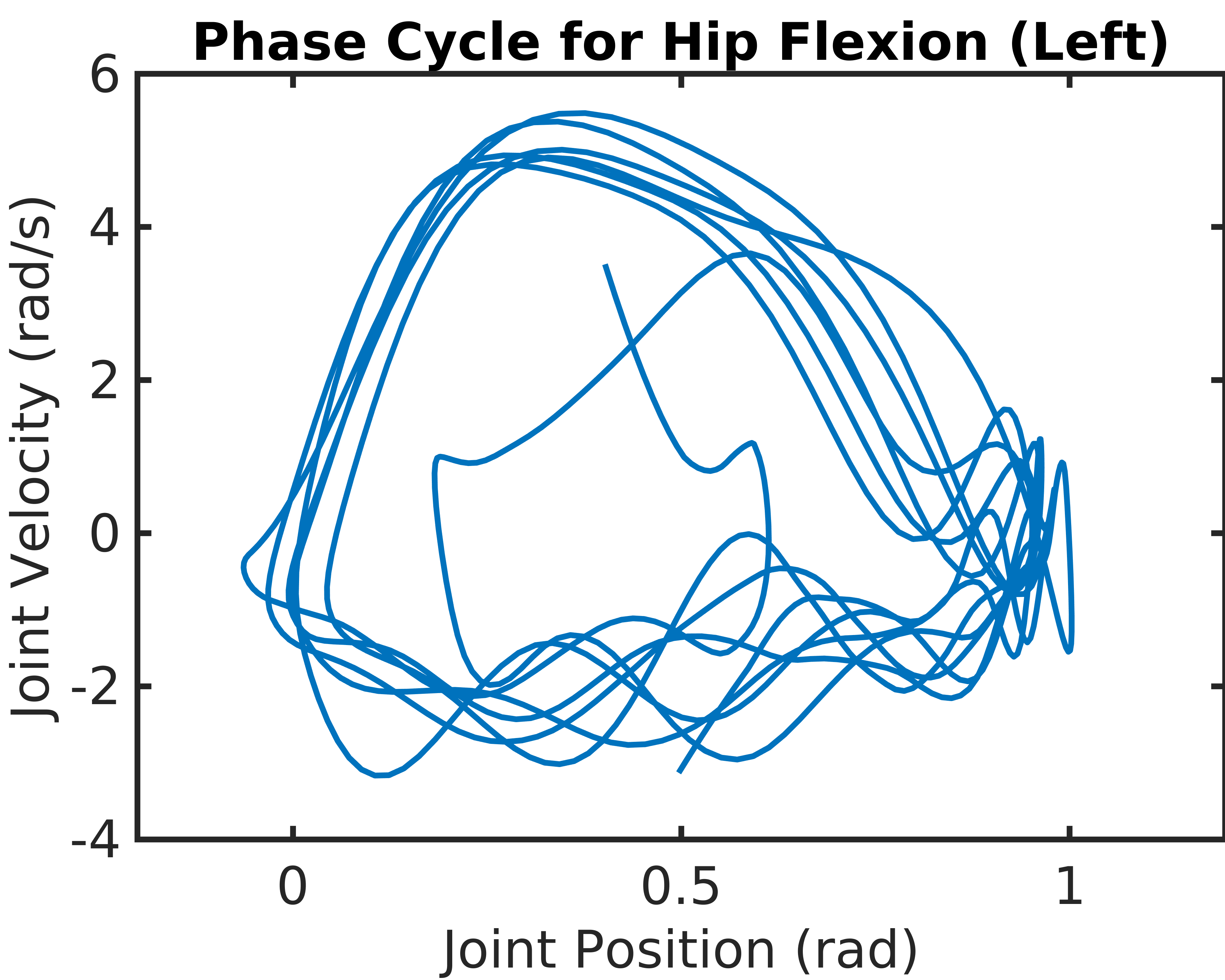}
  \caption{Left Hip Flexion}
  \label{fig:sfig17}
\end{subfigure}%
\begin{subfigure}{.35\textwidth}
  \centering
  \includegraphics[width=0.9\linewidth]{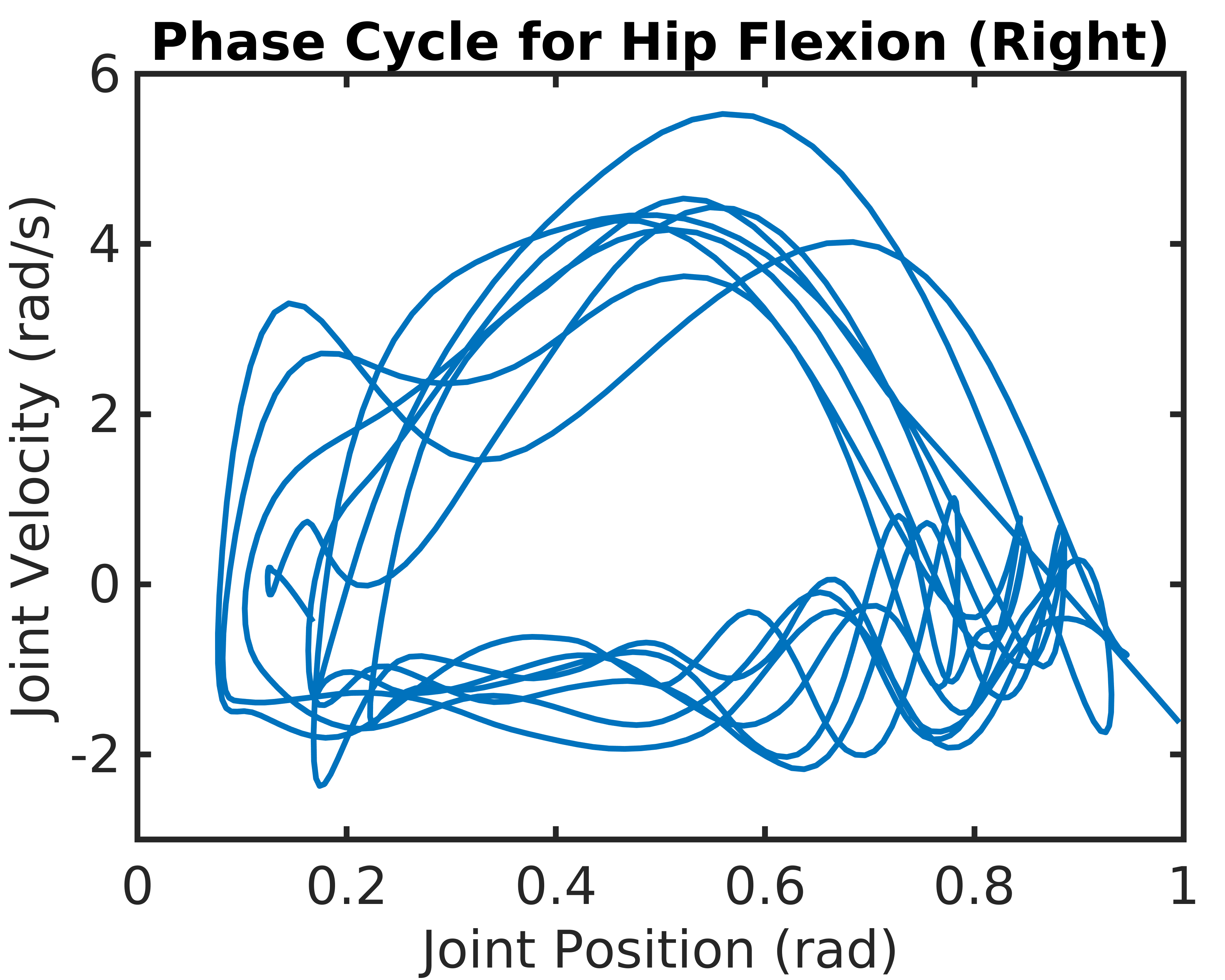}
  \caption{Right Hip Flexion}
  \label{fig:sfig229}
\end{subfigure}
\begin{subfigure}{.35\textwidth}
  \centering
  \includegraphics[width=0.9\linewidth]{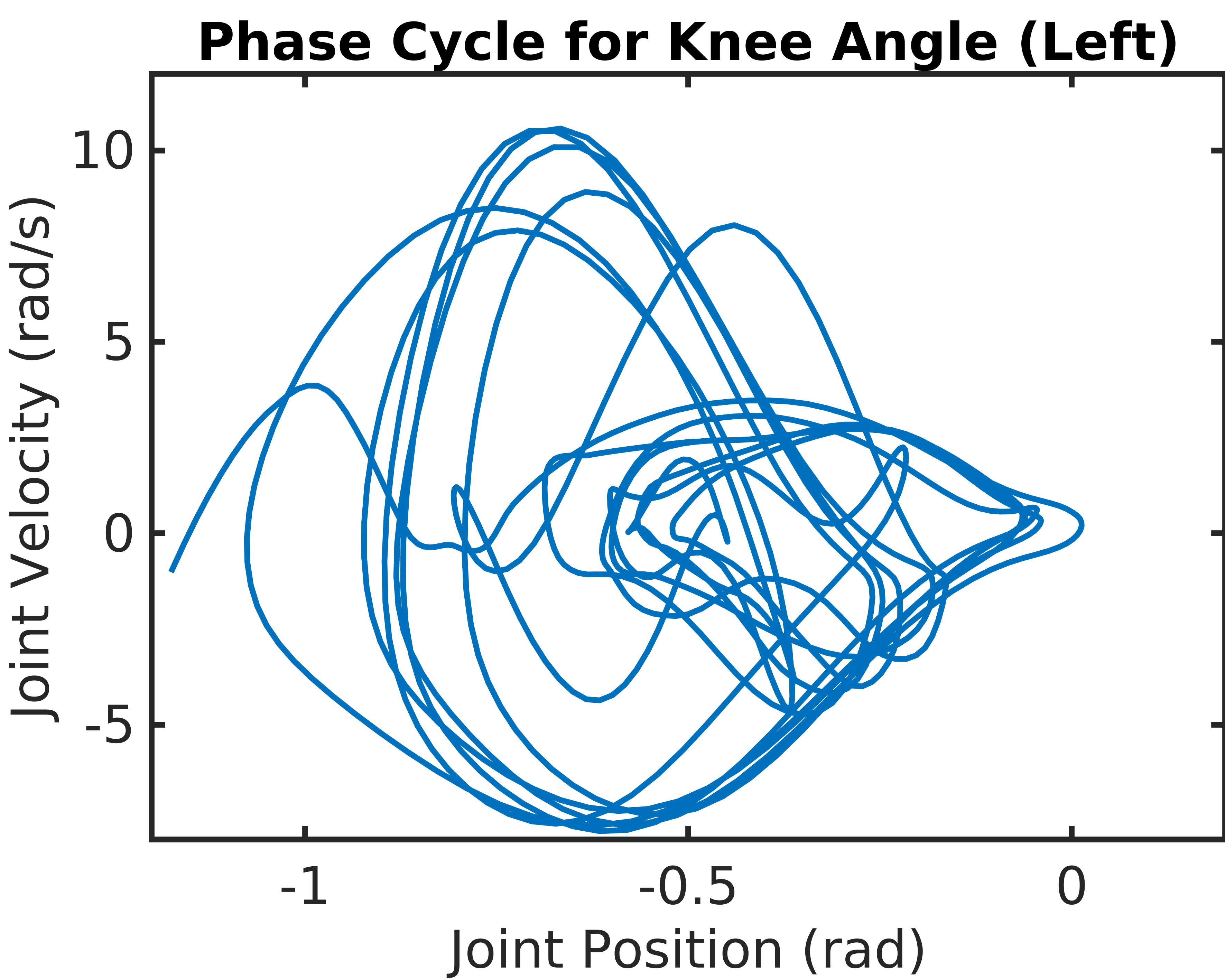}
  \caption{Left Knee Angle}
  \label{fig:sfig18}
\end{subfigure}%
\begin{subfigure}{.35\textwidth}
  \centering
  \includegraphics[width=0.9\linewidth]{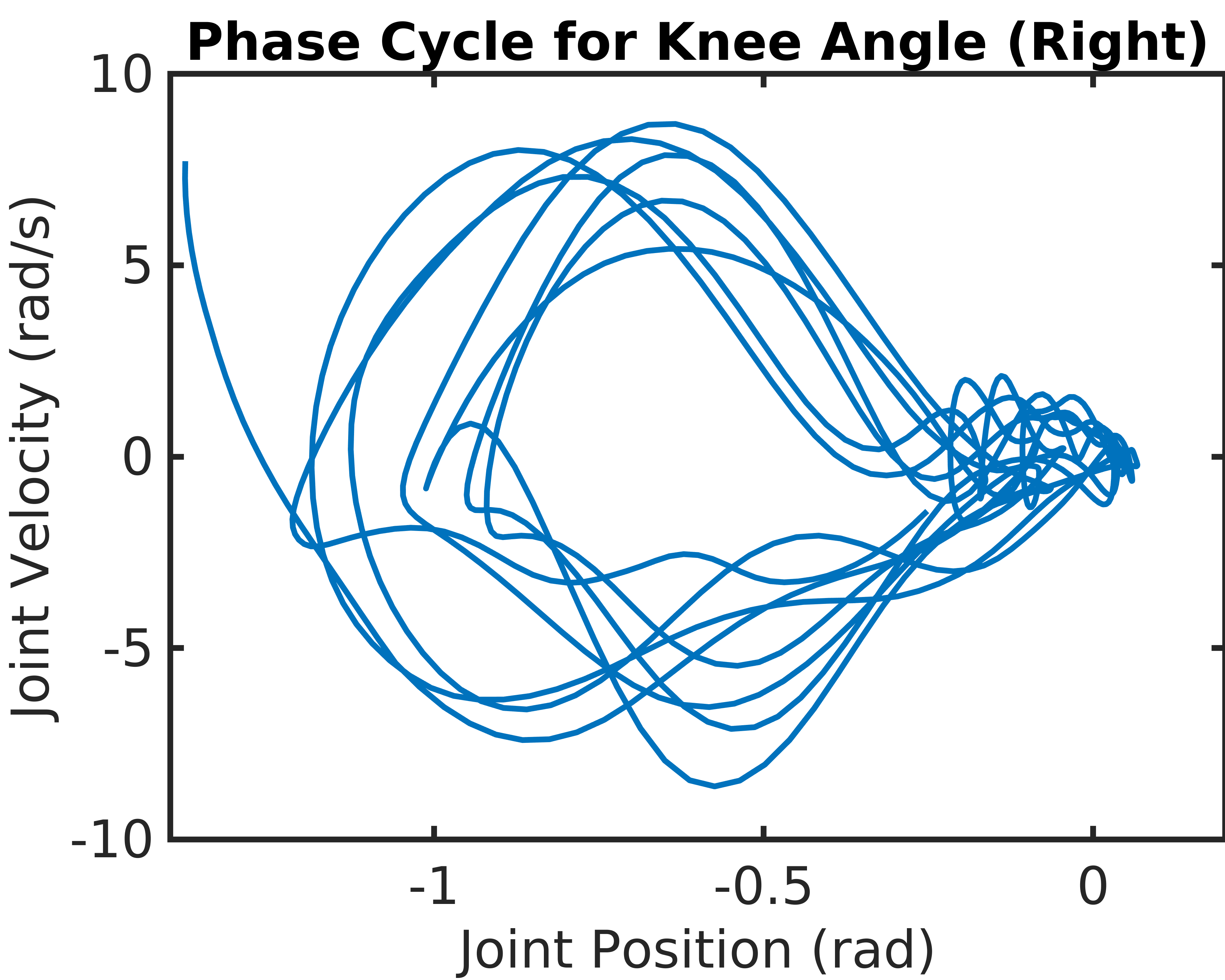}
  \caption{Right Knee Angle}
  \label{fig:sfig230}
\end{subfigure}
\caption{Learned Gait Patterns and phase cycles for MTU actuated walking. The generalization and symmetry between states can be realized.}
\label{fig:phasemus}
\end{figure}

\begin{figure}
\centering
\begin{subfigure}{.35\textwidth}
  \centering
  \includegraphics[width=0.9\linewidth]{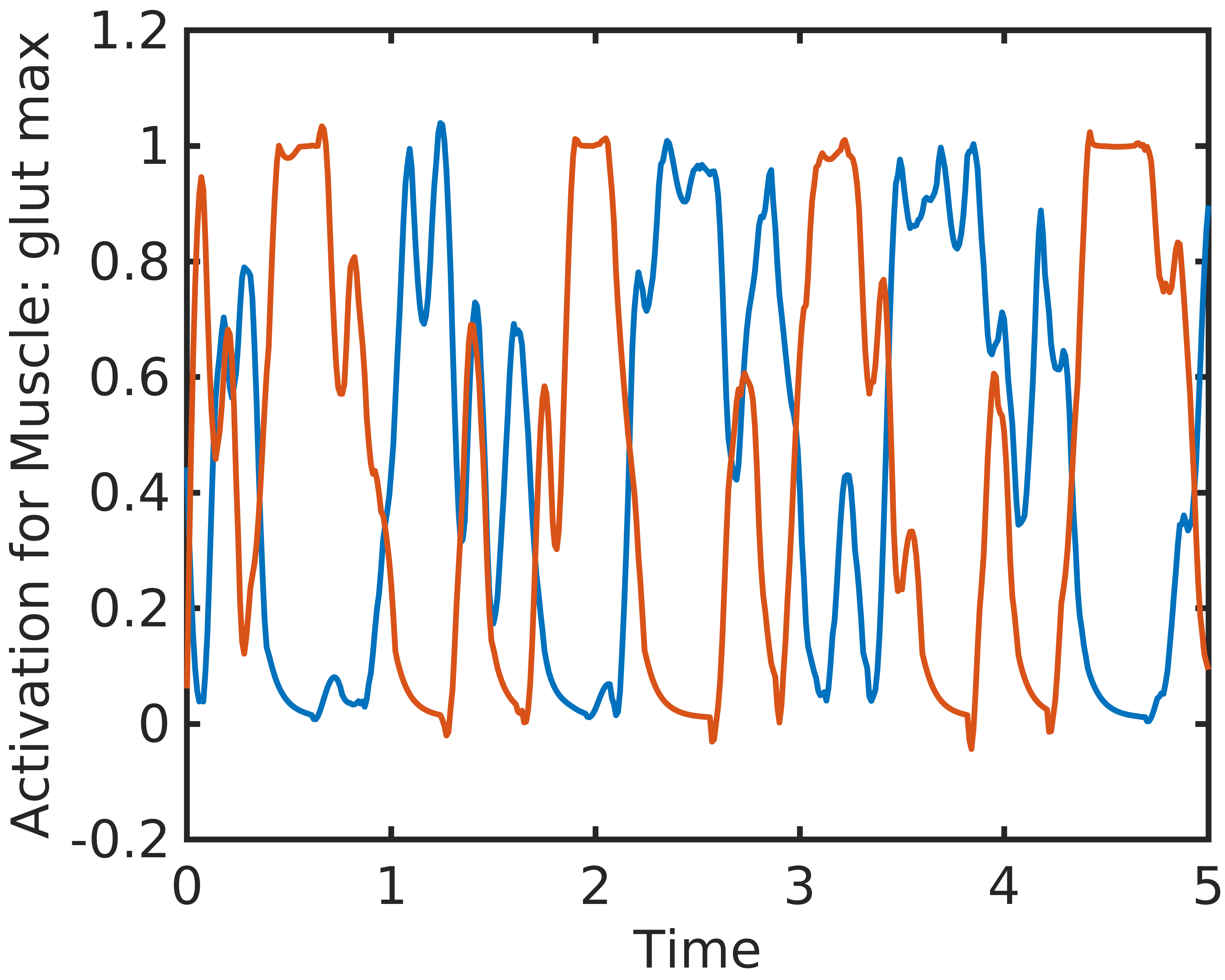}
  \caption{Gluteus Maximus}
  \label{fig:sfig19}
\end{subfigure}%
\begin{subfigure}{.35\textwidth}
  \centering
  \includegraphics[width=0.9\linewidth]{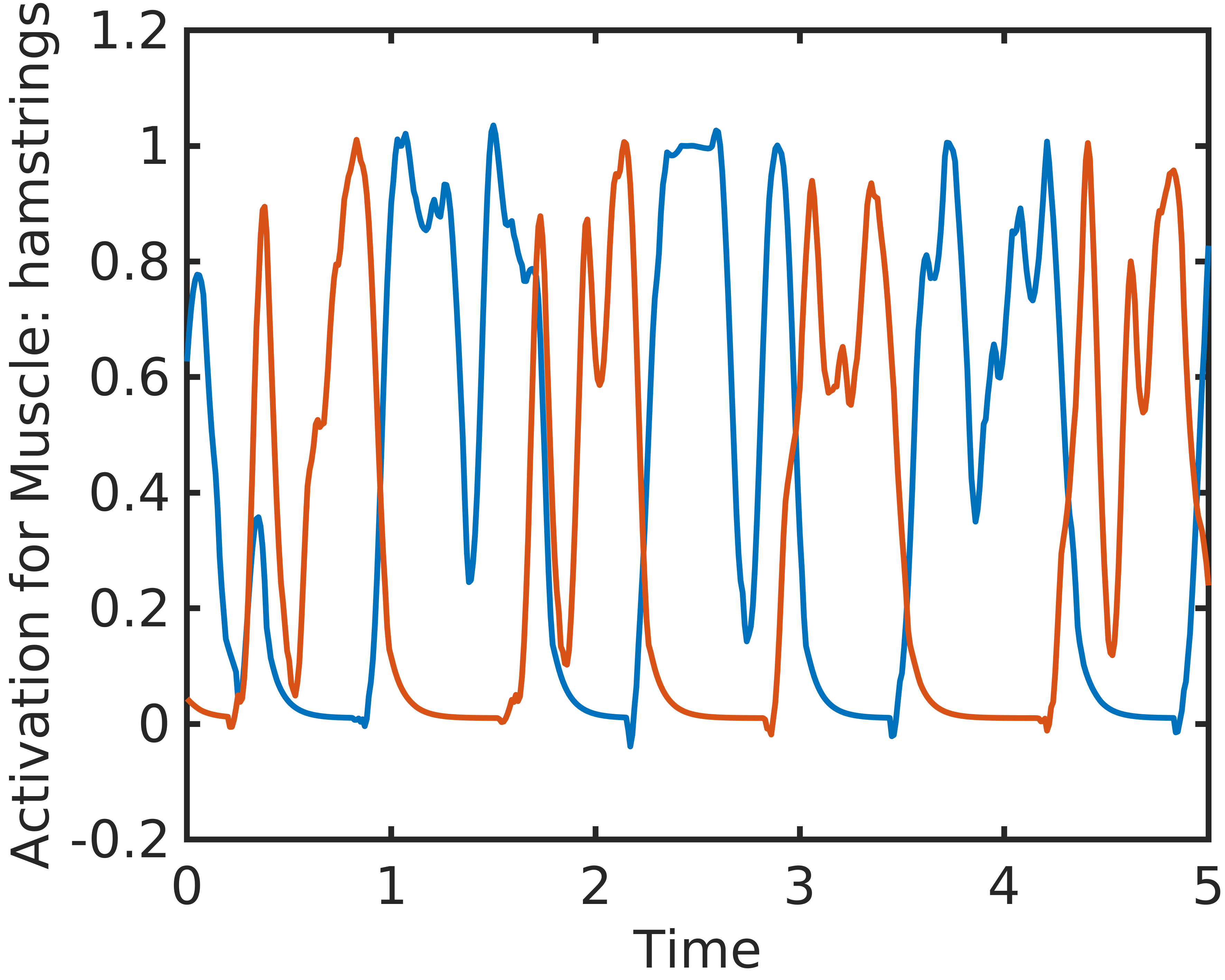}
  \caption{Hamstring}
  \label{fig:sfig231}
\end{subfigure}
\begin{subfigure}{.35\textwidth}
  \centering
  \includegraphics[width=0.9\linewidth]{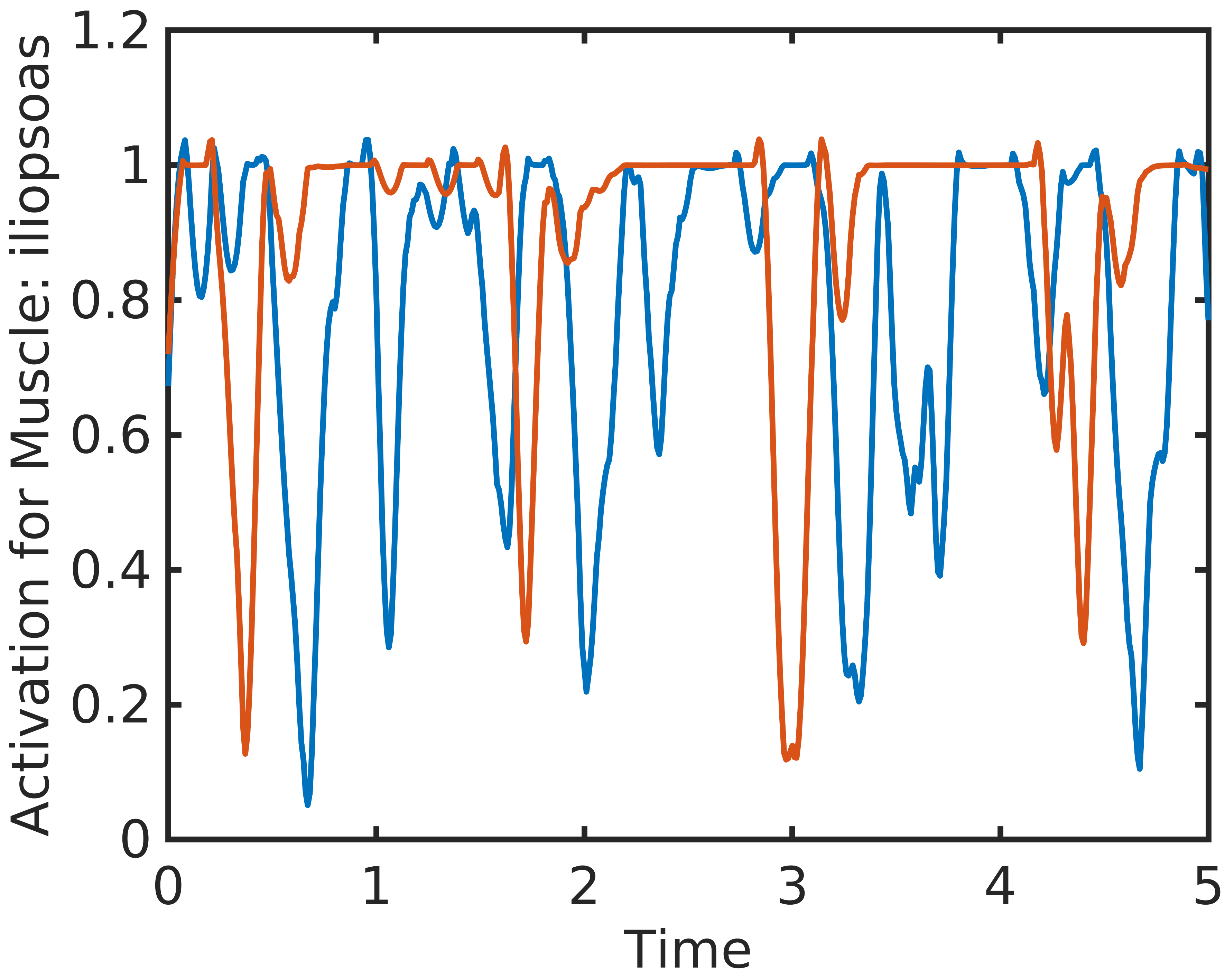}
  \caption{Iliopsoas}
  \label{fig:sfig2320}
\end{subfigure}%
\begin{subfigure}{.35\textwidth}
  \centering
  \includegraphics[width=0.9\linewidth]{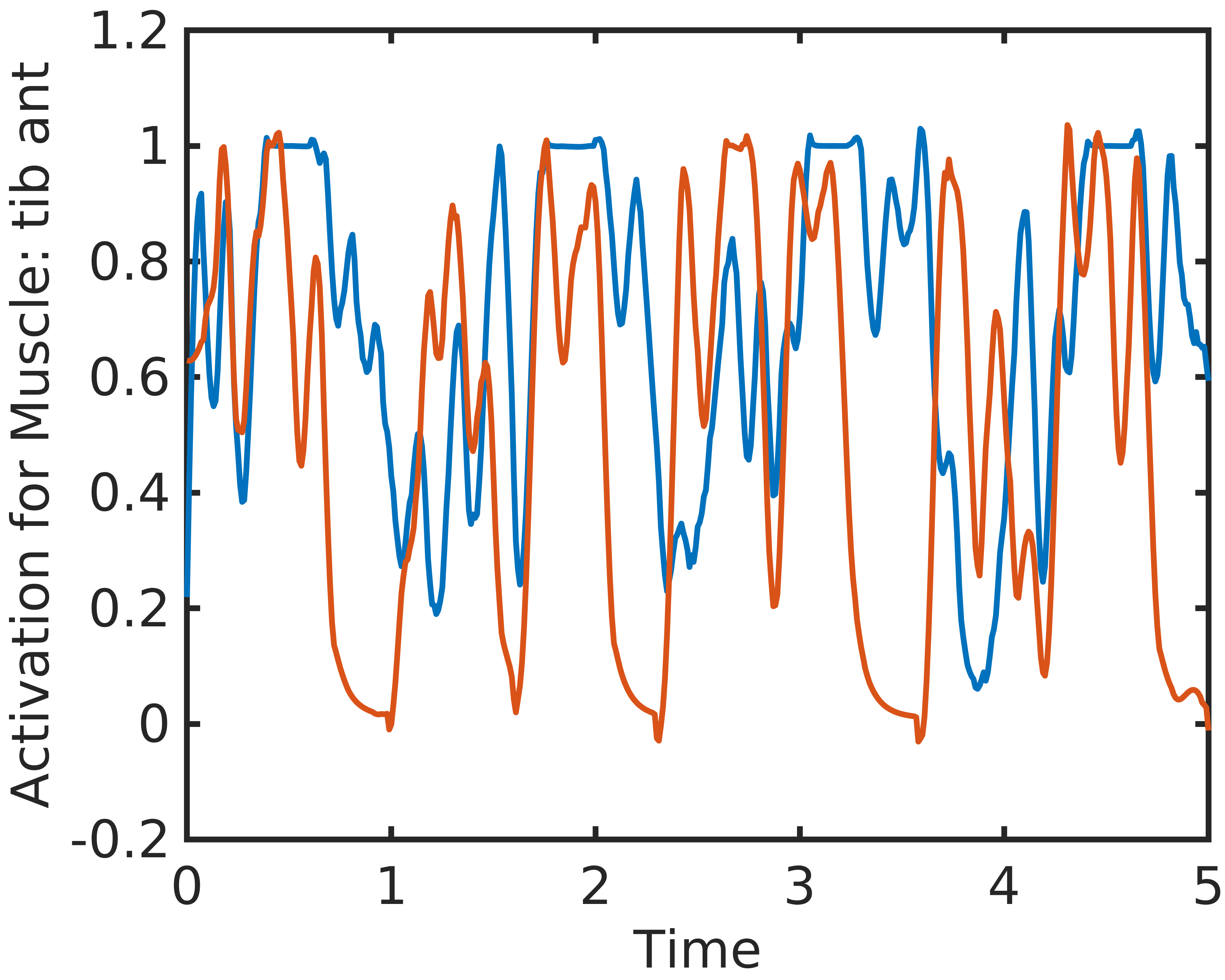}
  \caption{Tibia Anterior}
  \label{fig:sfig232}
\end{subfigure}
\caption{Muscle Activations for MTU actuated walking with proper validation of symmetry as observed for the activation curve (BLue for left and Red for right leg) for specific muscles.}
\label{fig:actmus}
\end{figure}

Based on the training configuration, the reference data available during the training comprised only a single gait cycle. Thus, this analysis is performed to test the benefit of using a phase variable and if the controller policy can correlate the periodicity in control actions with periodicity in phase. The results of this analysis are shown in Fig. \ref{fig:phasemus} where phase plots for left and right hip flexion and knee joints. This shows the learning of a definite and repetitive cycle, which allows for infinite walking strategy based on one gait cycle. This fact was also verified by visualizing the muscle activation plots in Fig. \ref{fig:actmus}. This should be noted that these plots show muscle activations, whereas the control actions are the muscle neural excitations. Four muscles were considered, namely, the Gluteus Maximus, Hamstrings, Iliopsoas, and Tibia Anterior for both the legs.

\newpage

\subsection{Torque Comparison Analysis with Torque Actuated Results}

Based on the muscle forces applied due to their activations, such forces were mapped to the generalized moments' vector using the moment arm matrix as a function of the model's state. Conceptually, the joint torques generated by a given MTU are a function of the current body configuration. A simple variable moment arm model is assumed for MTUs attached to the knee or ankle: $\tau = r_j \cos (\theta - \varphi^M_j) F^{MTU}$, where $\theta$ is the current knee or ankle angle in the sagittal plane, and $r_j$ is the maximum MTU-joint moment arm, which occurs at the joint angle $\varphi^M_j$. MTUs attached to the hip are assumed to have a constant moment arm: $\tau = r_j F_{MTU}$. The total lower extremity joint torques in the sagittal plane are obtained by summing over contributions from all relevant muscles given by:
\begin{gather}
    \tau^{hip} = \tau_{GLU} + \tau^{hip}_{HAM} - \tau_{ILI} \\
    \tau^{knee} = \tau_{VAS} - \tau^{knee}_{HAM} - \tau^{knee}_{GAS} \\
    \tau^{ankle} = \tau^{knee}_{GAS} + \tau_{SOL} - \tau_{[TIA}]
\end{gather}
where the subscript denotes the muscle exerting that force, and superscript represents the moment arm value at the considered joint corresponding to the muscle. A more detailed visualization regarding the muscle connections is already discussed in Fig. \ref{fig:modelpic}. The joint moments obtained from the above Eq.~(27-29) were compared with the control torques obtained from the torque actuated training, and the results are shown below in Fig. \ref{fig:torquecomp}.
\begin{figure}[b!]
\centering
\begin{subfigure}{.38\textwidth}
  \centering
  \includegraphics[width=0.9\linewidth]{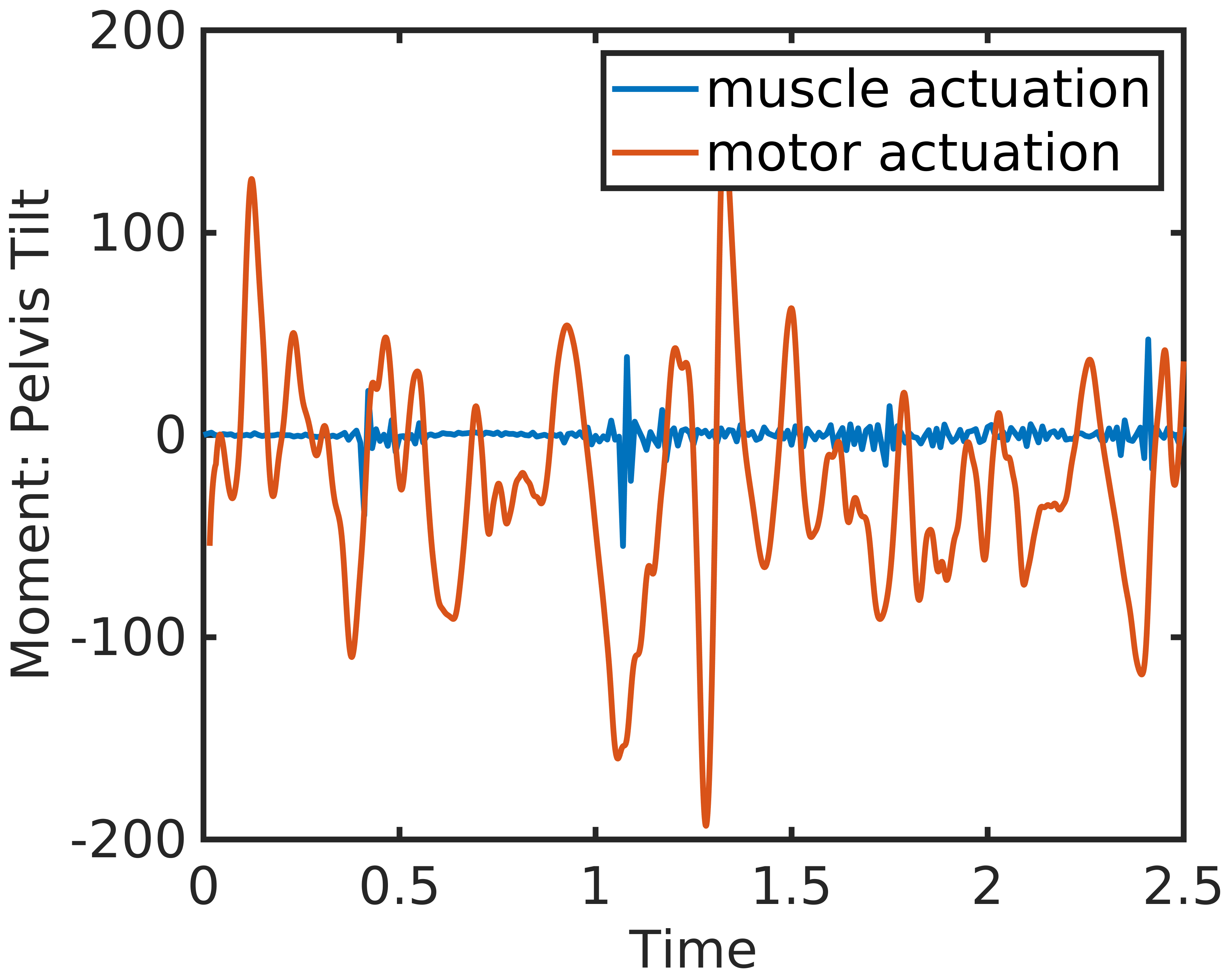}
  \caption{Pelvis Tilt}
  \label{fig:sfig21}
\end{subfigure}%
\begin{subfigure}{.38\textwidth}
  \centering
  \includegraphics[width=0.9\linewidth]{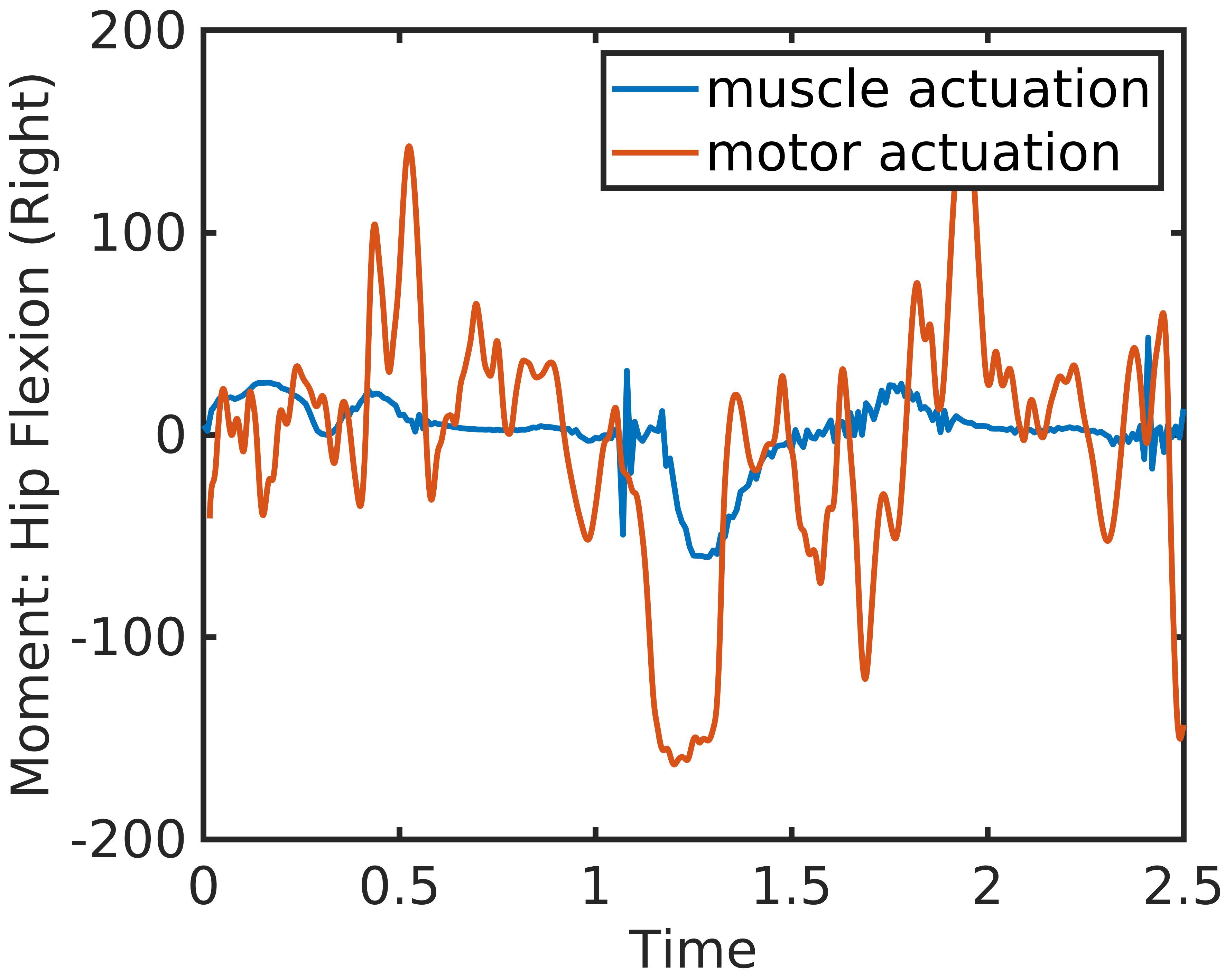}
  \caption{Right Hip Flexion}
  
\end{subfigure}
\begin{subfigure}{.38\textwidth}
  \centering
  \includegraphics[width=0.9\linewidth]{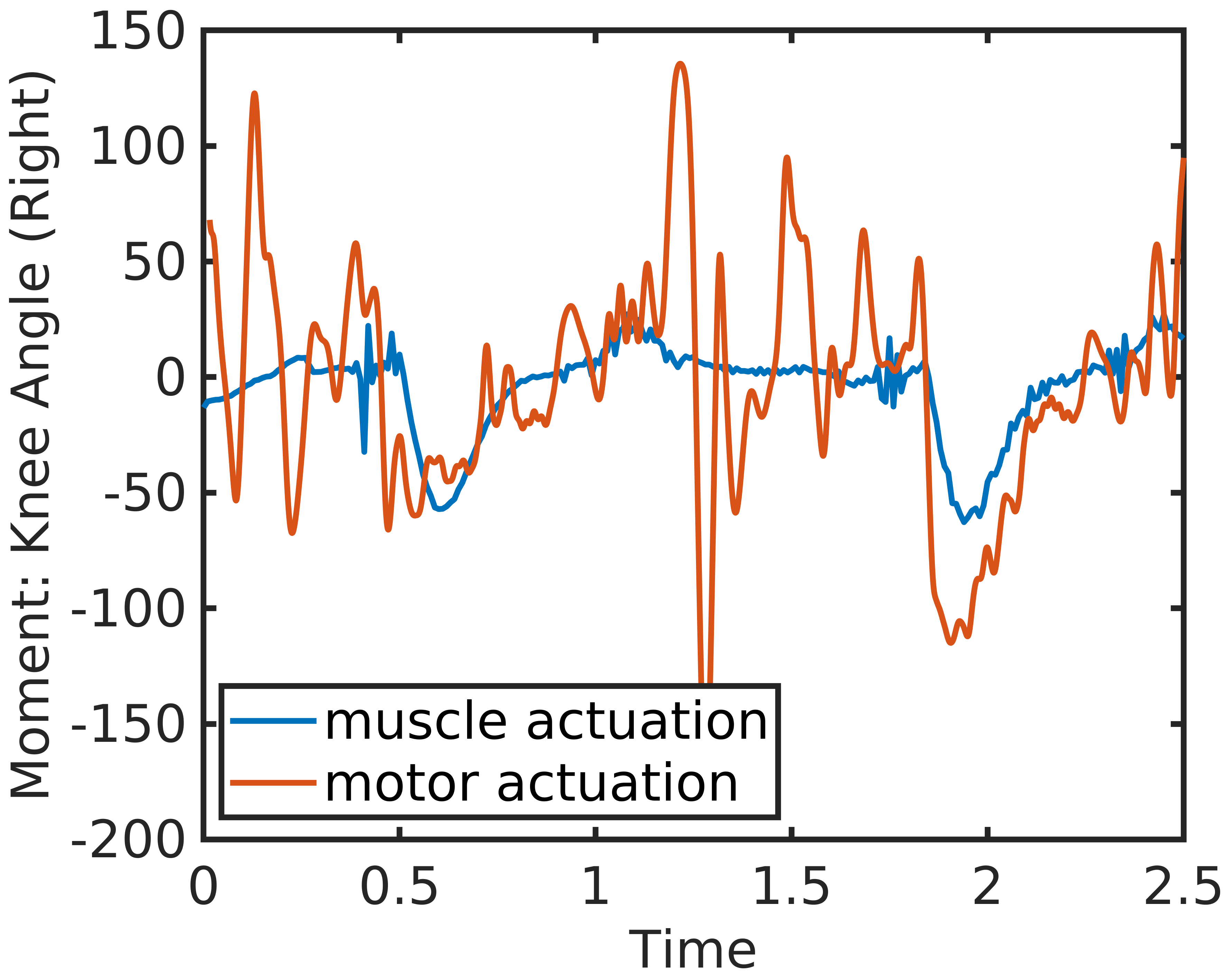}
  \caption{Right Knee Angle}
  \label{fig:sfig22}
\end{subfigure}%
\begin{subfigure}{.38\textwidth}
  \centering
  \includegraphics[width=0.9\linewidth]{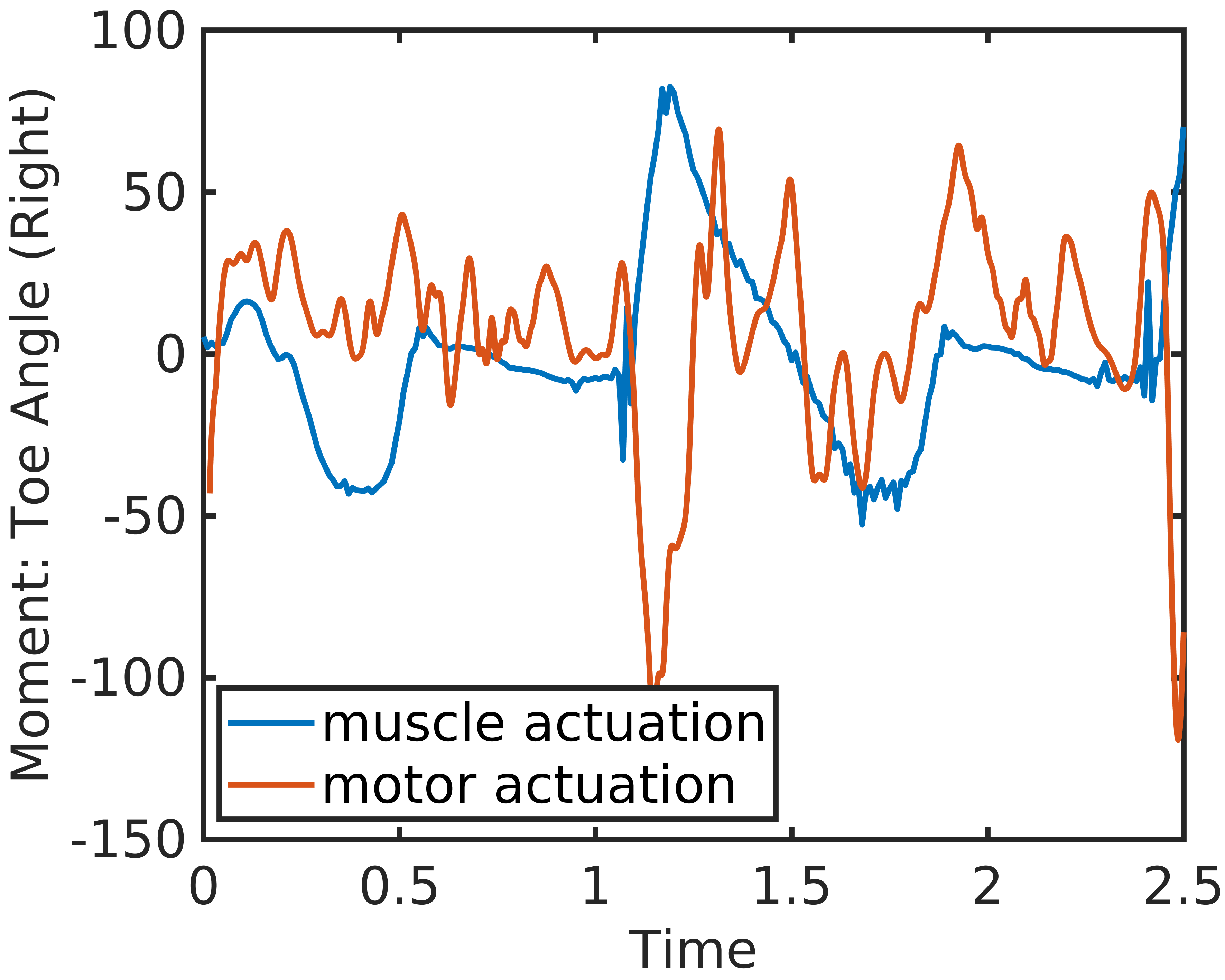}
  \caption{Right Toe Angle}
  
\end{subfigure}
\caption{Comparison of torque profiles based on torque actuated and MTU actuated framework for imitation of the walking gait.}
\label{fig:torquecomp}
\end{figure}

\newpage

\section{Algorithm Hyper-parameter Tuning and Results}
\label{sec:tuning}

The algorithm for imitating a given reference gait based on reinforcement learning has a lot of decision variables. These variables significantly affect the performance and quality of the solutions obtained. Various hyper-parameters have been introduced so far in Section D. The training configurations discussed hereafter aim to find the most optimal configuration strategically.

\subsection{Analysis with Varying Network Size}
\label{sec:tuning_nn}

\begin{figure}[b!]
\centering
\begin{subfigure}{.35\textwidth}
  \centering
  \includegraphics[width=0.9\linewidth]{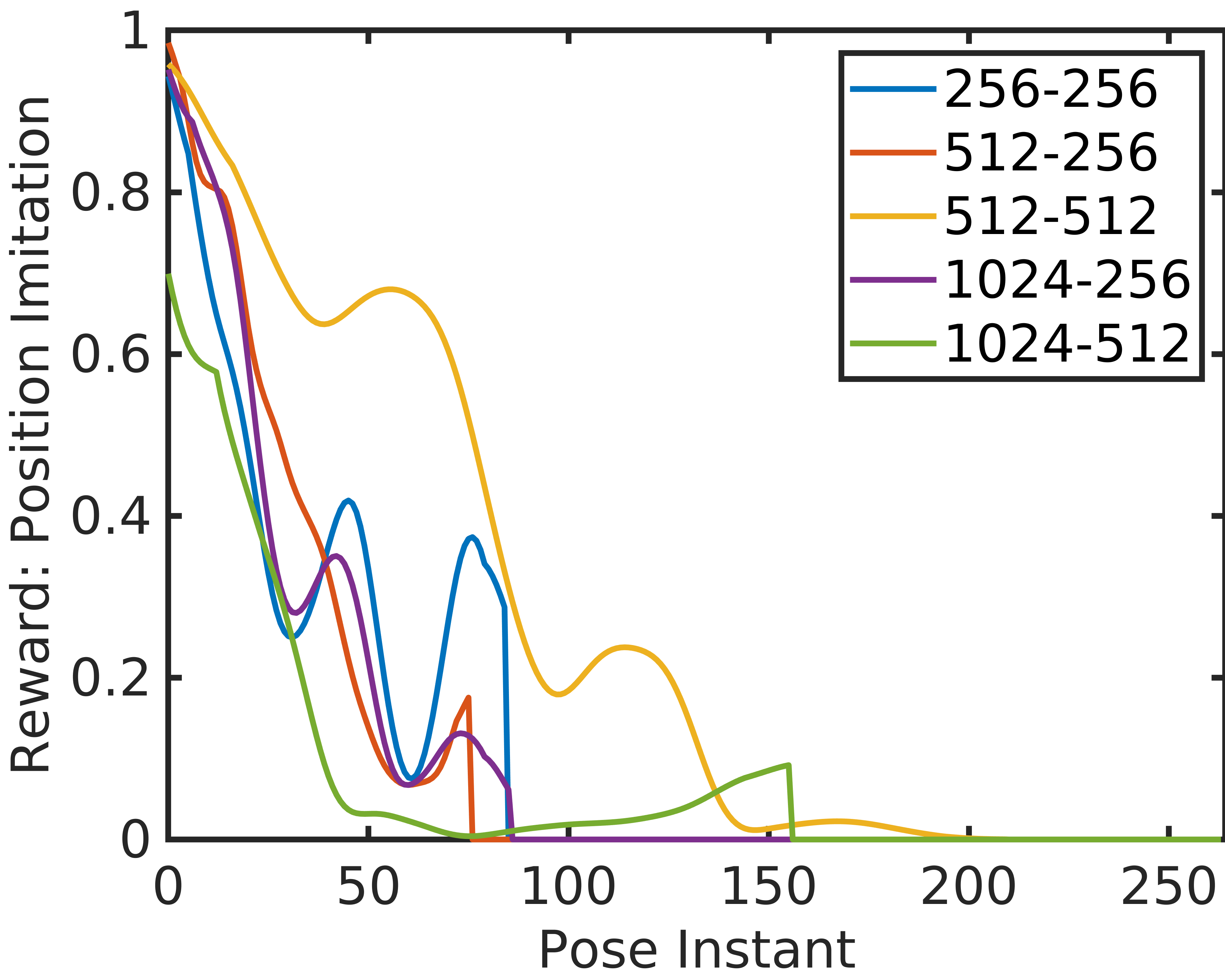}
  \caption{Position Imitation}
  \label{fig:sfig23}
\end{subfigure}%
\begin{subfigure}{.35\textwidth}
  \centering
  \includegraphics[width=0.9\linewidth]{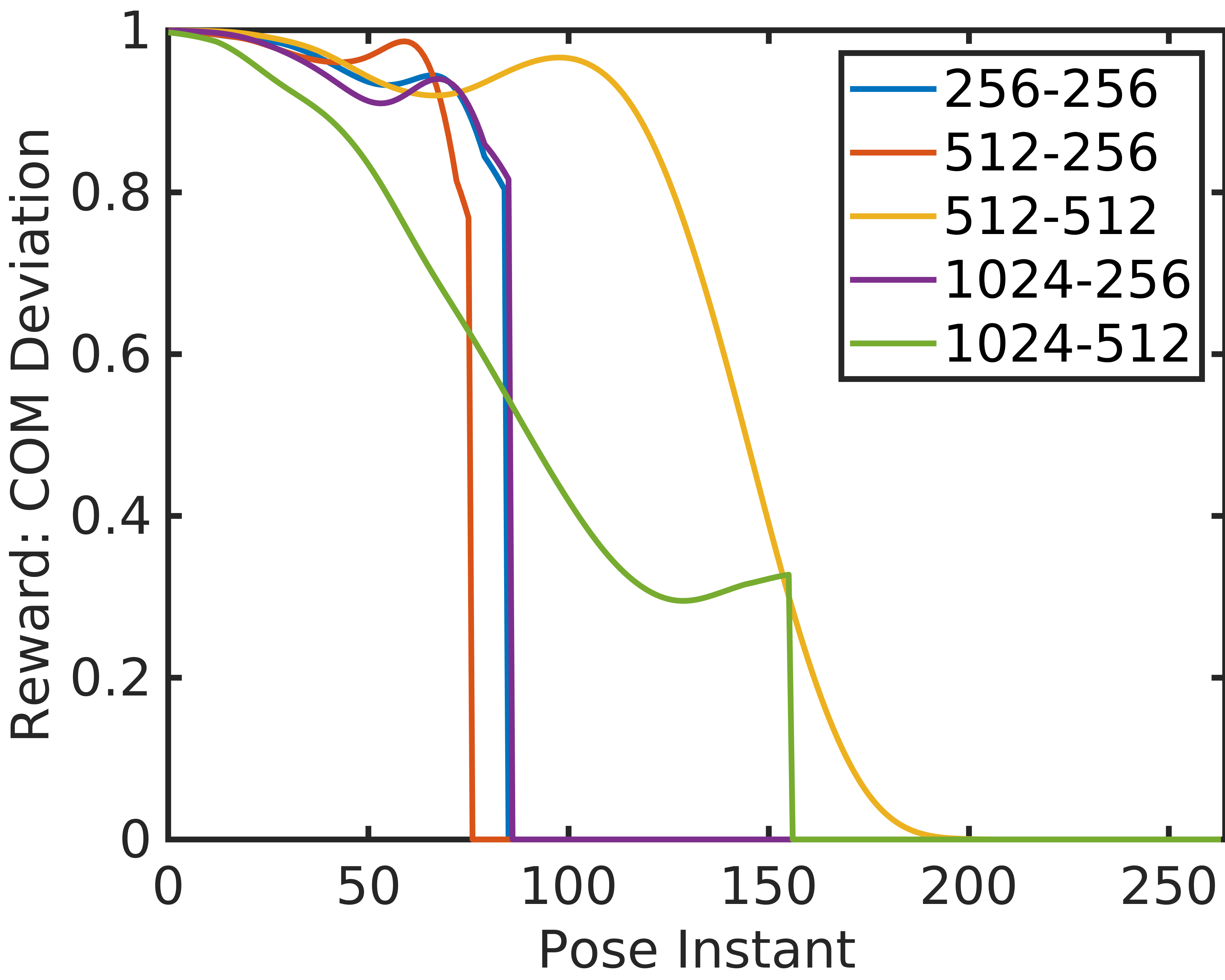}
  \caption{COM Deviation}
  
\end{subfigure}
\begin{subfigure}{.35\textwidth}
  \centering
  \includegraphics[width=0.9\linewidth]{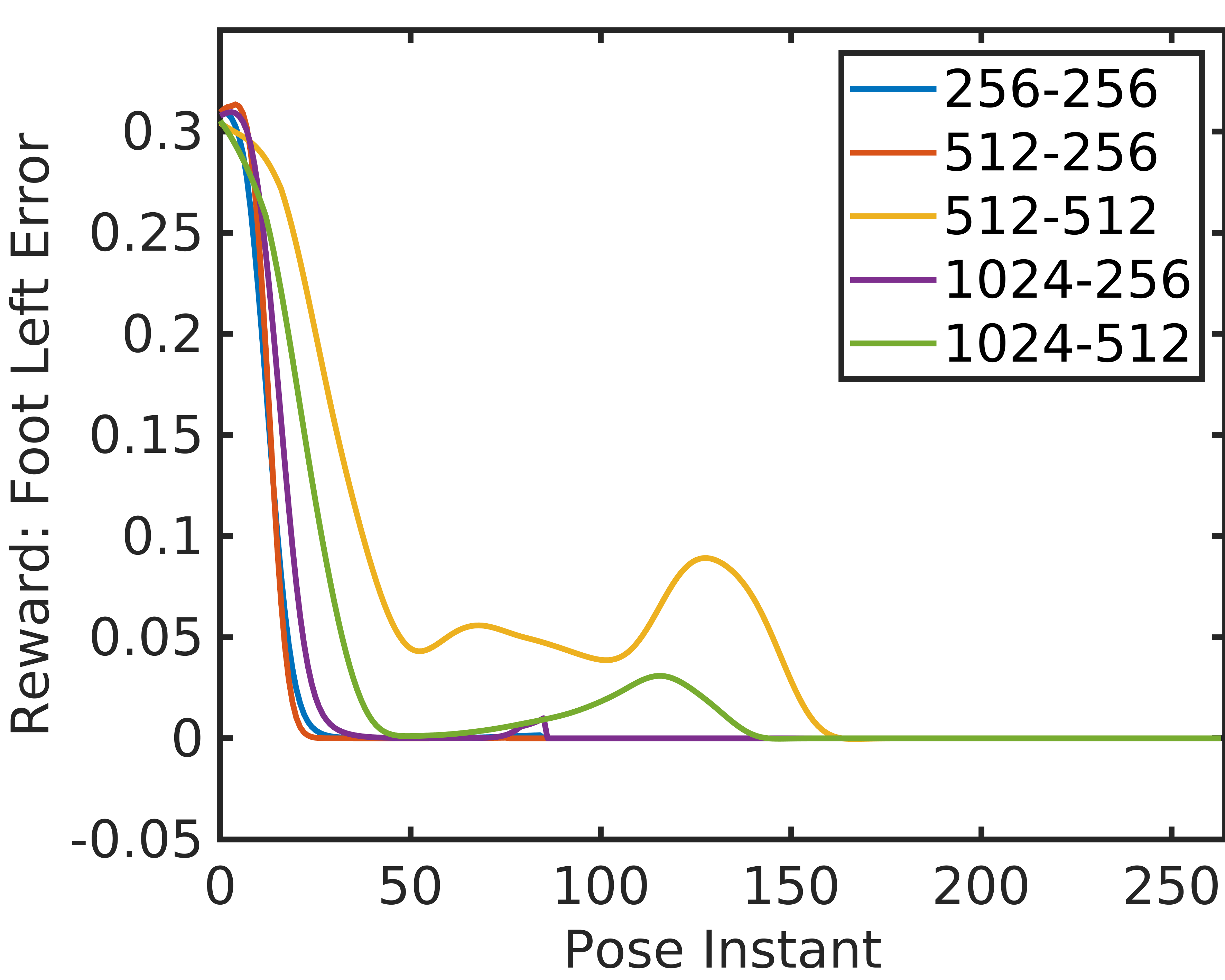}
  \caption{End-Effector Position}
  \label{fig:sfig24}
\end{subfigure}%
\begin{subfigure}{.35\textwidth}
  \centering
  \includegraphics[width=0.9\linewidth]{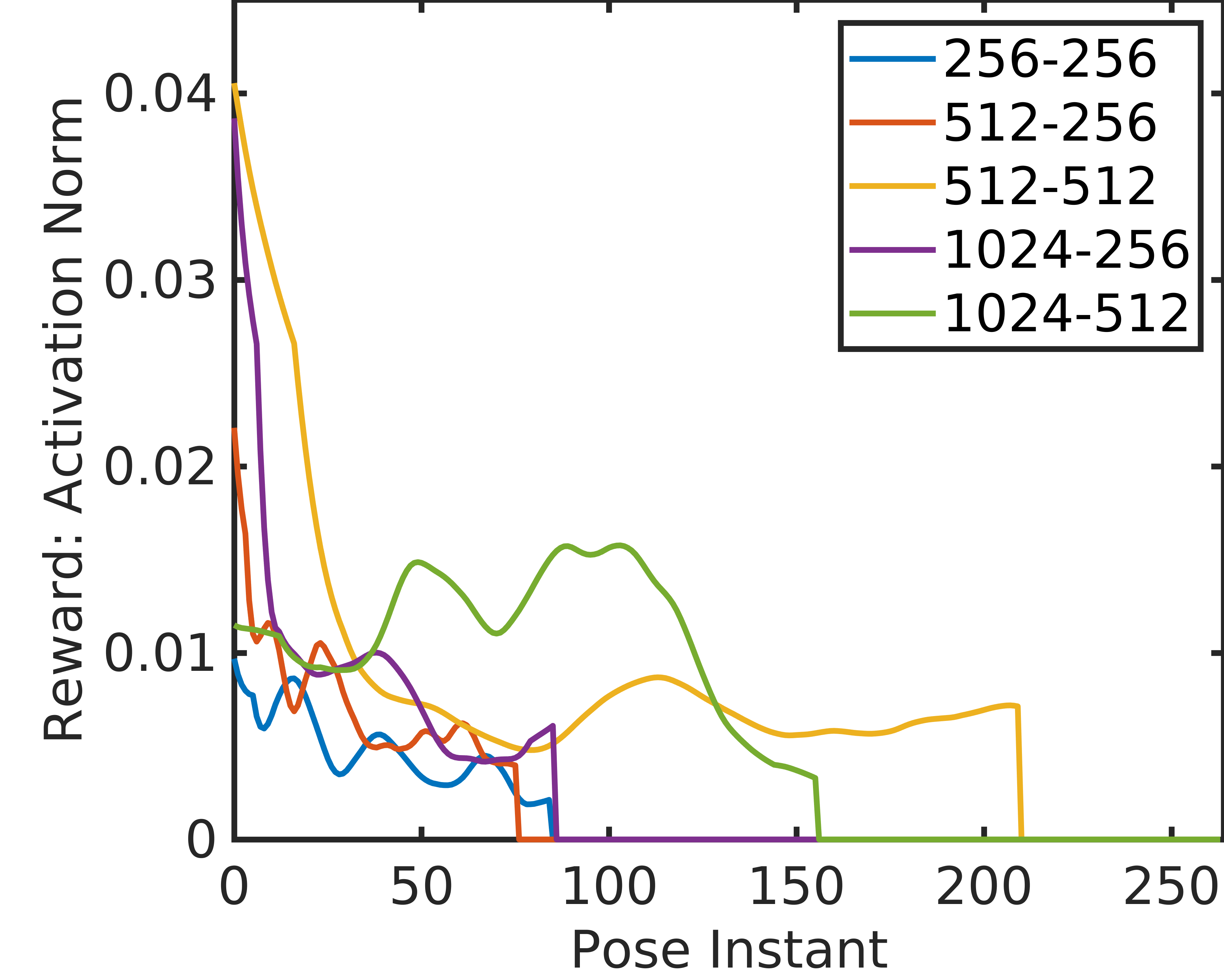}
  \caption{Muscle Activation}
  
\end{subfigure}
\caption{Changes in the individual reward terms with change in Policy (Actor Neural Network) architecture. The architecture with \{512, 512\} nodes performed the best i.e. the yellow curve.}
\label{fig:rewnn}
\end{figure}

The network size of the neural networks plays a vital role in representing the non-linear approximation of the transition dynamics of the interactions between the model and the environment. With the increasing number of layers and more hidden nodes, the number of parameters in the non-linear function increases. Although this might ensure a good approximation or representation of the actual transition dynamics, many parameters might overfit to some common observations and fail to generalize over the others or less likely observations. Also, the time to train the parameters increases, thus increasing the overall convergence time for the algorithm. The experiment was conducted to find the best network architecture with relatively better performance in a limited training time. With fixed number of hidden layers (i.e. 2), the hidden nodes for each layer were considered as $\{256, 256\}$, $\{256, 512\}$, $\{512, 512\}$, $\{1024, 256\}$ and $\{1024, 512\}$. The performance was evaluated based on individual reward response as in Section \ref{sec:algorithms_reward} and shown in Fig. \ref{fig:rewnn}. 

\newpage

\subsection{Analysis with Varying Observations}
\label{sec:tuning_obs}

\begin{figure}[b!]
\centering
\begin{subfigure}{.35\textwidth}
  \centering
  \includegraphics[width=0.9\linewidth]{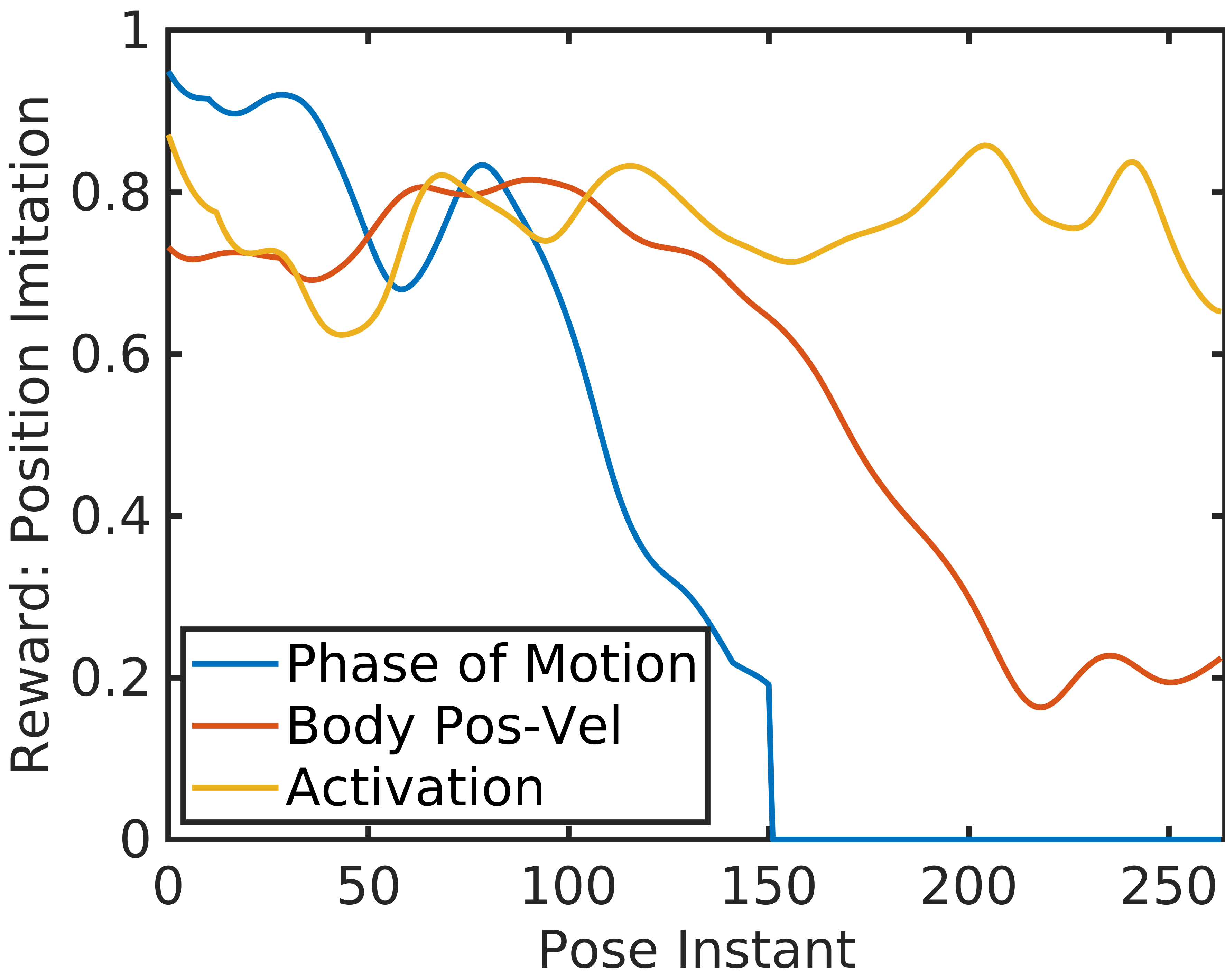}
  \caption{Position Imitation}
  \label{fig:sfig25}
\end{subfigure}%
\begin{subfigure}{.35\textwidth}
  \centering
  \includegraphics[width=0.9\linewidth]{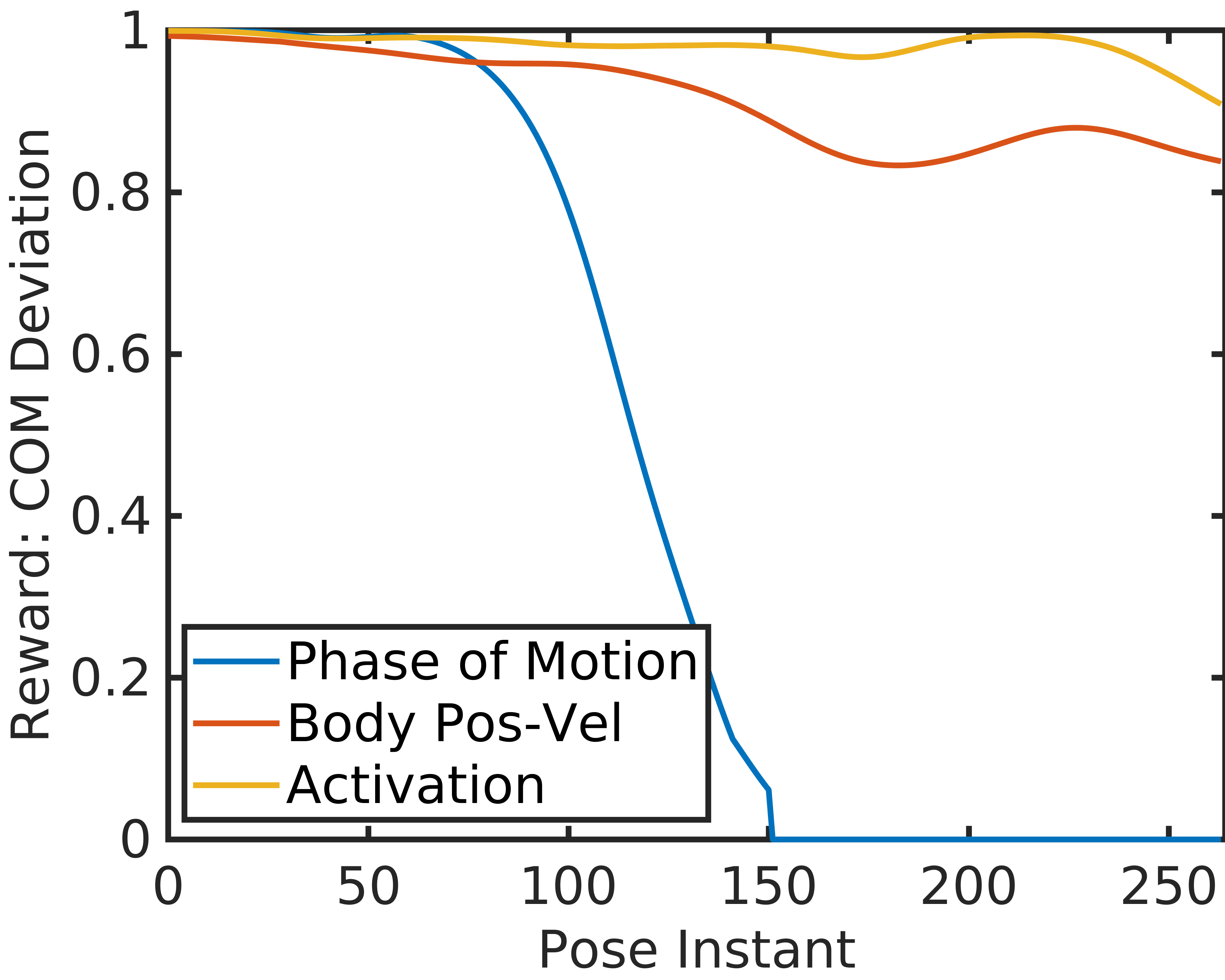}
  \caption{COM Deviation}
  
\end{subfigure}
\begin{subfigure}{.35\textwidth}
  \centering
  \includegraphics[width=0.9\linewidth]{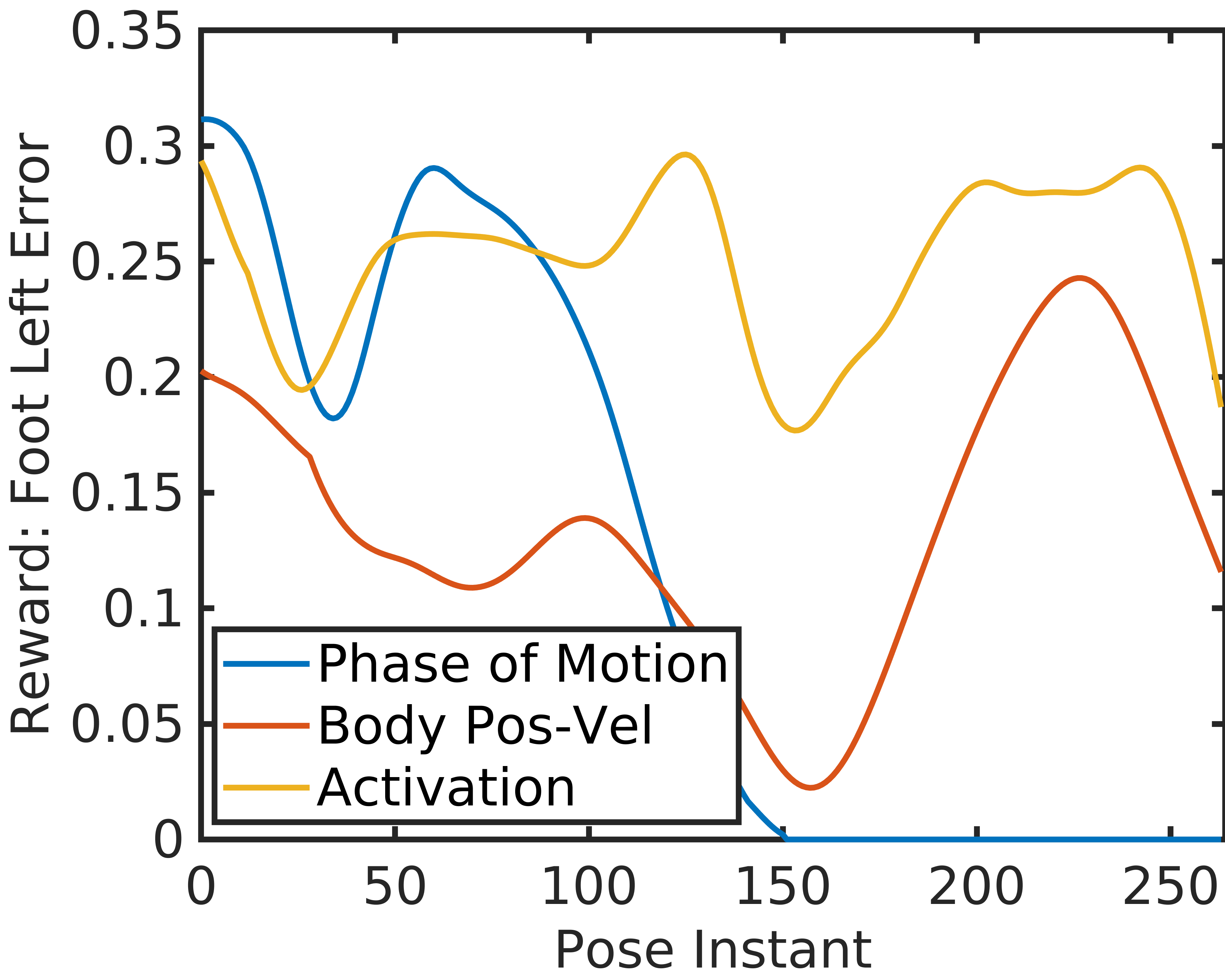}
  \caption{End-Effector Position}
  \label{fig:sfig26}
\end{subfigure}%
\begin{subfigure}{.35\textwidth}
  \centering
  \includegraphics[width=0.9\linewidth]{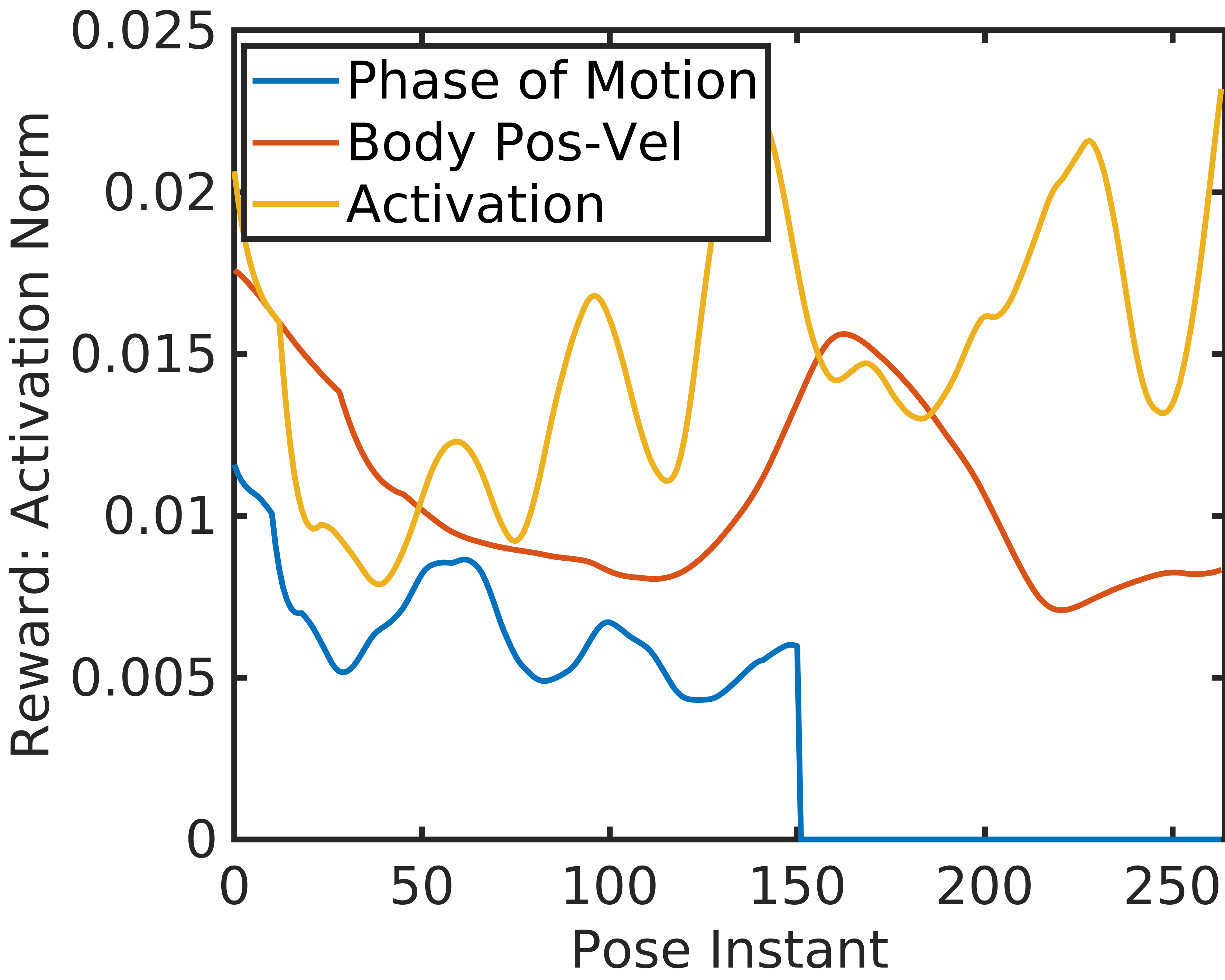}
  \caption{Muscle Activation}
  
\end{subfigure}
\caption{Changes in the individual reward terms with change in entities included in the Observations.}
\label{fig:rewobs}
\end{figure}

The lesser the quantity of observations, the easier is to learn their correspondence with the control actions. The experiments conducted and their performance are given in Table \ref{table:rewobs} and Fig. \ref{fig:rewobs} respectively in order to choose the most optimal contents of the observations.

\begin{longtable}[b!]{|p{0.20\linewidth}|p{0.5\linewidth}|p{0.1\linewidth}|}
\hline
\centering\arraybackslash \textbf{Experiment Name (in plot)} &  \centering\arraybackslash \textbf{Experiment Configuration} & \centering\arraybackslash \textbf{Chosen}\\
\hline
\centering\arraybackslash \textbf{Phase of Motion} &  Phase of the motion was not included. Observation space contained coordinate position, velocity and acceleration. For muscles, fiber length and velocity were included. The resulting policy was unable to balance after one gait cycle. & \centering\arraybackslash\xmark \\
\hline
\centering\arraybackslash \textbf{Body Pos-Vel} &  Phase and the body position and velocity were added to the observation in the previous experiment. The results improved with some increase in rewards & \centering\arraybackslash\xmark \\
\hline
\centering\arraybackslash \textbf{Activation} &  Activation of the muscles were added to ensure that the activation dynamics in modelled. With this addition, proper reward curves are observed. & \centering\arraybackslash\cmark \\
\hline
\caption{Experiment Details for Observation Space Tuning}
\label{table:rewobs}
\end{longtable}

\newpage

\subsection{Analysis with Varying Reward Terms}
\label{sec:tuning_rew}

\begin{figure}[b!]
\centering
\begin{subfigure}{.40\textwidth}
  \centering
  \includegraphics[width=0.9\linewidth]{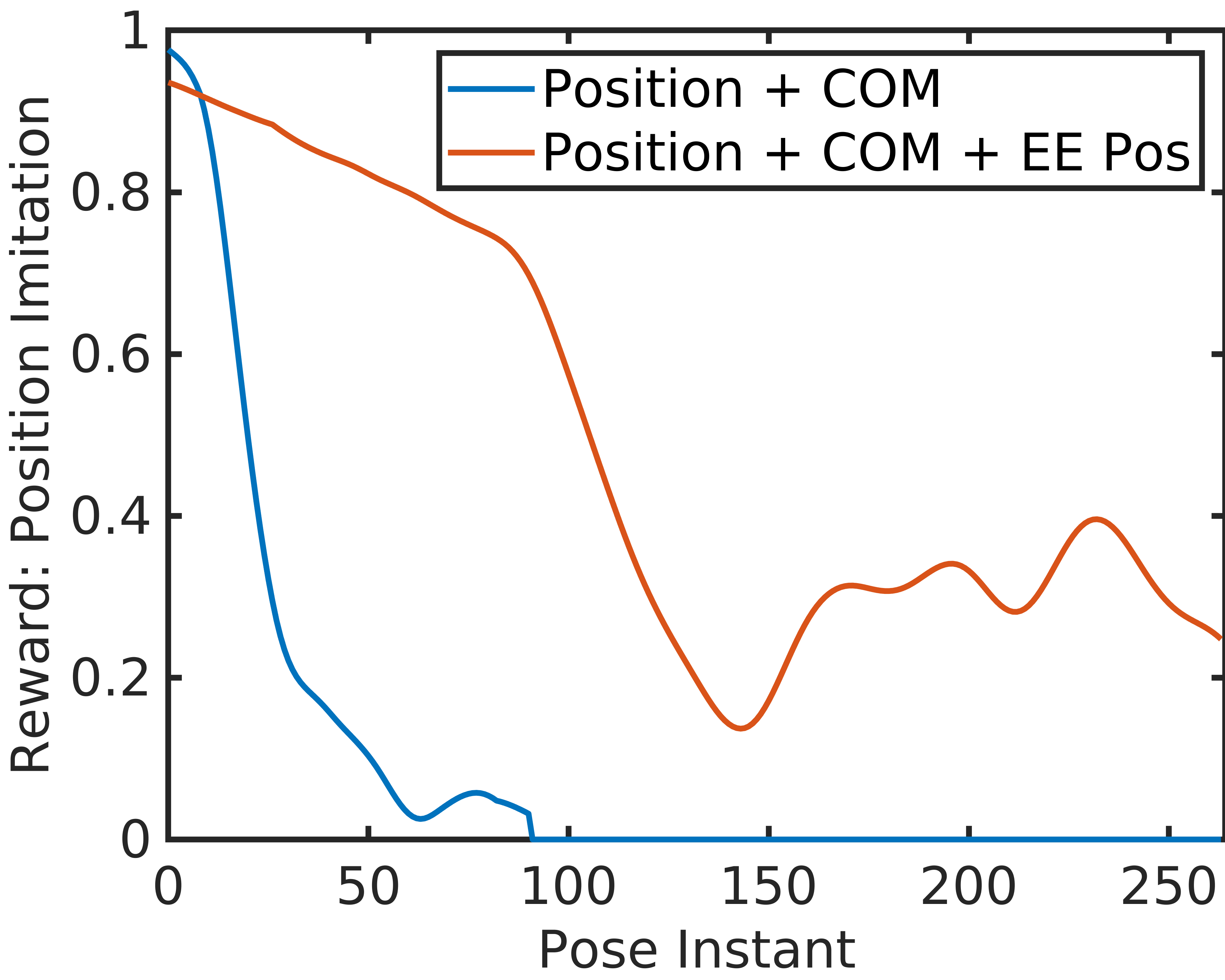}
  \caption{Position Imitation}
  \label{fig:sfig27}
\end{subfigure}%
\begin{subfigure}{.40\textwidth}
  \centering
  \includegraphics[width=0.9\linewidth]{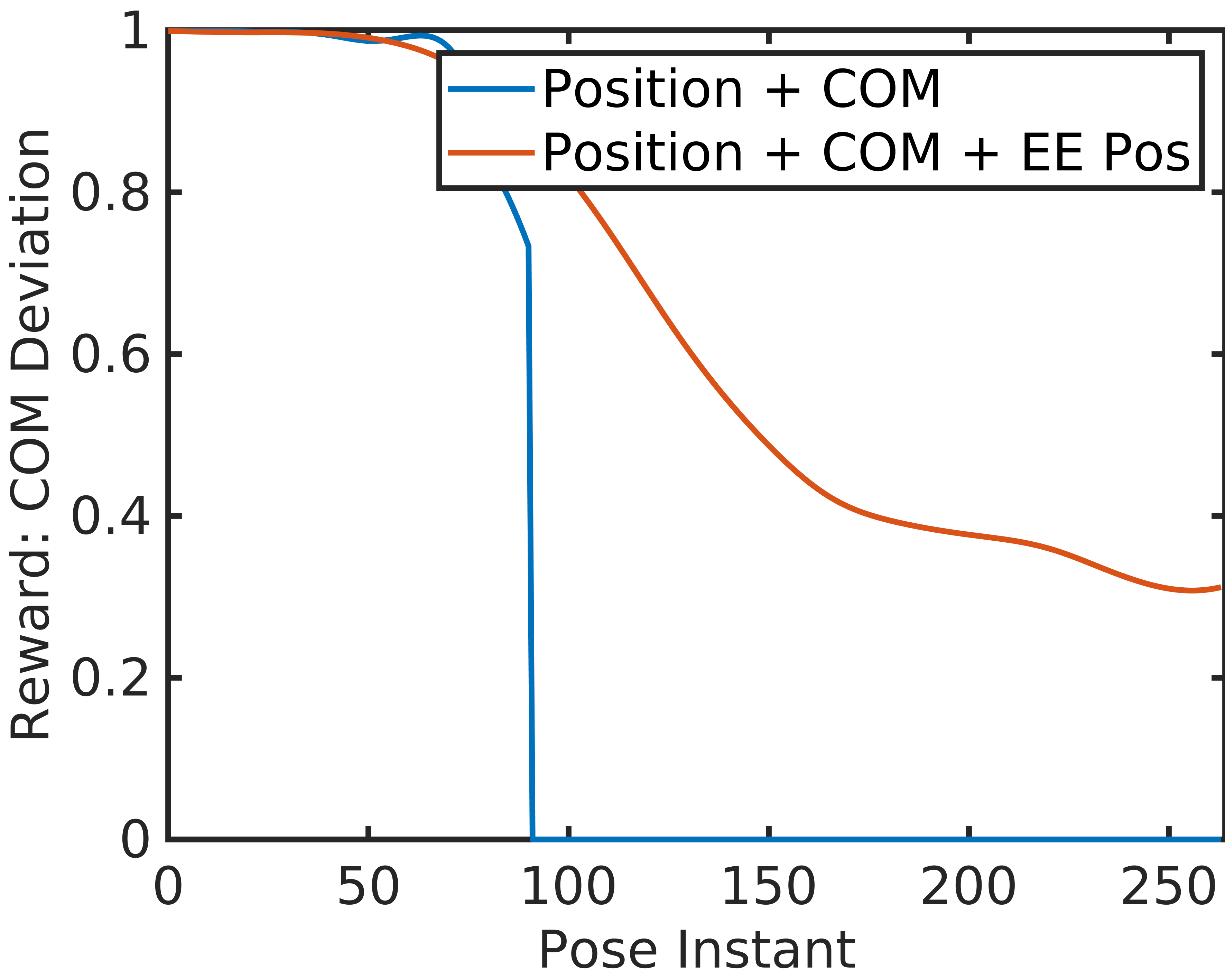}
  \caption{COM Deviation}
  
\end{subfigure}
\begin{subfigure}{.40\textwidth}
  \centering
  \includegraphics[width=0.9\linewidth]{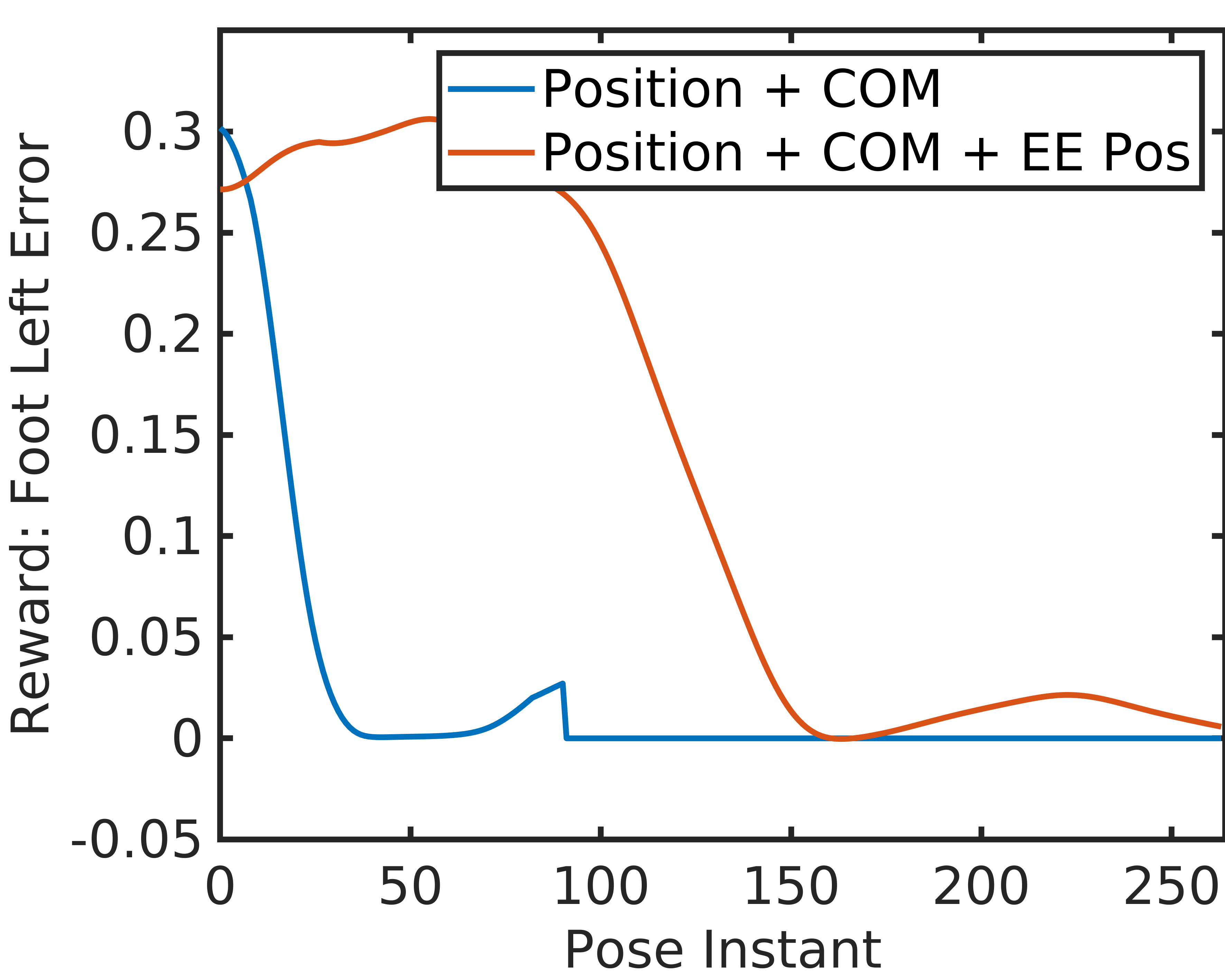}
  \caption{End-Effector Position}
  \label{fig:sfig28}
\end{subfigure}%
\begin{subfigure}{.40\textwidth}
  \centering
  \includegraphics[width=0.9\linewidth]{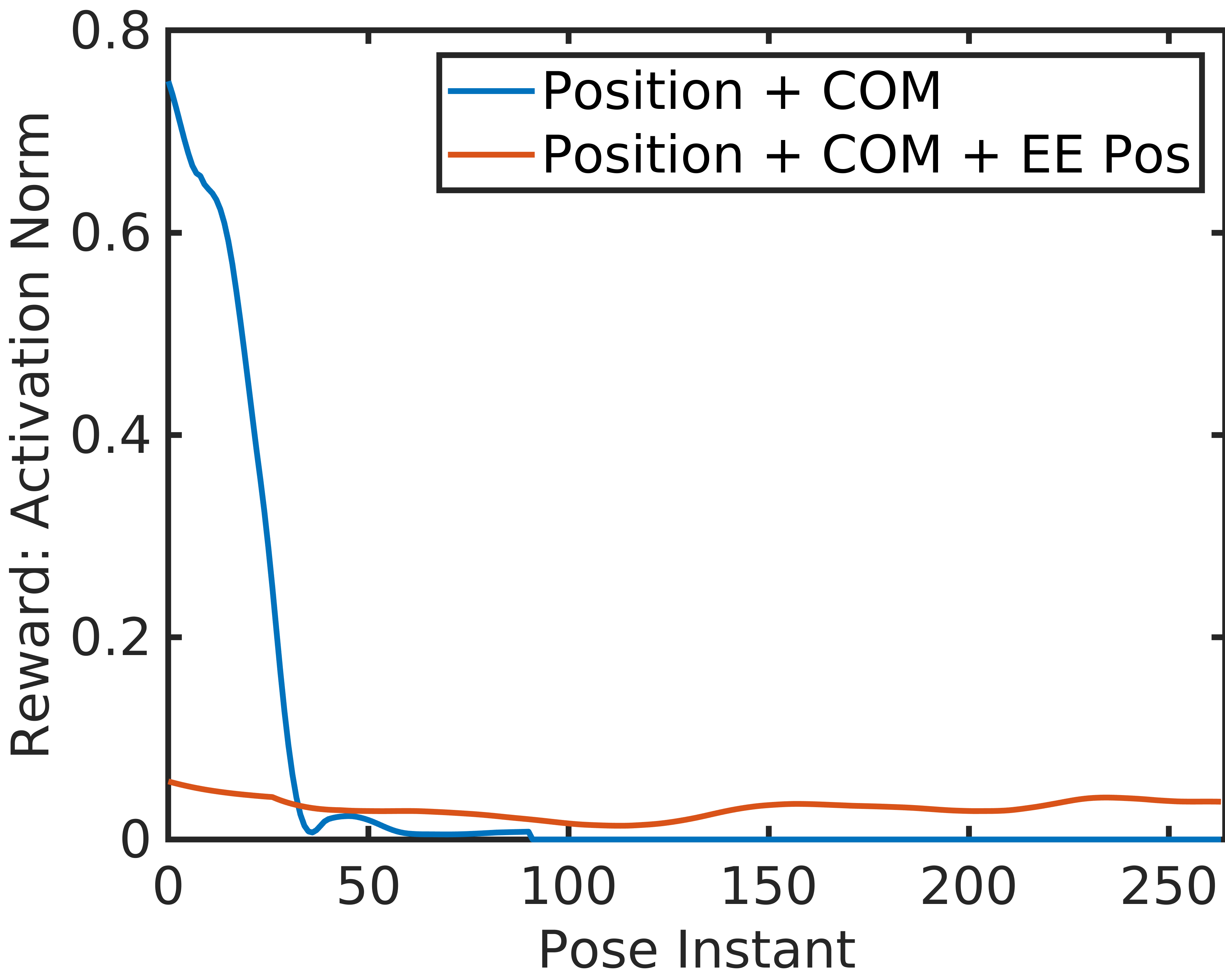}
  \caption{Muscle Activation}
  
\end{subfigure}
\caption{Changes in the individual reward terms with change in terms included in the Reward function.}
\label{fig:rewrew}
\end{figure}

The reward function is the optimization objective and should precisely represent the task to be solved. Too much restrictive rewards make it harder for the algorithm to learn, whereas leniency can lead to undesired outcomes. The experiments conducted and their performance are given in Table \ref{table:rewrew} and Fig. \ref{fig:rewrew} respectively in order to choose the most optimal contents of the reward function. The considered rewards are position imitation, COM imitation, foot position imitation, and minimal muscle excitations.

\begin{longtable}[b!]{|p{0.20\linewidth}|p{0.5\linewidth}|p{0.1\linewidth}|}
\hline
\centering\arraybackslash \textbf{Experiment Name (in plot)} &  \centering\arraybackslash \textbf{Experiment Configuration} & \centering\arraybackslash \textbf{Chosen}\\
\hline
\centering\arraybackslash \textbf{Position + COM} &  Position Imitation is the most basic reward. Apart from that COM deviation also plays a vital role in order to help in better generalization. However, the policy failed to generalize. & \centering\arraybackslash\xmark \\
\hline
\centering\arraybackslash \textbf{Position + COM + EE Pos} &  With addition of the foot (end-effector) position relative to pelvis, appropriate foot motion was obtained leading to more stable limit cycles and better generalization. & \centering\arraybackslash\cmark \\
\hline
\caption{Experiment Details for Reward Function Formulation Tuning}
\label{table:rewrew}
\end{longtable}

\newpage

\subsection{Analysis with Varying Discount Factor}
\label{sec:tuning_gam}

\begin{figure}[b!]
\centering
\begin{subfigure}{.40\textwidth}
  \centering
  \includegraphics[width=0.9\linewidth]{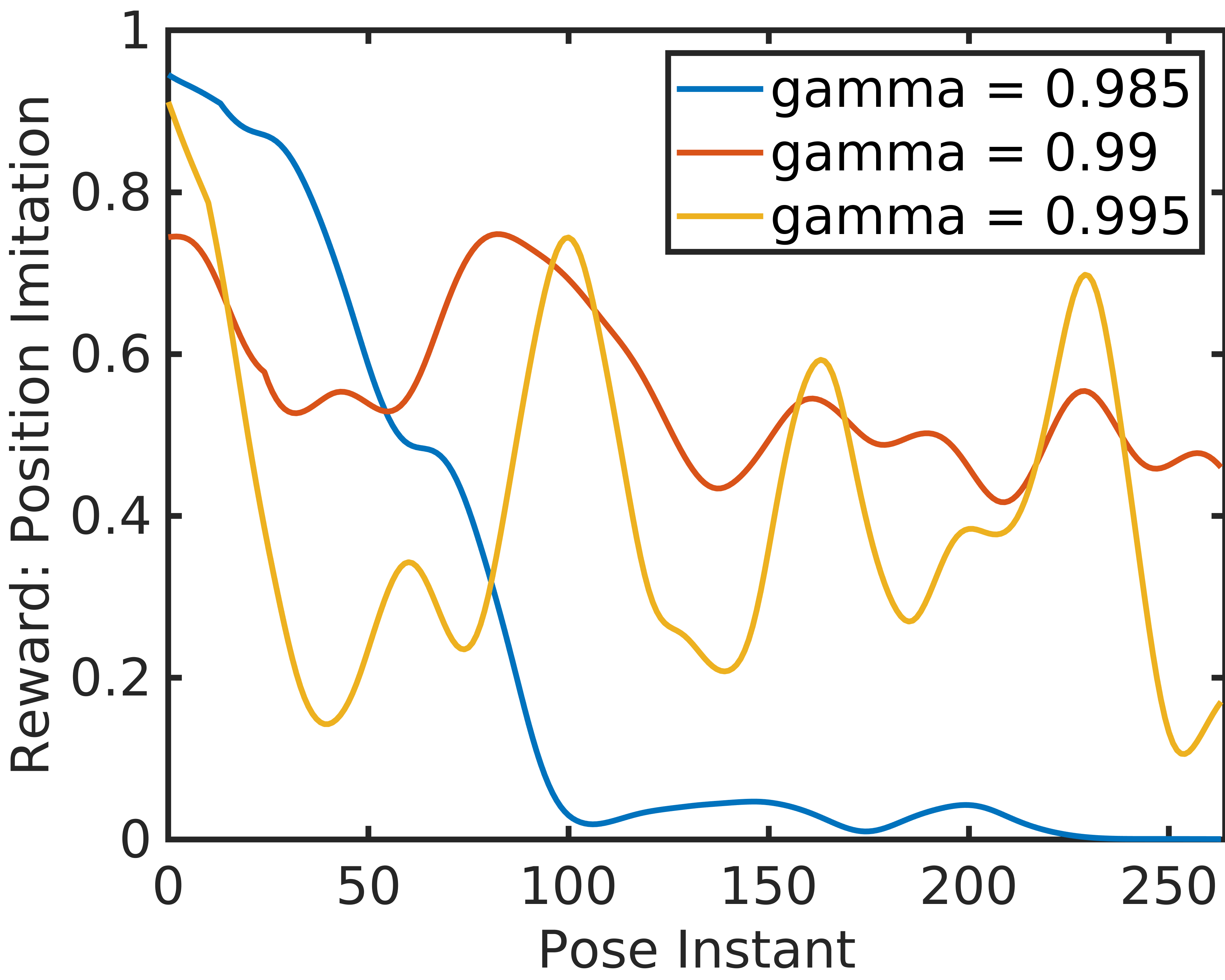}
  \caption{Position Imitation}
  \label{fig:sfig29}
\end{subfigure}%
\begin{subfigure}{.40\textwidth}
  \centering
  \includegraphics[width=0.9\linewidth]{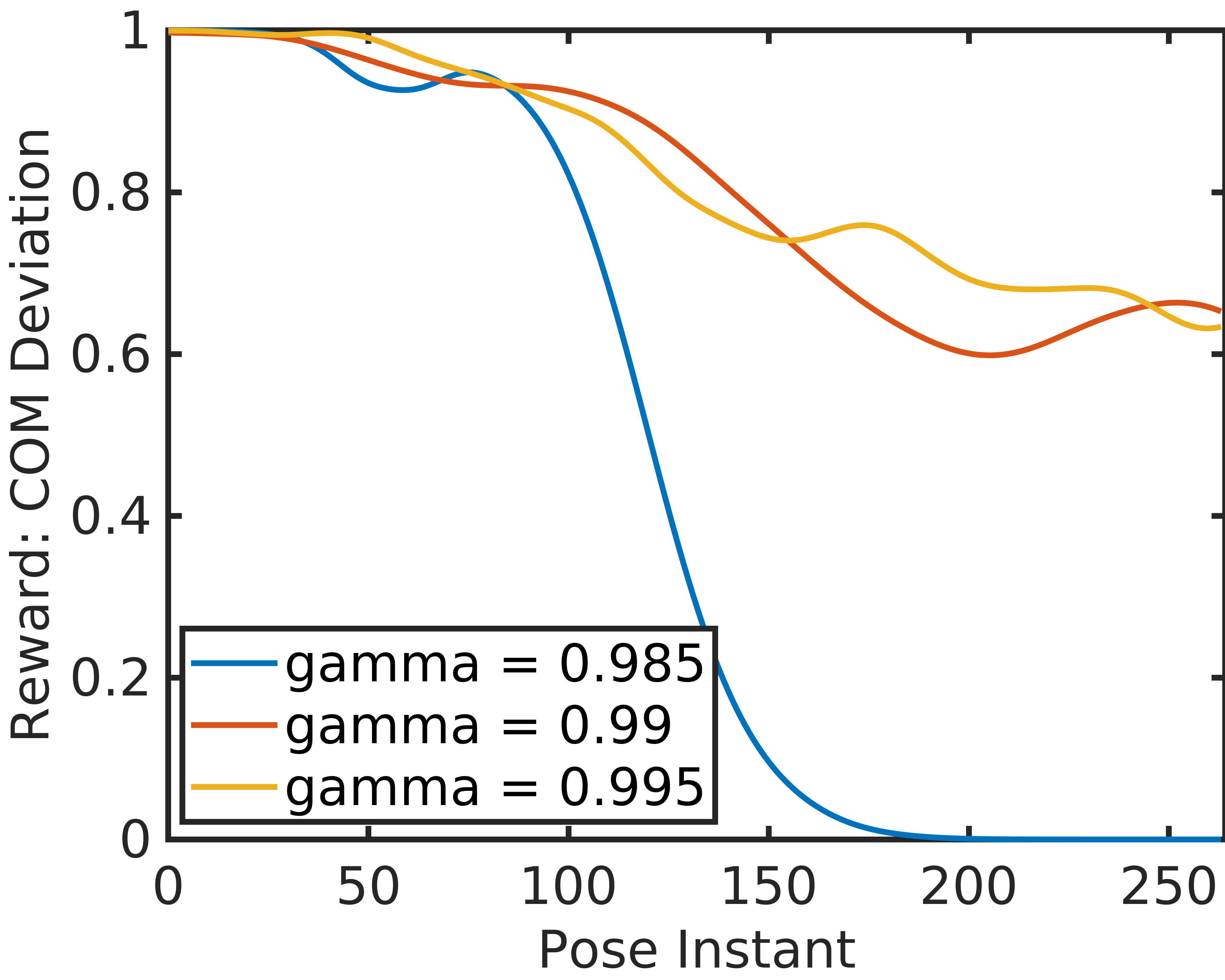}
  \caption{COM Deviation}
  \label{fig:sfig221}
\end{subfigure}
\begin{subfigure}{.40\textwidth}
  \centering
  \includegraphics[width=0.9\linewidth]{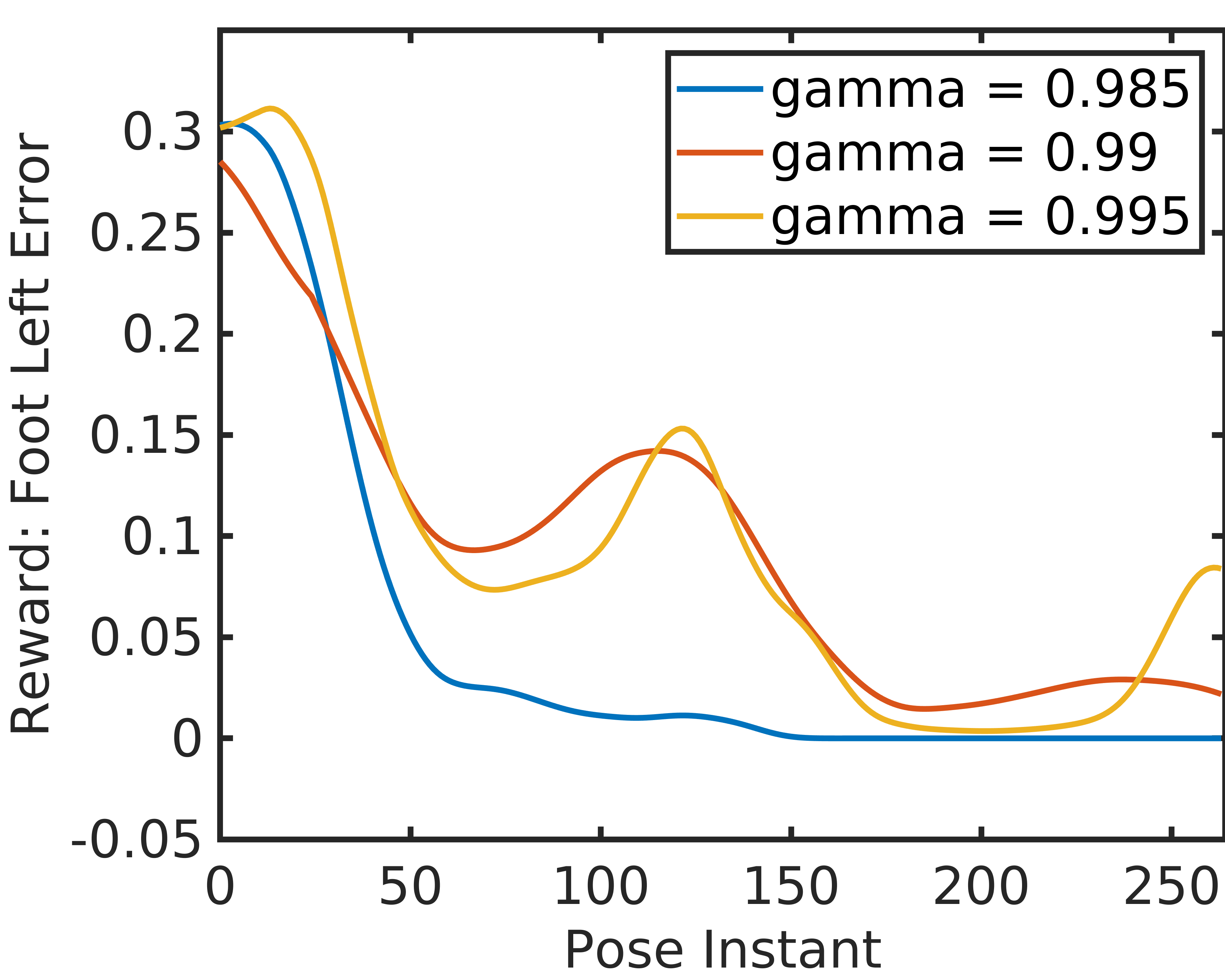}
  \caption{End-Effector Position}
  \label{fig:sfig30}
\end{subfigure}%
\begin{subfigure}{.40\textwidth}
  \centering
  \includegraphics[width=0.9\linewidth]{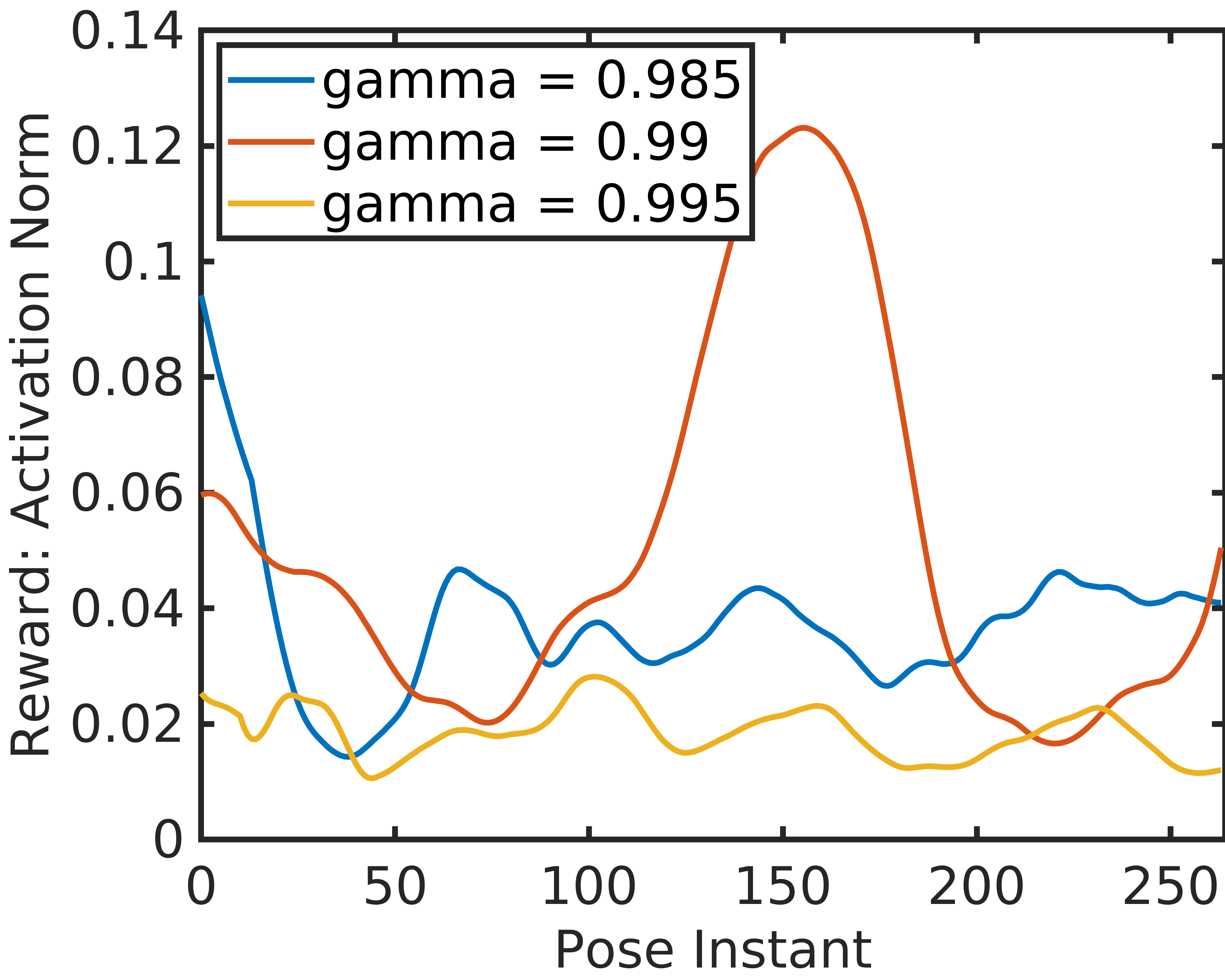}
  \caption{Muscle Activation}
  \label{fig:sfig222}
\end{subfigure}
\caption{Changes in the individual reward terms with change in reward discount factor, $\gamma$.}
\label{fig:rewgam}
\end{figure}

The discount factor ($\gamma$) is associated with the time horizon. Longer time horizons have much more variance due to presence of more irrelevant information, while the shorter time horizons are biased towards a greedy approach or reacting to short-term gains. The discount factor determines how much the RL agent is concerned about the rewards it receives in the distant future as compared to those in the immediate future. If $\gamma=0$, the agent will be wholly myopic and will only learn policies that focus on achieving rewards immediately. If $\gamma=1$, the agent will evaluate each of its actions on an infinite horizon setting based on the sum of all of its future rewards. Thus, the discount factor should be chosen in a way such that the time horizon contains all of the relevant rewards for a particular action, but not anymore. For a particular value of $\gamma$, the agent receives a discounted reward $G_t$, represented as a function of all the rewards that will be collected following the current policy and starting from the current state. This discounted reward is given by:
\begin{equation}
    G_t = R_t + \gamma R_{t+1} + \gamma^2 R_{t+2} + \dots
    \implies G_t = \sum^{\infty}_{k=0} \gamma^k R_{t+k}
\end{equation}
As $\gamma \in [0,1]$, the summation attenuates as $\gamma^k \longrightarrow 0$. For the experiment to choose the most favourable value of $\gamma$, three choices were considered, $\gamma = 0.985$, $\gamma = 0.99$ and $\gamma = 0.995$. Based on the reward analysis plots in Fig. \ref{fig:rewgam}, higher values of $\gamma = 0.99 \ \text{and} \ 0.995$ gave better results. However, the best one out of them could not be concluded based on this analysis.

\newpage

\chapter{Discussions}
\label{sec:discuss}

The results obtained with the imitation framework give many insights into the nature of the solution acquired based on the motor actuated and muscle-tendon actuated models. The results have illustrated a wide variety of comparison results ranging from motor actuated walking (normal and prosthetic), running and jumping(continuous and high) to MTU actuated walking and running. While many previous works have exhibited precise motion tracking, their formulations contain dynamic compensation terms. Their formulation perfectly fits into their purpose of high-quality animations and dynamic compensation terms nullifying external forces' effects to achieve precise tracking. However, such methods are not suited for practical applications. For example, if the motion is moving against gravity, there is a need for compensation to deal with the external force due to gravity. But suppose a similar motion is going towards the direction of gravity instead of taking advantage of the external gravity force. In that case, it ends up applying forces in the reverse direction to cancel them. Our current work is free from such compensations and directly deals with the dynamic interactions with minimum effort.

The observations provided to the learning agent most closely represent the actual perception qualities of a typical human. With this setting, the torque environment trainings have to deal with unknown ground reactions, and hence they fail to get a stable limit cycle for the ankle joint. Moreover, as seen from the torque profiles of the motor actuated solution, the torque profiles are spiky. This comes from the basic explanation that applying a high magnitude torque on a joint in reverse directions for two successive timesteps will eventually cancel their effects. Thus, the net result lies around a lower mean, but this effect will eventually burden the motors. Generally, filters like moving average, low pass, infinite impulse, and finite impulse filters are employed to resolve this issue. They smoothen the torque profiles to generate smooth motor actuation.

Muscle-tendon actuated control actions, on the other hand, were found to generate very smooth motions inherently. MTUs are not directly actuated but are excited to certain excitation levels using neural spikings. As already discussed in previous sections, the transition from the current activation level of the MTU to the excited level follows differential activation dynamics, which acts as a regularizer. As the human body has such effective mechanisms, no filters are required. Even with spiky excitation actions, the resulting joint moment profiles are significantly smoother than motor actuation. Such a smooth motion not only consumes less energy but also enables good tracking of the planned movement.

Finally, a hyperparameter optimization strategy was discussed, which enabled consideration of the most optimal configuration concerning the neural networks of the learning algorithm, the contents of the observation space, and the overall formulation of the reward function. After each successive set of evaluating experiments, the best configuration was transferred to the next set of experiments. Finally, it was concluded that the policy neural network configuration with two hidden layers of $\{512, 512\}$ hidden nodes was the most optimal, and the observation space should contain:
\begin{itemize}
    \item the phase variable for better understanding of the correlation of the periodicity of the observations
    \item body positions and velocities relative to the pelvis in order to get an intuition between the coordinate space and the task space
    \item muscle activations to model the non-linear approximation for the muscle activation dynamics
\end{itemize}

\newpage

\chapter{Conclusion and Future Works}
\label{sec:conclusion}

The presented work has successfully analyzed various aspects of imitating a reference gait using model-free reinforcement learning methods, namely, Deep Deterministic Policy Gradient (DDPG) and Proximal Policy Optimization (PPO). The task was performed using both motor actuated as well as muscle-tendon actuated models. While the former is of interest to the robotics community, the latter deals with the biomechanics of physiological gaits. Various fundamental gaits that are generally performed in day-to-day activities like walking, running and jumping were considered in the sagittal plane, and an inverse dynamics policy to successfully imitate them was presented. Muscle actuated systems were found to be superior in comparison to motor actuated systems with their inherent regularizations. The study concludes that understanding muscle models is the key to understanding human motions execution patterns, the most efficient motion strategy.

While this work successfully delivers a framework for further experiments, many insights for more complicated gaits are yet to be explored. Furthermore, as this study was constrained to the motion in the sagittal plane, a complete set of 3D gaits will be explored, and the framework will be generalized to more complicated gaits. On the learning part of the project, various model-based learning algorithms will be explored and exploited to let the agent understand the complex dynamics of the interactions between the agent and the environment with deeper insights and precise approximation. 

\newpage

\begin{appendices}

\chapter{Additional Environment}
\label{appendix:A}

\begin{figure}[thpb]
      \centering
      \includegraphics[width=\linewidth]{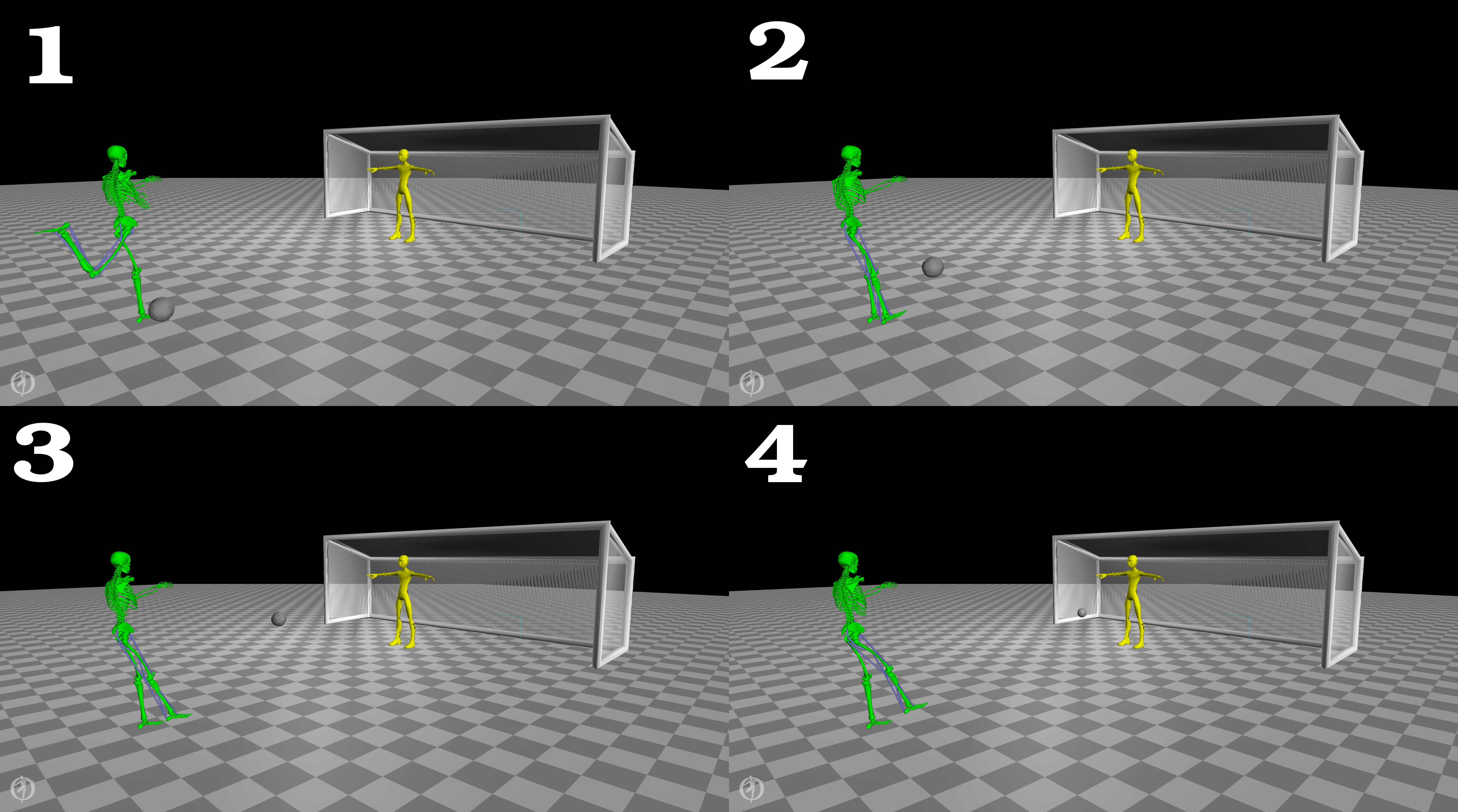}
      \caption[]{Kicking Environment was implemented and solved using the same RL framework except the imitation setting}
      \label{fig:kicking}
   \end{figure}
   
The environment and task description are given as follows:
\begin{itemize}
    \item \textbf{Environment Description:} The environment as shown in Fig. \ref{fig:kicking} represents a soccer penalty environment where a ball has to be directed in such a way that it falls within the range of the goal but outside the range of the goal-keeper. Appropriate ranges are defined using collision bodies of required sizes.
    \item \textbf{Model Description:} The model of the shooter is controllable and can be transformed between motor and MTU actuation. While one leg is fixed to the ground, there is an added degree of rotation about the pelvis.
    \item \textbf{Task Description:} Multiple tasks were tried, which included shooting at the goal and shooting at any target location in space. The idea is to understand how a ball can be shot in a proper direction with minimum effort.
    \item \textbf{Reward Description:} Rewards were formulated based on a linear combination of task completion reward and effort done by the shooter during the motion. Task completion rewards is indirectly proportional to the shortest distance between the ball's path and the target location.
\end{itemize}

\newpage

\chapter{Video Links for all Illustrations}
\label{appendix:B}

The complete set of videos include: 
\begin{itemize}
    \item Torque based Normal Walking: \\
Link: \href{https://utkarshmishra04.github.io/videos/BTP2021/comparison_walking_motor.mp4}{utkarshmishra04.github.io/videos/BTP2021/comparison\_walking\_motor.mp4}
    \item Torque based Prosthetic Walking (Locked Knee):\\
Link: \href{https://utkarshmishra04.github.io/videos/BTP2021/prosthetic_walking_motor.mp4}{utkarshmishra04.github.io/videos/BTP2021/prosthetic\_walking\_motor.mp4}
    \item Torque based Prosthetic Walking (Single Leg):\\
Link: \href{https://utkarshmishra04.github.io/videos/BTP2021/prosthetic_single_motor.mp4}{utkarshmishra04.github.io/videos/BTP2021/prosthetic\_single\_motor.mp4}
    \item Torque based Normal Running:\\
Link: \href{https://utkarshmishra04.github.io/videos/BTP2021/comparison_running_motor.mp4}{utkarshmishra04.github.io/videos/BTP2021/comparison\_running\_motor.mp4}
    \item Torque based High Jump (no stable landing):\\
Link: \href{https://utkarshmishra04.github.io/videos/BTP2021/highjump_motor.mp4}{utkarshmishra04.github.io/videos/BTP2021/highjump\_motor.mp4}
    \item Torque based Repeated Jumps (with stable landing):\\
Link: \href{https://utkarshmishra04.github.io/videos/BTP2021/continuous_jump_motor.mp4}{utkarshmishra04.github.io/videos/BTP2021/continuous\_jump\_motor.mp4}
    \item MTU based walking compared with reference:\\
Link: \href{https://utkarshmishra04.github.io/videos/BTP2021/comparison_walking_mtu.mp4}{utkarshmishra04.github.io/videos/BTP2021/comparison\_walking\_mtu.mp4}
    \item All Hyper-parameter experiments and evolution of an optimal policy:\\
Link: \href{https://utkarshmishra04.github.io/videos/BTP2021/all_experiments_mtu.mp4}{utkarshmishra04.github.io/videos/BTP2021/all\_experiments\_mtu.mp4}
    \item MTU based Prosthetic Walking (Locked Knee):\\
Link: \href{https://utkarshmishra04.github.io/videos/BTP2021/prosthetic_walking_mtu.mp4}{utkarshmishra04.github.io/videos/BTP2021/prosthetic\_walking\_mtu.mp4}
    \item Preliminary Solution for Kicking Task Environment:\\
Link: \href{https://utkarshmishra04.github.io/videos/BTP2021/kicking_preliminary.mp4}{utkarshmishra04.github.io/videos/BTP2021/kicking\_preliminary.mp4}
\end{itemize}

More Videos: \href{https://utkarshmishra04.github.io/redirects/BTP2021videos.html}{utkarshmishra04.github.io/redirects/BTP2021videos.html}

\end{appendices}


\newpage

\begin{thebibliography}{99}

\bibitem{c1} Delp, S. L., Anderson, F. C., Arnold, A. S., Loan, P., Habib, A., John, C. T., Guendelman, E., \& Thelen, D. G. (2007). OpenSim: Open-source software to create and analyze dynamic simulations of movement. IEEE Transactions on Biomedical Engineering. https://doi.org/10.1109/TBME.2007.901024

\bibitem{c2} Thelen, D. G., Anderson, F. C., \& Delp, S. L. (2003). Generating dynamic simulations of movement using computed muscle control. Journal of Biomechanics. https://doi.org/10.1016/S0021-9290(02)00432-3

\bibitem{c4} Van Den Bogert, A. J., Blana, D., \& Heinrich, D. (2011). Implicit methods for efficient musculoskeletal simulation and optimal control. Procedia IUTAM. https://doi.org/10.1016/j.piutam.2011.04.027

\bibitem{c5} Lee, L. F., \& Umberger, B. R. (2016). Generating optimal control simulations of musculoskeletal movement using OpenSim and MATLAB. PeerJ. https://doi.org/10.7717/peerj.1638

\bibitem{c6} Dembia, C. L., Bianco, N. A., Falisse, A., Hicks, J. L., \& Delp, S. L. (2019). OpenSim Moco: Musculoskeletal optimal control. In bioRxiv. https://doi.org/10.1101/839381

\bibitem{c7} Geijtenbeek, T. (2019). SCONE: Open Source Software for Predictive Simulation of Biological Motion. Journal of Open Source Software. https://doi.org/10.21105/joss.01421

\bibitem{c8} Peng, X. Bin, Berseth, G., Yin, K., \& Van De Panne, M. (2017). DeepLoco: Dynamic locomotion skills using hierarchical deep reinforcement learning. ACM Transactions on Graphics. https://doi.org/10.1145/3072959.3073602

\bibitem{c9} Peng, X. Bin, \& van de Panne, M. (2017). Learning locomotion skills using deep RL: Does the choice of action space maer? Proceedings - SCA 2017: ACM SIGGRAPH / Eurographics Symposium on Computer Animation. https://doi.org/10.1145/3099564.3099567

\bibitem{c10} Peng, X. Bin, Abbeel, P., Levine, S., \& Van De Panne, M. (2018). DeepMimic: Example-guided deep reinforcement learning of physics-based character skills. ACM Transactions on Graphics. https://doi.org/10.1145/3197517.3201311

\bibitem{c11} Lee, S., Park, M., Lee, K., \& Lee, J. (2019). Scalable muscle-actuated human simulation and control. ACM Transactions on Graphics. https://doi.org/10.1145/3306346.3322972

\bibitem{c12} Liu, L., \& Hodgins, J. (2018). Learning basketball dribbling skills using trajectory optimization and deep reinforcement learning. ACM Transactions on Graphics. https://doi.org/10.1145/3197517.3201315

\bibitem{c13} Yuan, Y., \& Kitani, K. M. (2020). Residual Force Control for Agile Human Behavior Imitation and Extended Motion Synthesis. In arXiv.

\bibitem{c14} Schulman, J., Wolski, F., Dhariwal, P., Radford, A., \& Klimov, O. (2017). Proximal policy optimization algorithms. In arXiv.

\bibitem{c15} Schulman, J., Levine, S., Moritz, P., Jordan, M., \& Abbeel, P. (2015). Trust region policy optimization. 32nd International Conference on Machine Learning, ICML 2015.

\bibitem{c16} Lillicrap, T. P., Hunt, J. J., Pritzel, A., Heess, N., Erez, T., Tassa, Y., Silver, D., \& Wierstra, D. (2016). Continuous control with deep reinforcement learning. 4th International Conference on Learning Representations, ICLR 2016 - Conference Track Proceedings.

\bibitem{wang2012} Jack M. Wang, Samuel R. Hamner, Scott L. Delp, and Vladlen Koltun. 2012. Optimizing locomotion controllers using biologically-based actuators and objectives. ACM Trans. Graph. 31, 4, Article 25 (July 2012), 11 pages. DOI:https://doi.org/10.1145/2185520.2185521

\end{thebibliography}
\end{document}